\documentclass{article}
\pdfoutput=1

\usepackage{arxiv}

\usepackage[utf8]{inputenc} 
\usepackage[T1]{fontenc}    
\usepackage{hyperref}       
\usepackage{nicefrac}       
\usepackage{microtype}      
\usepackage[numbers]{natbib}
\usepackage{doi}

\usepackage{tikz}
\usepackage{forest}
\usepackage{makecell}
\usepackage{multirow}
\usetikzlibrary{arrows.meta,shadows.blur}
\usepackage{amsmath}
\usepackage{stmaryrd}
\usepackage{hhline}
\usepackage{amsfonts}
\usepackage{mathtools}
\usepackage{fontawesome}
\usepackage{dsfont}
\usepackage{subcaption}
\usepackage{lscape}
\usepackage{tikz}
\usepackage{colortbl}
\usetikzlibrary{mindmap,trees}
\newtheorem{property}{Property}[section]
\newtheorem{definition}{Definition}[section]
\newtheorem{observation}{Observation}[section]
\newtheorem{query}{Query}[section]

\definecolor{scorecolor}{HTML}{8DD3C7}

\title{Semi-Supervised Constrained Clustering: An In-Depth Overview, Ranked Taxonomy and Future Research Directions}

\author{
        \href{https://orcid.org/0000-0001-7881-2023}{\includegraphics[scale=0.06]{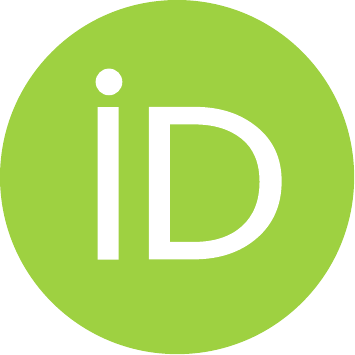}\hspace{1mm}Germ\'an Gonz\'alez-Almagro }\\
	DaSCI Andalusian Institute \\ DECSAI \\
	University of Granada\\
	Granada, Spain\\
	\texttt{germangalmagro@ugr.es} \\
        \And
        Daniel Peralta \\
	IDLab, Department of \\ Information Technology \\
	Ghent University - imec \\
	Ghent, Belgium\\
	\texttt{daniel.peralta@ugent.be} \\
        \And
        Eli De Poorter \\
	IDLab, Department of \\ Information Technology \\
	Ghent University - imec \\
	Ghent, Belgium\\
	\texttt{eli.depoorter@ugent.be} \\
	\And
	Jos\'e-Ram\'on Cano \\
        DaSCI Andalusian Institute \\
	Dept. of Computer Science \\
        University of Ja\'en\\
	Ja\'en, Spain\\
	\texttt{jrcano@ujaen.es} \\
	\And
	Salvador Garc\'ia \\
	DaSCI Andalusian Institute \\ DECSAI \\
	University of Granada\\
	Granada, Spain\\
	\texttt{salvagl@decsai.ugr.es} \\
}

\renewcommand{\headeright}{}
\renewcommand{\undertitle}{}
\renewcommand{\shorttitle}{}

\hypersetup{
pdftitle={Semi-Supervised Constrained Clustering: An In-Depth Overview, Ranked Taxonomy and Future Research Directions},
pdfsubject={Computer Sciences, Data Science},
pdfauthor={Germ\'an Gonz\'alez-Almagro },
pdfkeywords={Semi-supervised learning, Background knowledge, Pairwise instance-level constraints, Constrained clustering, Taxonomy}
}

\begin{document}
\maketitle

\begin{abstract}

Clustering is a well-known unsupervised machine learning approach capable of automatically grouping discrete sets of instances with similar characteristics. Constrained clustering is a semi-supervised extension to this process that can be used when expert knowledge is available to indicate constraints that can be exploited. Well-known examples of such constraints are must-link (indicating that two instances belong to the same group) and cannot-link (two instances definitely do not belong together). The research area of constrained clustering has grown significantly over the years with a large variety of new algorithms and more advanced types of constraints being proposed. However, no unifying overview is available to easily understand the wide variety of available methods, constraints and benchmarks. To remedy this, this study presents in-detail the background of constrained clustering and provides a novel ranked taxonomy of the types of constraints that can be used in constrained clustering. In addition, it focuses on the instance-level pairwise constraints, and gives an overview of its applications and its historical context. Finally, it presents a statistical analysis covering 307 constrained clustering methods, categorizes them according to their features, and provides a ranking score indicating which methods have the most potential based on their popularity and validation quality. Finally, based upon this analysis, potential pitfalls and future research directions are provided.

\end{abstract}


\section{Introduction}

Two major approaches characterize machine learning: supervised learning and unsupervised learning~\cite{bishop2006pattern}. In supervised learning, the goal is to build a classifier or regressor that, trained with a set of examples (or instances) $X$ and their corresponding output value $Y$, can predict the value of unseen inputs. In unsupervised learning, only the set of examples $X$ is available, and no output value is provided. In the latter, the goal is to discover some underlying structure in $X$. For example, in unsupervised clustering the goal is to infer a mapping from the input to clusters (groups) of similar instances. Generally, the set of examples $X$ is known as the dataset, and the set of output values $Y$ is known as the labels set.

Semi-Supervised Learning (SSL)~\cite{chapelle2010semi} is the branch of machine learning that tries to combine the benefits of these two approaches. To do so, it makes use of both unlabeled data and labeled data, or other kinds expert knowledge. For example, when considering classification or regression, in addition to a set with labeled data, an additional set of unlabeled data may be available, which can contain valuable information. Similarly, when considering clustering problems, a smaller subset of labeled data (or other types of knowledge about the dataset) may be available. Generally, some kind of information that does not fit within the supervised or unsupervised learning paradigm may be available to perform machine learning tasks. Ignoring or excluding this information does not optimally use all available information, thus the need of SSL~\cite{van2020survey}.

\subsection{On the feasibility of semi-supervised learning}

With regards to the applicability of SSL, a natural question arises~\cite{chapelle2010semi}: in comparison with supervised and unsupervised learning, can SSL obtain better results? It could be easily inferred that the answer to this question is ``yes'', otherwise neither this study nor all the cited before would exist. However, there is an important condition imposed for the answer to be affirmative: the distribution of instances in $X$ must be representative of the true distribution of the data. Formally, the underlying marginal distribution $p(X)$ over the input space must contain information about the posterior distribution $p(Y|X)$. Then, SSL is capable of making use of unlabeled data to obtain information about $p(X)$ and, therefore, about $p(Y|X)$~\citep{van2020survey}. Luckily, this condition appears to be fulfilled in most real-world learning problems, as suggested by the wide variety of fields in which SSL is successfully applied. Nonetheless, the way in which $p(X)$ and $p(Y|X)$ are related is not always the same. This gives place to the SSL assumptions, introduced in~\cite{chapelle2010semi} and formalized in~\cite{van2020survey}. A brief summary of these assumptions following~\cite{van2020survey} is presented, please refer to the cited studies for more details.

\begin{itemize}

    \item \textbf{Smoothness assumption}: two instances that are close in the input space should have the same label.
    
    \item \textbf{Low-density assumption}: decision boundaries should preferably pass through low-density regions in space.
    
    \item \textbf{Manifold assumption}: in problems in which data can be represented in Euclidean space, instances in the high-dimensional input space are usually gathered along lower-dimensional structures, known as manifolds: locally Euclidean topological spaces.
    
    \item \textbf{Cluster assumption}: data points which belong to the same cluster also belong to the same class. This assumption can be seen as a generalization of the other three specific assumptions.

\end{itemize}

As in other machine learning paradigms, the transduction versus induction dichotomy can be made within SSL. Usually, semi-supervised classification methods cope the SSL field, therefore the aforementioned dichotomy is explained in terms of classification as follows: 

\begin{itemize}

    \item \textbf{Inductive methods}: inductive methods aim to build a classifier capable of outputting a label for any instance in the input space. Unlabeled data can be used to train the classifier, but the predictions for unseen instances are independent of each other once the training phase is completed. An example of inductive method in supervised learning is linear regression~\cite{van2020survey}.
    
    \item \textbf{Transductive methods}: transductive methods do not build a classifier for the entire input space. Their predictions are limited to the data used during the training phase. Transductive methods do no have separated training and testing phases. An example of transductive method in unsupervised learning is hierarchical clustering~\cite{van2020survey}.
    
\end{itemize}

Classification methods within SSL can be clearly separated following the definitions above. However, when it comes to clustering, this distinction becomes unclear. Clustering methods within the SSL learning paradigm are usually considered to be transductive, as their output is still a set of labels partitioning the dataset and not a classification rule~\cite{chapelle2010semi}. On the other hand, some authors claim that partitional clustering methods can be considered as inductive methods, because their assignation rule can be used to predict the cluster membership of unseen instances. Hierarchical clustering methods would belong the transductive learning category, as no assignation rule can be derived from them~\cite{miyamoto2011inductive}. The differences between partitional and hierarchical clustering will be formalized later in Section~\ref{sec:constrained_clustering}.

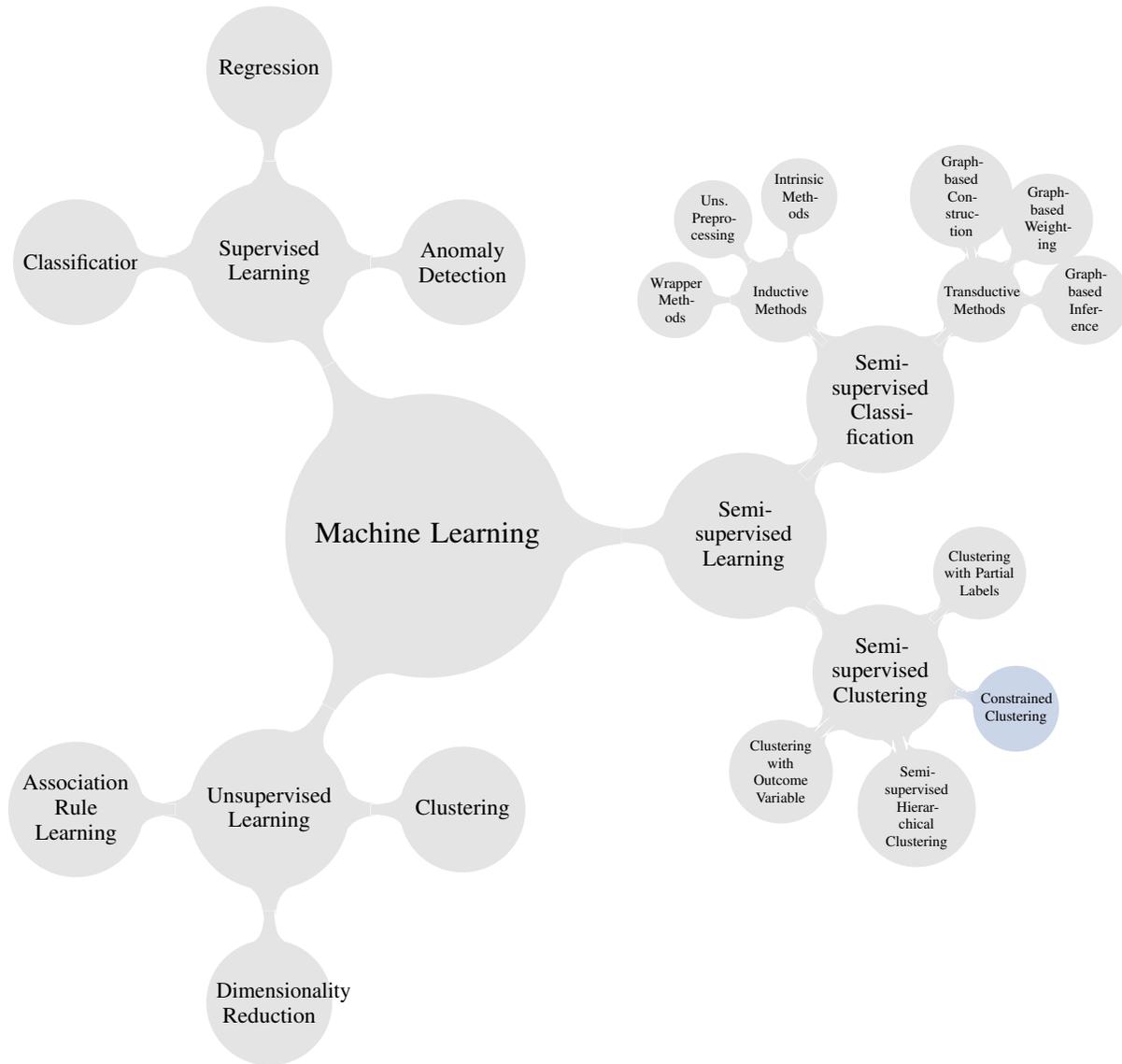
\begin{figure}[!ht]
	\centering
	\definecolor{airforceblue}{rgb}{0.36, 0.54, 0.66}
\definecolor{standardGray}{HTML}{bdbbbb}

\definecolor{color1}{HTML}{bdbbbb}
\definecolor{color2}{HTML}{bdbbbb}
\definecolor{color3}{HTML}{bdbbbb}
\definecolor{color4}{HTML}{bdbbbb}
\definecolor{color5}{HTML}{CBD5E8}

\begin{tikzpicture}[mindmap, grow cyclic, every node/.style=concept, concept color=color1!40,
	level 1/.append style={level distance=4.5cm,sibling angle=120},
	level 2/.append style={level distance=2.75cm,sibling angle=90},
	level 3/.append style={level distance=2cm,sibling angle=45},
	level 4/.append style={level distance=1.5cm,sibling angle=50}]

\node[concept]  {Machine Learning}
[clockwise from=120]
child [concept color=color2!40] { node {Supervised Learning}
    [clockwise from=180]
    child [concept color=color4!40] { node {Classification}}
	child [concept color=color4!40] { node {Regression}}
	child [concept color=color4!40] { node {Anomaly Detection}}
}
child [concept color=color2!40] { node {Semi-supervised Learning}
    [clockwise from=45]
    child[level 3/.append style={sibling angle=90}, concept color=color3!40] {node [concept] {Semi-supervised Classification}
        [clockwise from=135]
        child[concept color=color4!40] {node {Inductive Methods}
            [clockwise from=180]
            child [concept] { node {Wrapper Methods}}
        	child [concept] { node {Uns. Preprocessing}}
        	child [concept] { node {Intrinsic Methods}}
        }
        child [concept color=color4!40] {node {Transductive Methods}
            [clockwise from=100]
            child [concept] { node {Graph-based Construction}}
        	child [concept] { node {Graph-based Weighting}}
        	child [concept] { node {Graph-based Inference}}
        }
    }
    child[level 3/.append style={sibling angle=60}, concept color=color3!40] {node [concept] {Semi-supervised Clustering}
        [clockwise from=45]
	    child [concept color=color4!40] { node {Clustering with Partial Labels}}
    	child [concept color=color5] { node {Constrained Clustering}}
    	child [concept color=color4!40] { node {Semi-supervised Hierarchical Clustering}}
    	child [concept color=color4!40] { node {Clustering with Outcome Variable}}
    }
}
child [concept color=color2!40] { node {Unsupervised Learning}
    [clockwise from=0]
    child [concept color=color4!40] { node {Clustering}}
	child [concept color=color4!40] { node {Dimensionality Reduction}}
	child [concept color=color4!40] { node {Association Rule Learning}}
};
\end{tikzpicture}
	\caption{Mindmap of the machine learning overall landscape.}
	\label{fig:ml_mindmap}
\end{figure}

Figure~\ref{fig:ml_mindmap} helps us contextualize semi-supervised learning and its derivatives within the overall machine learning landscape. General SSL literature~\cite{zhu2005semi,chapelle2010semi,zhu2009introduction} usually divides SSL methods into two categories: semi-supervised classification and semi-supervised clustering. Further dichotomies have been made in later literature. In~\cite{van2020survey,zhou2021semi} semi-supervised classification methods are taxonomized taking into account the inductive versus transductive dichotomy. Some of the categories found in these taxonomies have been further studied:~\cite{subramanya2014graph} proposes a taxonomy for graph-based semi-supervised methods, and~\cite{triguero2015self} does the same for the self-labeling field. Concerning semi-supervised clustering,~\cite{bair2013semi} proposes a high level taxonomy with 4 types of methods, while~\cite{davidson2007survey,basu2008constrained} focus on the specific area of constrained clustering. The supervised and unsupervised learning paradigms are included in Figure~\ref{fig:ml_mindmap} for the sake of contextualization only. Consequently only classic and widely-known tasks belonging to these areas have been included in the diagram.

\subsection{Related work}

The semi-supervised clustering area has been widely studied and successfully applied in many fields since its inception. It can be seen as a generalization of the classic clustering problem which is able to include background knowledge into the clustering process~\cite{chapelle2010semi}. Many types of background knowledge have been considered in semi-supervised clustering~\cite{bair2013semi}, although the most studied one is the instance-level pairwise must-link and cannot-link constraints~\cite{basu2008constrained}. This type of background knowledge relates instances indicating if they belong to the same class (must-link) or to different classes (cannot-link). The problem of performing clustering in the presence of this type of background knowledge is referred to in literature as \textit{Constrained Clustering} (CC) (marked in Figure~\ref{fig:ml_mindmap} in blue).

This study carries out a comprehensive review of constrained clustering methods. It also proposes an objective scoring system, which addresses the potential and popularity of existing methods, and can be used to produce a sorted ranking for all of them. To the best of our knowledge, no similar study has been published before. Existing literature is either limited to the theoretical background on the topic, very limited in the number of methods reviewed, or outdated due to the rapid advance of the field. The earliest survey including constrained clustering in the reviewed studies can be found in~\cite{grira2004unsupervised}, although it is very limited in content. In~\cite{davidson2007survey}, the first survey focusing specifically on constrained clustering is proposed. It introduces many of the foundational concepts of subsequent studies and provides the first comprehensive reference on the area. However, this study was published in 2007, and even then it was limited to very few methods. The first book fully devoted to constrained clustering was published in~\cite{basu2008constrained} (2008). It provides unified formal background within the area and detailed studies on state-of-the-art methods.

\subsection{Remainder of this paper}

The rest of this study is organized as follows. Section~\ref{sec:Clustering_with_BK} presents a taxonomy of types of background knowledge with which semi-supervised clustering can work, including equivalencies between them in Subsection~\ref{subsec:equivalencies}. Section~\ref{sec:constrained_clustering} formalizes afterwards the constrained clustering problem, starting with basic background on classic clustering (Subsection~\ref{subsec:classic_clustering}) and pairwise constraints (Subsections~\ref{subsec:pairwise_constraints} and~\ref{subsec:fsblt_problem}), which is followed by a quick note on the history of constrained clustering (Subsection~\ref{subsec:cc_history}), and a comprehensive review on the applications of constrained clustering (Subsection~\ref{subsec:cc_applications}). Subsequently, advanced concepts regarding constrained clustering are introduced in Section~\ref{sec:cc_concepts_structures}. A statistical study on the experimental elements used to demonstrate the capabilities of CC methods is proposed in Section~\ref{sec:exp_setup}. This statistical study is used as the basis of the scoring system, which is presented in Section~\ref{sec:scoring_system} and used in subsequent sections to produce a ranking for all reviewed methods. Section~\ref{sec:Taxonomy} proposes a ranked taxonomic review of constrained clustering methods. A statistical analysis of the taxonomy is presented in Section~\ref{sec:statistic_analysis}. Finally, Section~\ref{sec:conclusions} presents conclusions, criticisms and future research guidelines.

\section{Clustering with Background Knowledge} \label{sec:Clustering_with_BK}

\begin{figure}
    \centering
    \includegraphics[width=\linewidth]{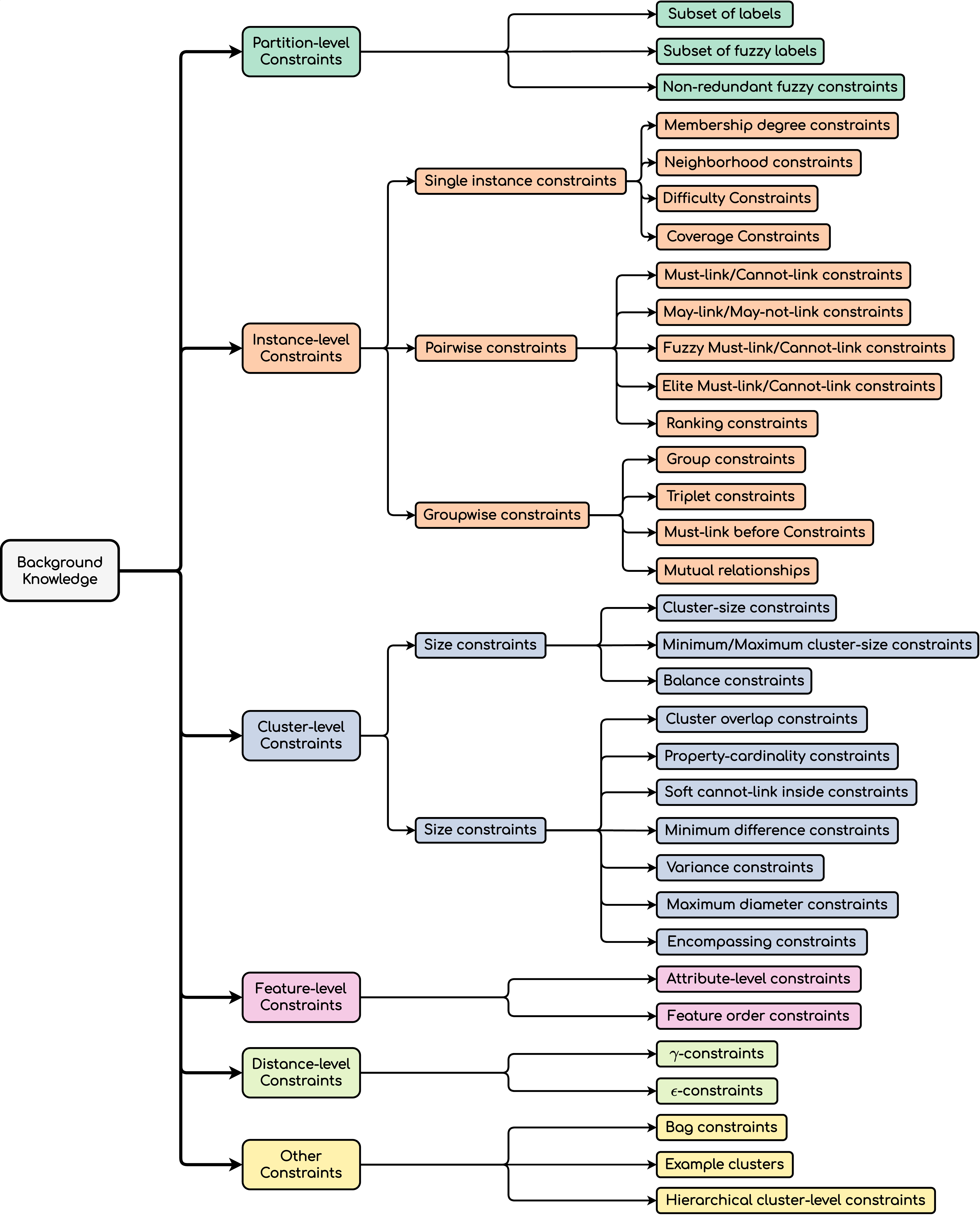}
    \caption{Taxonomy of types of background knowledge}
    \label{fig:BK_taxonomy}
\end{figure}

In this section, a comprehensive literature review on the types of background knowledge that have been used by semi-supervised clustering algorithm is carried out. In general terms, 5 families of background knowledge have been identified: partition-level constraints, instance-level constraints, cluster-level constraints, feature-level constraints, and distance constraints. Background information which does not belong to any of the mentioned categories has been placed together in a miscellaneous category. Figure~\ref{fig:BK_taxonomy} shows a visual representation of this taxonomy. All 5 families are composed by smaller, more specific categories which are detailed below.

\subsection{Partition-level Constraints} 

Partition-level constraints refer to restrictions imposed on the partition generated by the semi-supervised clustering algorithm~\cite{liu2015clustering,smieja2017semi,basu2003semi,liu2017partition}. Their most common form is a subset of labeled data, which is often referred simply as ``partition-level constraints'', although other categories within this type of background knowledge can be found:

\begin{itemize}

    \item \textbf{Subset of labels}: they consist of a subset of instances from the dataset for which labels are available. The resulting partition must be consistent with the given labels~\cite{basu2002semi,bohm2008hissclu,lelis2009semi,liu2010non,liu2011constrained,marussy2013success,lan2014soft,treechalong2015semi,choo2015weakly,vu2017graph,ienco2018semi,liu2019clustering,xie2019new,herrmann2007label}.
    
    \item \textbf{Subset of fuzzy labeled data}: used in fuzzy semi-supervised clustering algorithms. It consists of a subset of instances for which fuzzy labels are provided~\cite{zhang2009semi,yang2015robust,antoine2018evidential,gan2019safe,mei2018semisupervised,antoine2021fast}.
    
    \item \textbf{Non-redundant clusters constraints}: they constraint the output partition so that clusters in it must be orthogonal to each other, therefore maximizing their conditional mutual information and producing non-redundant clusters~\cite{gondek2007non}.
        
\end{itemize}

\subsection{Instance-level Constraints} 

Instance-level constraints can refer to single instances, pairs of instances or groups of multiple instances. In the case of single instance constraints, they are used to describe particular features of said instances or to restrict the features of the cluster they can belong to:

 \begin{itemize}
    
    \item \textbf{Membership degree constraints}: used in fuzzy semi-supervised clustering algorithm to provide prior membership degrees for some instances~\cite{yasunori2009semi,yin2012semi,zeng2013study}.
    
    \item \textbf{Neighborhood constraints}: they link instances to their neighborhood, with the latter being defined differently for every problem~\cite{nghiem2020constrained}.
    
    \item \textbf{Instance difficulty constraints}: they are referred to single instances and specify how hard it is to determine the cluster an instance belongs to, so that the semi-supervised clustering algorithm can focus on easy instances first~\cite{zhang2019framework}.
    
    \item \textbf{Coverage constraints}: for clustering algorithms which allow instances to belong to multiple clusters at the same time, this type of constraint limits the number of times an instance can be covered by different clusters~\cite{mueller2010integer}.
    
 \end{itemize}
 
Instance-level pairwise constraints involve pairs of instances and are used to indicate positive or negative relationships. The former refers to features that instances have in common, such as class or relevance, while the latter refers to the opposite case. Even if instance-level must-link and cannot-link are the most common form of instance-level constraints, the latter can be given in multiple ways:

\begin{itemize}

    \item \textbf{Must-link/Cannot-link constraints}: must-links involve pairs of instances that are known to belong to the same class. Therefore they must belong to the same cluster in the output partition. Cannot-link are used to indicate the opposite (the two instances involved in them are known to belong to different classes an thus they need to be placed in different clusters)~\cite{davidson2007survey}.

    \item \textbf{May-link/May-not-link constraints}: they represent soft must-link and cannot-link constraints respectively. This means that they can be violated in the output partition to some extent. They can be used in combination with the hard must-link and cannot-link constraints~\cite{ares2009avoiding}.
    
    \item \textbf{Fuzzy Must-link/Cannot-link constraints}: pairwise positive/negative relationships with and associated degree of belief~\cite{diaz2015use}.
    
    \item \textbf{Elite Must-link/Cannot-link constraints}: refined ML and CL constraints. They have the property of being unarguably satisfied in every optimal partition of the dataset~\cite{jiang2013extracting}.
    
    \item \textbf{Ranking constraints}: in contexts in which output class labels (clusters) can be ordered, ranking constraints are used to indicate whether an instance should be assigned a class label (cluster) higher that the class label of another instance~\cite{xiao2015maximum}.
    
\end{itemize}

The last form of instance-level constraints are group constraints, which are used to gather group of instances that are known to share features or to be different to each other in some aspect of their nature. They can also be used to set relative comparisons between a fixed number of constraints. Overall, they can be classified as follows: 

\begin{itemize}

    \item \textbf{Group constraints}: also referred to as \textbf{grouping information}~\cite{qian2016affinity,yu2004segmentation}. They specify the certainty of each or several instances belonging to the same cluster. Note that group constraint cannot be used to specify groups of instances that must not belong to the same cluster~\cite{law2004clustering}.
    
    \item \textbf{Triplet constraints}: also known as \textbf{relative constraints}~\cite{pei2016comparing,liu2011clustering,amid2015kernel}. They involve three instances: an anchor instance $a$, a positive instance $b$, and a negative instance $c$. A triplet constraint indicates that $a$ is more similar to $b$ than $c$~\cite{zhang2019framework,ienco2019deep}.
    
    \item \textbf{Must-link-before}: these are ML constraints specifically designed to be applied in hierarchical clustering setups. They involve triplets of constraints and their basic idea is to link instances positively not only in the output partition, but also in the hierarchy (dendrogram) produced by hierarchical clustering methods~\cite{bade2014hierarchical}.
    
    \item \textbf{Mutual relationships}: they establish a relation in groups of instances that is not known in advance and is determined during the clustering process. For example, a group of instances in the same mutual relation may be determined to belong to the same cluster during the cluster process, or contrarily they may be determined to not belong to the same cluster. Contrary to ML and CL constraints, mutual relations do not specify whether the nature of the relation they describe is positive or negative as part of the prior knowledge~\cite{endo2011fuzzy}.
    
\end{itemize}

\subsection{Cluster-level Constraints} 

Cluster-level constraints are used to restrict a wide variety of features related to clusters without specifying which instances must belong to these clusters. They are considered to be one of the most useful types of background knowledge, as they can convey large amounts of information compared to the amount of expert knowledge available. Size constraints are one of the forms in which cluster-level constraints can be found. They constraint the number of instances that clusters can have in the output partition and can be divided in three categories:

\begin{itemize}

    \item \textbf{Cluster-size constraints}: also called \textbf{cardinality constraints}~\cite{gallegos2010using}. They specify the number of instances each cluster must have in the output partition. The number of instances in a cluster may vary from a cluster to another~\cite{zhu2010data,tang2019size}.
    
    \item \textbf{Maximum/minimum cluster-size constraints}: they specify the maximum/minimum size a cluster can have in the output partition without specifying the exact size of each cluster~\cite{babaki2014constrained,dao2013declarative,bradley2000constrained,lei2013size,rebollo2013modification}. They may also be referred to as \textbf{significance constraints}~\cite{ge2007constraint}.
    
    \item \textbf{Balance constraints}: also known as \textbf{global size constraints}~\cite{zhang2019framework} applied in the cluster-level, they try to even the number of instances in every cluster (all cluster should be approximately the same size)~\cite{zhong2003unified,banerjee2006scalable,zhu2010data,tang2019optimizing}.
    
\end{itemize}

Apart from the size of the cluster, cluster-level constraints can restrict a wide variety of cluster features, ranging from their shape or separation, to the kind of instances they may contain:

\begin{itemize}
    
    \item \textbf{Cluster-overlap constraints}: they constraint the amount of overlap between clusters~\cite{nghiem2020constrained,mueller2010integer}.
    
    \item \textbf{Property-cardinality constraints}: they constraint the amount of a specific type of instance a cluster can contain~\cite{nghiem2020constrained}.
    
    \item \textbf{Soft cannot-link inside cluster constraints}: they require that the number of pairs of instances in a cluster which have a cannot-link constraint among them to be bounded~\cite{babaki2014constrained}.
    
    \item \textbf{Minimum difference constraints}: applied to pair of clusters, they require clusters to be similar or different to some degree~\cite{babaki2014constrained}.
    
    \item \textbf{Variance constraints}: they impose maximum or minimum values for the variance clusters must feature in the output partition~\cite{ge2007constraint}.
    
    \item \textbf{Maximum diameter constraints}: they specify an upper/lower bound on the diameter of the clusters~\cite{dao2013declarative}.
    
    \item \textbf{Encompassing constraints}: they determine whether clusters are allowed to encompass each other, i.e., they are allowed to form a hierarchy~\cite{mueller2010integer}.
    
\end{itemize}

\subsection{Feature-level Constraints} 

Feature-level constraints constraint instances by their feature values or directly relate pairs of feature to each other to indicate degrees of importance. Two types of feature-level constraints can be found:

\begin{itemize}

    \item \textbf{Attribute-level constraints}: they constraint the number of possible assignations for instances with specific values for specific features~\cite{nghiem2020constrained}.
    
    \item \textbf{Feature order constraints}: also called \textbf{feature order preferences}. They involve pairs of features and determine which one of them is more important. This is, what features need to be paid more attention to when performing comparisons to decide cluster memberships~\cite{sun2010clustering}.
    
\end{itemize}

\subsection{Distance Constraints} 

Distance constraints represent a very particular case of constraint-based information, as they relate pair of instances indirectly and in a global way. That means distance constraints can always be translated to instance-level must-link constraints~\cite{davidson2007survey}. Two types of distance constraints are defined in literature:

\begin{itemize}
    
    \item \textbf{$\gamma$-constraints}: also called \textbf{minimum margin}~\cite{dao2013declarative} or \textbf{minimum separation}~\cite{davidson2010sat}. They require the distance between two points of different clusters to be superior to a given threshold called $\gamma$~\cite{davidson2007survey,dao2013declarative,davidson2005agglomerative}.
    
    \item \textbf{$\epsilon$-constraints}: they require for each instance to have in its neighborhood of radius $\epsilon$ at least another point of the same cluster~\cite{davidson2007survey,dao2013declarative,davidson2005agglomerative}.
    
\end{itemize}

\subsection{Other Types of Constraints}

Finally, authors have proposed forms of background knowledge that do not fit into any of the previous categories:

\begin{itemize}
    
    \item \textbf{Bag constraints}: specific to the multi-instance multi-label framework, where datasets are given in the form of bags, with each bag containing multiple instances and labels, which provided only at the bag-level. Bag constraints specify similarities between bags~\cite{pei2014constrained}.
    
    \item \textbf{Example clusters}: predefined clusters in the dataset given to the clustering algorithm, which is required to output a partition which is consistent with example clusters. This information can be converted to instance-level pairwise constraints~\cite{hu2013generalizing,vens2013semi}.
    
    \item \textbf{Hierarchical cluster-level constraints}: sometimes also referred to as \textbf{ranking constraints}~\cite{ben2012mathcal,ben2012towards}. These constraints are designed to be applied only in semi-supervised hierarchical clustering methods. Given pairs of clusters, they specify which action (merge, split, remove, etc.) must be taken over them in successive steps of the clustering process that builds the output dendrogram~\cite{nogueira2017integrating}.
    
\end{itemize}

\subsection{Constraints Usability}

After analyzing the wide variety of forms in which constraints can be given, it is reasonable to ask which type of constraint is more effective for general purposes. There is not in fact a unique answer to this question, as it highly depends on the problem or applications and the type of information available to solve it. In~\cite{pei2016comparing} an empirical setup that tries to answer this question in a reduced semi-supervised environment is proposed. It only considers instance-level pairwise must-link and cannot-link constraints and subsets of labeled data as available sources of background knowledge. Three questions tried to be answered in the mentioned study, which can be reformulated to include a broader scope as follows:

\begin{itemize}
    \item Given the same amount of oracle effort, which type of background knowledge is more effective at aiding clustering?
    \item Which type of constraint is easier to obtain from the oracle?
    \item Which type of constraint is more reliable?
\end{itemize}

What it is meant here with an oracle is always understood as the source of background knowledge. This oracle can be a human, an automatic classifier, a crowdsourcing setup to gather information from distributed sources, etc. It is essential for any real-world or in-lab application of semi-supervised clustering to address these three questions.

\subsection{Equivalencies Between Types of Background Knowledge} \label{subsec:equivalencies}

\begin{figure}[!ht]
	\centering
	\includegraphics[width=\linewidth]{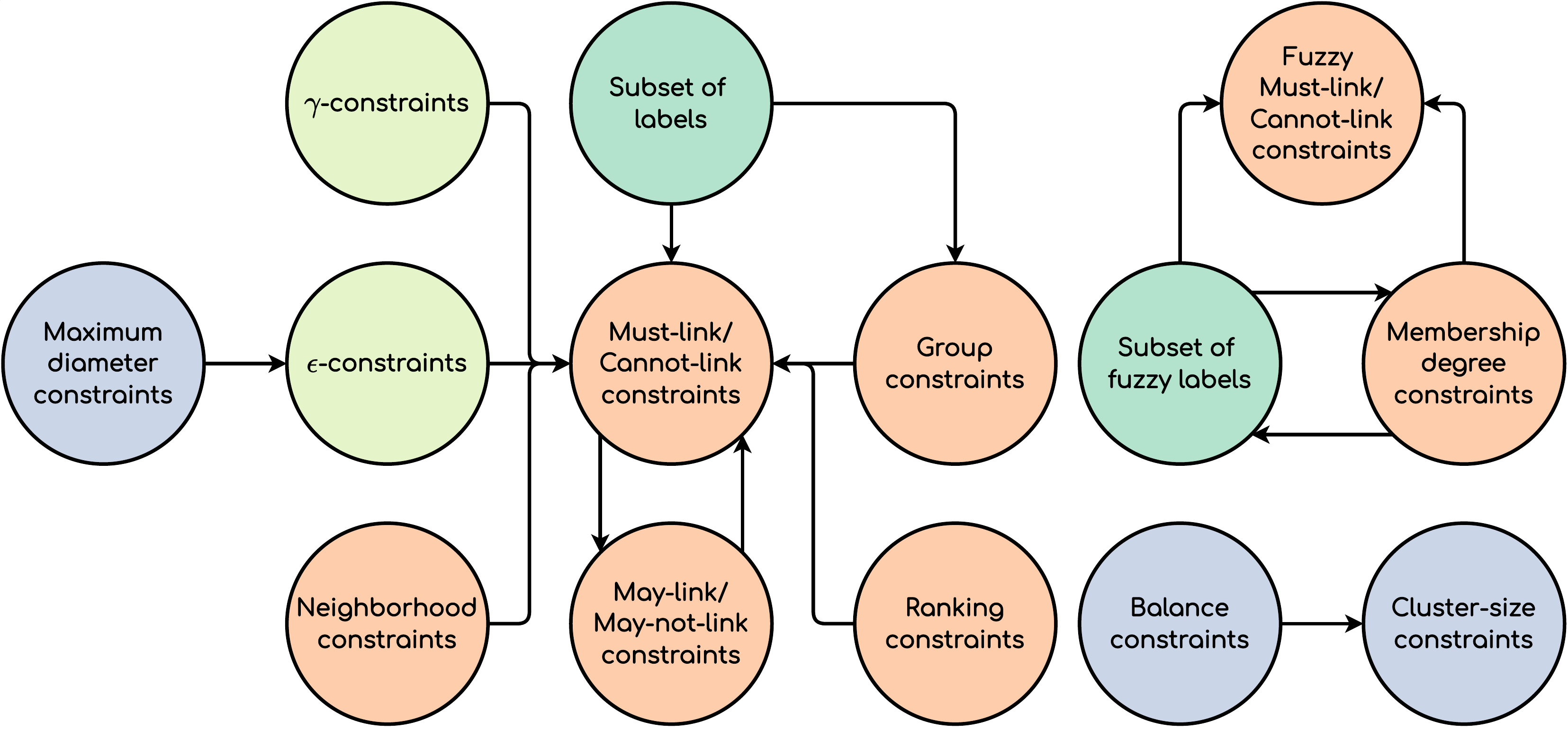}
	\caption{Graphical representation of direct equivalences between types of background knowledge.}
	\label{fig:BK_equivalencies}
\end{figure}

It is well known that some categories of background knowledge are neither isolated nor hermetic. Some types of constraints can be converted to another in a direct manner. Distance constraints can be translated to must-link constraints~\cite{davidson2007survey}, or a subset of labeled data can always be transformed in a set of must-link and cannot-link constraints~\cite{qian2016affinity}. The aim of this section is to provide intuition on all possible transformations without the need of a formal definition/notation for them, as this would require the length of a monography. Previous work on this line has been carried out in~\cite{qian2016affinity}, although within a much limited scope regarding the types of background knowledge considered. Figure~\ref{fig:BK_equivalencies} depicts equivalences found between the types of background knowledge introduced in Section~\ref{sec:Clustering_with_BK}.

In Figure~\ref{fig:BK_equivalencies}, only conversions without loss of information are considered, e.g.: fuzzy must-link/cannot-link constraints could be converted to must-link/cannot-link constraints by considering only those whose degree of belief is over 50\%, although this would involve losing not only constraints, but also the degree of belief information. Even if these kind of transformations are possible, they are not considered here, as they imply losing information.

\section{Instance-Level Pairwise Constrained Clustering} \label{sec:constrained_clustering}

Among all types of background knowledge reviewed in Section~\ref{sec:Clustering_with_BK}, pairwise constraints are undoubtedly one of the most studied topics, particularly basic must-link and cannot-link constraints, as it is shown later in this study. From now on, and for the sake of readability, \textbf{must-link and cannot-link constraints will be referred to simply as pairwise constraints}. In this section, the basic concepts of classic clustering and semi-supervised partitional and hierarchical clustering under pairwise constraints are introduced. \textbf{This problem is known in literature simply as Constrained Clustering (CC)}~\cite{davidson2007survey}. 

\subsection{Background on Classic Clustering} \label{subsec:classic_clustering}

Partitional clustering can be defined as the task of grouping the instances of a dataset into $K$ clusters. A dataset $X$ consists of $n$ instances, and each instance is described by $u$ features. More formally, $X = \{x_1, \cdots, x_n\}$, with the $i$th instance noted as $x_i = (x_{[i,1]}, \cdots, x_{[i,u]})$. A typical clustering algorithm assigns a class label $l_i$ to each instance $x_i \in X$. As a result, we obtain the list of labels $L = [l_1, \cdots, l_n]$, with $l_i \in \{1, \cdots, K\}$, that effectively splits $X$ into $K$ non-overlapping clusters $c_i$ to form a partition called $C$. The list of labels producing partition $C$ is referred to as $L^C$. The criterion used to assign an instance to a given cluster is the similarity to the rest of elements in that cluster, and the dissimilarity to the rest of instances of the dataset. This value can be obtained with some kind of distance measurement~\cite{jain1999data}.

Hierarchical clustering methods produce an informative hierarchical structure of clusters called dendrogram. Partitions as described above, with a number of clusters ranging from 1 to $n$, can always be obtained from a dendrogram by just selecting a level from its hierarchy and partitioning the dataset according to its structure. Typically, agglomerative hierarchical clustering methods start with a large number of clusters and iteratively merge them according to some affinity criteria until a stopping condition is reached. Every merge produces a new level in the hierarchy of the dendrogram. Formally, given an initial partition with $n_c$ clusters $C = \{c_1, \cdots, c_{n_c}\}$ (usually $n_c = n$), a traditional agglomerative constrained clustering method selects two clusters to merge by applying Equation~\ref{eq:general_aff}.

\begin{equation}
\{c_i, c_j\} =  \underset{c_i,c_j \in C, i\neq j}{\text{argmax}}A(c_i,c_j),
\label{eq:general_aff}
\end{equation}

\noindent with $A(\cdot, \cdot)$ being a function used to determine the affinity between the two clusters given as arguments. This function needs to be carefully chosen for every problem, as it greatly affects the result of the clustering process. Some conventional methods to measure affinity between clusters are worth mentioning, such as single linkage, average linkage and complete linkage~\cite{jain1999data}. Nevertheless, different measures are used in out-of-lab applications, as the manifold structures usually found in real-world datasets can be hardly captured by the classic affinity measures mentioned above. Typically, classic partitional clustering methods are algorithmically less complex than hierarchical clustering methods, with the former featuring $\mathcal{O}(n)$ complexity and the latter $\mathcal{O}(n^2)$~\cite{davidson2007survey}.

\subsection{Background on Pairwise Constraints} \label{subsec:pairwise_constraints}

In most clustering applications, it is common to have some kind of information about the dataset that will be analyzed. In CC this information is given in the form of pairs of instances that must, or must not, be assigned to the same cluster. We can now formalize these two types of constraints: 

\begin{itemize}

	\item Must-link (ML) constraints $C_=(x_i,x_j)$: instances $x_i$ and $x_j$ from $X$ must be placed in the same cluster. The set of ML constraints is referred to as $C_=$.

	\item Cannot-link (CL) constraints $C_{\neq}(x_i,x_j)$: instances $x_i$ and $x_j$ from $X$ cannot be assigned to the same cluster. The set of CL constraints is referred to as $C_{\neq}$.

\end{itemize}

The goal of constrained clustering is to find a partition (or clustering) of $K$ clusters $C = \{c_1, \cdots, c_K\}$ of the dataset $X$ that ideally satisfies all constraints in the union of both constraint sets, called $CS = C_= \bigcup C_{\neq}$. As in the original clustering problem, the sum of instances in each cluster $c_i$ is equal to the number of instances in $X$, which we have defined as $n = |X| = \sum_{i = 1}^{K} |c_i|$.

Knowing how a constraint is defined, ML constraints are an example of an equivalence relation; therefore, ML constraints are reflexive, transitive and symmetric. This way, given constraints $C_=(x_a,x_b)$ and $C_=(x_b,x_c)$, then $C_=(x_a,x_c)$ is verified. In addition to this, if $x_a \in c_i$ and $x_b \in c_j$ are related by $C_=(x_a,x_b)$, then $C_=(x_c,x_d)$ is verified for any $x_c \in c_i$ and $x_d \in c_j$~\cite{davidson2007survey}.

It can also be proven that CL constraints do not constitute an equivalence relation. However, analogously, given $x_a \in c_i$ and $x_b \in c_j$, and the constraint $C_{\neq}(x_a,x_b)$, then it is also true that $C_{\neq}(x_c,x_d)$ for any $x_c \in c_i$ and $x_d \in c_j$~\cite{davidson2007survey}.

Regarding the degree in which constraints need to be met in the output partition/dendrogram of any CC algorithm, a simple dichotomy can be made: hard pairwise constraints must necessarily be satisfied, while soft pairwise constraints can be violated to a variable extent. This distinction is introduced in~\cite{davidson2007survey} and adopted by later studies, eventually producing the ``may constraints'' (may-link/may-not link constraints mentioned in Section~\ref{sec:Clustering_with_BK}), which can be seen as the formalization of soft constraints. However, the scientific community still refers to ``may constraints'' as soft constraints in the majority of the cases and the terms may-link and may-not link are used only in cases in which both soft and hard constraints are mixed and can be considered by the same CC algorithm.  The major advantages in favor of soft over hard constraints are found in the resiliency to noise in the constraint set, the flexibility on the design of cost/objective functions, and their optimization procedures. The ability to consider soft, hard, or both types of constraints is a defining element for CC methods.

In~\cite{davidson2006measuring} two measures designed to characterize the quality of a given constraint set are proposed: \textbf{informativeness} (or informativity~\cite{davidson2007survey}) is used to determine the amount of information in the constraint set that the CC algorithm could determine on its own, and \textbf{coherence}, which measures the amount of agreement between the constraints themselves. These two measures were proposed in early stages of the development of the CC area; however, they have not been used consistently in later studies.

\subsection{The Feasibility Problem} \label{subsec:fsblt_problem}

Given that CC adds a new element to the clustering problem, we must consider how it affects the complexity of the problem in both of its forms: partitional and hierarchical. Intuitively, the clustering problem goes from its classic formulation ``find the best partition for a given dataset'' to its constrained form ``find the best partition for a given dataset satisfying all constraints in the constraint set''. The formalization of this problem is tackled in~\cite{davidson2005clustering,davidson2007survey,davidson2009using}, where the feasibility problems for partitional and hierarchical CC are defined as in~\ref{def:feass_prblm_PCC} and~\ref{def:feass_prblm_HCC} respectively, where $CS = C_{\neq} \cup C_=$ (the joint constraint set). Given these two definitions, we say that \textbf{a partition $C$ for a dataset $X$ is feasible when all constraints in $CS$ are satisfied by $C$}. Note that there exist constraint sets for which a feasible partition can never be found, e.g., no feasible partition exist for $CS_1 = \{C_=(x_1,x_2), C_{\neq}(x_1,x_2)\}$ regardless of the value of $K$. Similarly, the feasibility of partitions such as $CS_2 = \{C_{\neq}(x_1,x_2), C_{\neq}{\neq}(x_2,x_3)C_{\neq},(x_1x_3)\}$ depends on the value of $K$. In this case, the feasibility problem for $CS_2$ can be solved for $K=3$ but not for $K=2$.

\begin{definition}

	\textbf{Feasibility Problem for Partitional CC}: given a dataset $X$, a constraint set $CS$, and the bounds on the number of clusters $k_l \leq K \leq k_u$, is there a partition $C$ of $X$ with $K$ clusters that satisfies all constraints in $CS$?~\cite{davidson2005clustering}
	\label{def:feass_prblm_PCC}

\end{definition}

In~\cite{davidson2005clustering} it is proven that, when $k_l = 1$ and $k_u \ge 3$, the feasibility problem for partitional CC is $\mathbf{NP}$-complete, by reducing it from the Graph K-Colorability problem. It is also proven that it is not harder, so both have the same complexity. Table~\ref{tab:fsblt_prblm_complexity} shows the complexity of the feasibility for different types of constraints.

\begin{definition}
	
	\textbf{Feasibility Problem for Hierarchical CC}: given a dataset $X$, the constraint sets $CS$, and the symmetric distance measure $D(x_i, x_j) \ge 0$ for each pair of instances: Can $X$ be partitioned into clusters so that all constraints in $CS$ are satisfied?~\cite{davidson2009using}
	\label{def:feass_prblm_HCC}
	
\end{definition}

Please note that the definition of the feasibility problem for partitional CC (in Definition~\ref{def:feass_prblm_PCC}) is significantly different from the definition of the feasibility problem for hierarchical CC (in~\ref{def:feass_prblm_HCC}). Particularly, the formulation for the hierarchical CC does not include any restriction on the number of clusters $K$, which is equivalent to considering that any level of the dendrogram can be used to produce the partition that satisfies all constraints~\cite{davidson2009using}.  In~\cite{davidson2005agglomerative} a reduction from the One-in-three 3SAT with positive literals problem (which is $\mathbf{NP}$-complete) for the problem in Definition~\ref{def:feass_prblm_HCC} is used to prove the complexities presented in Table~\ref{tab:fsblt_prblm_complexity} for the hierarchical CC problem. It is worth mentioning that, for the hierarchical CC problem, the dead-ends problem arises: a hierarchical CC algorithm may find scenarios where no merge/split can be carried out without violating a constraint. Previous solutions based on the transitive closure of the constraint sets have been proposed to this problem, although they imply not generating a full dendrogram~\cite{davidson2005clustering}.

\begin{table}[!h]
	\centering
	\setlength{\tabcolsep}{7pt}
	\renewcommand{\arraystretch}{1.3}
		\begin{tabular}{c c c c}
			\hline
			Constraints & Partitional CC & Hierarchical CC & Dead Ends?\\
			\hline
			ML & $\mathbf{P}$ & $\mathbf{P}$ & No\\
			CL & $\mathbf{NP}$-complete & $\mathbf{NP}$-complete & Yes\\
			ML and CL & $\mathbf{NP}$-complete & $\mathbf{NP}$-complete & Yes\\
			\hline

		\end{tabular}
	\caption{Feasibility problem complexities for partitional and hierarchical CC and dead-ends found in hierarchical CC~\cite{davidson2005clustering}.}
	\label{tab:fsblt_prblm_complexity}
\end{table}

Overall, complexity results in Table~\ref{tab:fsblt_prblm_complexity} show that the feasibility problem under CL constraints is intractable, hence constrained clustering is intractable too. This leads to Observation~\ref{obv:not_help}.  For more details on the complexity of constrained clustering please see~\cite{davidson2005clustering}. 

\begin{observation}
    \textbf{Knowing that a feasible solution exists does not help us find it}. The results from Table~\ref{tab:fsblt_prblm_complexity} imply that the fact that there is a feasible solution for a given set of constraints does not mean it will be easy to find.
    \label{obv:not_help}
\end{observation}

With respect to the dead-ends problem, a full dendrogram considering constraints can be obtained by switching from a hard interpretation of constraints to a soft one. This means that every level in the dendrogram tries to satisfy as many constraints as possible, but constraint violations are allowed in order for the algorithm to never reach a dead-end.

Some interesting results, both positive and negative, about the nature of pairwise constraints are proved and discussed in~\cite{davidson2007survey}, as well as some workarounds for problems related to the use of constraints in clustering.

\subsection{Early History of Constrained Clustering} \label{subsec:cc_history}

The Constrained Clustering problem has been rediscovered and renamed throughout years of evolution, firstly in mathematical science, secondly in Computer Science. The first reference to the CC problem was proposed by Harary in~\cite{harary1953notion} as early as in the year 1953. Harary introduced the \textit{signed graph}, which is an undirected graph with +1 or -1 labels on its edges, respectively indicating similarity of dissimilarity between the vertices they connect. This can be directly translated to the ML and CL constraints that shape the CC problem. Besides, Harary introduced the concept of \textit{imbalance} for a 2-way partitioning of such signed graph, which referred to the number of vertices violated by the partitioning. The aim of Harary was to find highly related groups of vertices within a psychological interpretation of the problem: positive edges correspond to pairs of people who like one another, and negative edges to pairs who dislike one another.

It was not until year 2000 that the name Constrained Clustering made its first appearance by the work of K. Wagstaff and C. Cardie in~\cite{wagstaff2000clusteringB}, which is a brief paper that introduces later work by the same authors in which the first two CC algorithms in the history of Computer Science are proposed: COP-COBWEB~\cite{wagstaff2000clustering} and COP-K-Means~\cite{wagstaff2001constrained} in 2000 and 2001 respectively. These two papers set the precedent for a new area in semi-supervised learning known as Constrained Clustering, providing experimental procedures and baselines to compare with.

On the one hand, and following the trend set by K. Wagstaff and C. Cardie, although in separate studies, S. Basu proposes in 2003 the first two soft constrained approaches to CC in~\cite{basu2003comparing}: the PCK-Means and MPCK-Means algorithms. Later, in the year 2005, I. Davidson and S.S. Ravi would propose the first hierarchical approaches to the CC problem~\cite{davidson2005agglomerative}. On the other hand, E. Xing et al. propose the first distance metric learning based approach to CC with their CSI algorithm~\cite{xing2002distance}, also known in literature simply as Xing's algorithm. Finally, in 2008, S. Basu, I. Davidson and K. Wagstaff joined forces to produce the first book fully dedicated to constrained clustering in~\cite{basu2008constrained}.

Research in CC has followed the general trend in Computer Science ever since. Ranging from well-studied classic clustering approaches, such as fuzzy clustering~\cite{li2018k}, spectral clustering~\cite{nie2021semi} or non-negative matrix factorization~\cite{zong2018multi}, to modern and general optimization models like classic~\cite{denoeux2021nn} or deep~\cite{amirizadeh2021cdec} neural networks and evolutive~\cite{gonzalez2021me} or non-evolutive~\cite{gonzalez2020dils} metaheuristic algorithms.

\subsection{Applications of Constrained Clustering} \label{subsec:cc_applications}

\begin{figure}[!ht]
	\centering
	\includegraphics[width=0.7\linewidth]{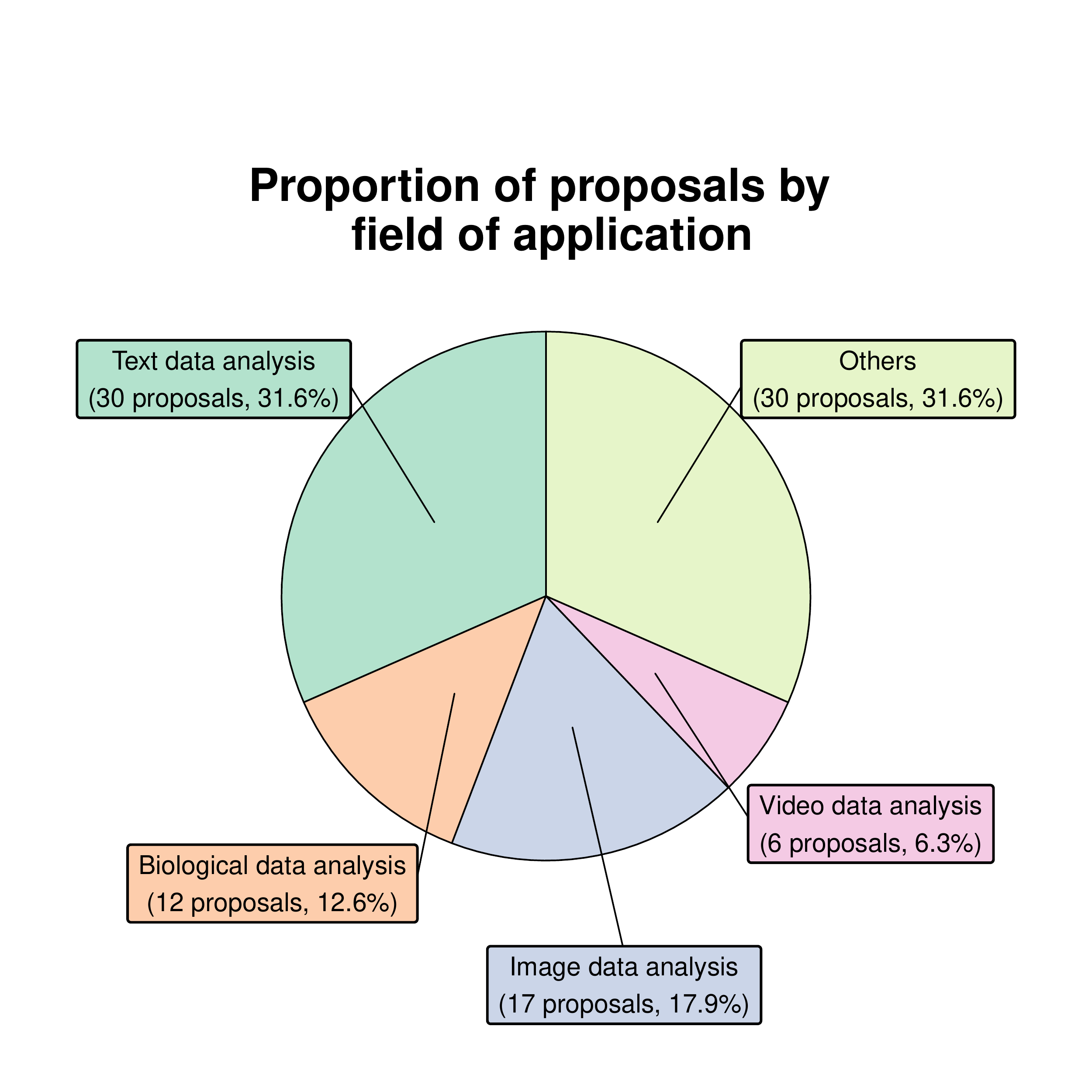}
	\caption{Piechart showing a summary of the overall CC application field.}
	\label{fig:pieplot_applications}
\end{figure}

CC has been applied in many fields since its inception. The first applications are gathered in~\cite{davidson2007survey}, which include clustering of image data, video and sound data, biological data, text data, web data, and the first application of CC found in~\cite{wagstaff2001constrained}, which is lane finding for vehicles in GPS data. Figure~\ref{fig:pieplot_applications} shows a summary of the overall CC application field.

Table~\ref{tab:wide_field_app} gathers CC applications, sorting them by application field and indicating the specific purpose of every application. The field with the largest number of publications is text data analysis; within it,  document clustering has attracted the most publications. Text data clustering is followed by three other wide application fields, which are biological data analysis, image data analysis and video data analysis.

\begin{table}[!h]
	\centering
	\setlength{\tabcolsep}{7pt}
	\renewcommand{\arraystretch}{1.3}
	\resizebox{\textwidth}{!}{
		\begin{tabular}{c  c  cl c c}
			\hline
			Field of application & No. of studies && Specific application & No. of studies & References \\
			\cline{1-2} \cline{4-6}
			\multirow{7}{*}{Text data analysis} & \multirow{7}{*}{30}
            && Document clustering & 19 & \makecell{\cite{huang2007semi}\cite{huang2009active}\cite{ma2010orthogonal}\cite{yan2013fuzzy}\\\cite{liu2013clustering}\cite{ma2013nonnegative}\cite{sutanto2014ranking}\cite{hu2016document}\\\cite{diaz2016automatic}\cite{trabelsi2016mining}\cite{balafar2020active}\cite{buatoom2020document}\\\cite{van2017cobra}\cite{hu2008towards}\cite{chen2009semi}\cite{ma2010orthogonal}\\\cite{zhao2012effective}\cite{yan2013semi}\cite{gu2012efficient}} \\
            &&& Text Clustering & 4 &~\cite{ares2009avoiding}\cite{parapar2012language}\cite{song2012constrained}\cite{zhu2011text} \\
            &&& Verb clustering & 1 &~\cite{vlachos2009unsupervised} \\
            &&& Word disambiguation & 1 &~\cite{sugiyama2009semi} \\
            &&& Microblog clustering & 3 &\cite{ma2014semi}\cite{ma2015semi}\cite{ma2019term} \\
            &&& Document filtering & 1 &~\cite{tang2005user} \\
            &&& Clustering in online forums & 1 &~\cite{duan2015user} \\
			\hline
            \multirow{6}{*}{Biological data analysis} & \multirow{6}{*}{12}
            && Gene Expression & 7 & \makecell{\cite{tseng2008constrained}\cite{maraziotis2012semi}\cite{saha2015use}\\\cite{maraziotis2006semi}\cite{maraziotis2012semi}\cite{yu2014double}\cite{ceccarelli2005semi}} \\
            &&& Gene Clustering & 1 &~\cite{zeng2007effectiveness} \\
            &&& RNA-seq Data Clustering & 1 &~\cite{tian2021model} \\
            &&& Regulatory Module Discovery & 1 &~\cite{mishra2008semi} \\
            &&& Fiber Segmentation & 1 &~\cite{abdala2010fiber} \\
            &&& Biomolecular Data & 1 &~\cite{yu2011knowledge} \\
			\hline
		\multirow{10}{*}{Image data analysis} & \multirow{10}{*}{17}
            && Medical image  & 2 &~\cite{verma2016improved}\cite{veras2018medical} \\
            &&& Image segmentation & 3 &~\cite{wang2018non}\cite{guo2014consistent}\cite{li2020spatial} \\
            &&& Image clustering & 3 &~\cite{biswas2011large}\cite{hoi2010semi}\cite{biswas2014active} \\
            &&& Image categorization & 2 &~\cite{grira2006fuzzy}\cite{frigui2008image} \\
            &&& Image annotation & 2 &~\cite{ismail2012automatic}\cite{xiaoguang2006image} \\
            &&& Image indexing & 1 &~\cite{lai2014new} \\
            &&& Point of interest mining & 1 &~\cite{bui2017point} \\
            &&& Multi-target detection & 1 &~\cite{li2017multi} \\
            &&& Face recognition & 1 &~\cite{vu2019efficient} \\
            &&& Satellite Image Time Series & 1 &~\cite{lafabregue2019deep} \\
			\hline
            \multirow{4}{*}{Video data analysis} & \multirow{4}{*}{6}
            && Extracting moving people & 1 &~\cite{niebles2008extracting} \\
            &&& Web Video Categorization & 1 &~\cite{mahmood2013semi} \\
            &&& Face Clustering & 2 &~\cite{wu2013constrained}\cite{wu2017coupled} \\
            &&& Face Tracking & 2 &~\cite{wu2013simultaneous}\cite{wu2017coupled} \\
            \hline
		\end{tabular}}
	\caption{Comprehensive listing of applications of CC in wide application fields.}
	\label{tab:wide_field_app}
\end{table}

However, some CC applications are very specific and cannot be grouped into wider application fields. Studies which bring forward this kind of applications are listed in Table~\ref{tab:particular_app}. 

\begin{table}[!h]
	\centering
	\setlength{\tabcolsep}{7pt}
	\renewcommand{\arraystretch}{1.3}
	\resizebox{\textwidth}{!}{
		\begin{tabular}{l c c l c}
			\hline
			Field of Application & References && Field of Application & References \\
			\cline{1-2} \cline{4-5}
			Identifying speakers in a conversation through audio data &~\cite{xie2006improved} && Lane finding for vehicles in GPS data &~\cite{wagstaff2001constrained} \\
			
			Clustering of software requirements &~\cite{duan2008consensus} && Optimization of rural ecological endowment industry &~\cite{zhao2022transformation} \\
		
			Machinery fault diagnosis &~\cite{li2010evolving} && Job-shopping scheduling &~\cite{el2022decomposition} \\
			
			Patient Segmentation from medical data &~\cite{han2016intelligible}\cite{zhang2019semi} && Trace-clustering &~\cite{de2021expert} \\
			
			Direct marketing applications &~\cite{chang2011group}~\cite{yu2021trust}~\cite{seret2014new}~\cite{akaichi2021pairwise} && Discovering educational-based life patterns &~\cite{zhang2021semi} \\
			
			Group extraction from professional social network &~\cite{ben2014group} && Oil price prediction &~\cite{boesen2021data} \\
			
			Clustering of cognitive radio sensor networks & ~\cite{shah2014spectrum} && Traffic analysis &~\cite{zhang2019towards}\cite{malzer2021constraint} \\
			
			District design &~\cite{joshi2011redistricting}\cite{song2021adaptive} && Vocabulary maintenance policy for CBR systems &~\cite{ben2019cevm} \\
			
			Sentiment analysis &~\cite{araujo2014audio}\cite{xiong2015exploiting} && Obstructive sleep apnea analysis &~\cite{mai2018evolutionary} \\ 
			
			Sketch symbol recognition &~\cite{tirkaz2012sketched} && Internet traffic classification &~\cite{wang2013internet}\\
			
			Robot navigation systems &~\cite{semnani2016constrained} && Social event detection &~\cite{sutanto2014ranking}\\
			
			Terrorist community detection &~\cite{saidi2018novel} && & \\
			\hline
			
		\end{tabular}}
	\caption{Comprehensive listing of particular applications of CC.}
	\label{tab:particular_app}
\end{table}

\section{Constrained Clustering Concepts and Structures} \label{sec:cc_concepts_structures}

Within the CC research field, some concepts and data structures are repeatedly mentioned and used by researchers. The goal of this section is to provide a formal definition of these concepts, as they will be mentioned later and are necessary for the reader to have a good understanding of the methods described later in Section~\ref{sec:Taxonomy}. From now on, and for the sake of readability and ease of writing, \textbf{we refer to instances involved in a constraint simply as \textit{constrained instances}, and to ML constraints and CL constraints as simply ML and CL, respectively}. Instances involved in ML are referred to as ML-constrained instances and instances involved in CL are referred to as CL-constrained instances.

\paragraph{\textbf{The Constraint Matrix}} This is one of the most, if not the most, basic and most frequently used data structures to store the information contained in the constraint set. It is a symmetric matrix, with as many rows and columns as instances in the dataset, filled with three values: 0 to indicate no constraint between the instances associated with the row and column in which it is stored, 1 is used for ML and -1 is used for CL. Formally, the Constraint Matrix is a matrix $CM_{n\times n}$ filled as in Equation~\ref{eq:const_mat}.

\begin{equation}
CM_{[i,j]} = CM_{[j,i]} = \left\{ \begin{array}{lc}
1 & \text{if} \;\; C_=(x_i, x_j) \in CS\\
-1 & \text{if} \;\; C_{\neq}(x_i, x_j) \in CS\\
0 & \text{otherwise}
\end{array}\right.
\label{eq:const_mat}
\end{equation}

Please note that, following this definition for the constraint matrix, its diagonal may be assigned to all 1 or all 0. That possibility depends on whether ML with the form $C_=(x_i,x_i)$ are included or not in $CS$, respectively. The inclusion of such constraints may be convenient in some cases. Variants of this matrix are also commonly used. In some cases, the constraint matrix can store any value in the range $[-1,1]$, with negative values indicating the weight or degree of belief for CL and positive values doing so for ML.

\paragraph{\textbf{The Constrain List}} It is a list, with length equal to the number of constraints, that stores triplets with two values used to specify two instances and a third value used to indicate the type of constraint between them (1 for ML and -1 for CL). Formally, the Constraint List $CL$ contains $|C_=|$ triplets with the form $[i,j,1]$ for ML such that $C_=(x_i, x_j)$ and $|C_{\neq}|$ triplets with the form $[i,j,-1]$ for CL such that $C_{\neq}(x_i, x_j)$.

The Constraint List is used in methods in which the number of violated constraints needs to be repeatedly computed over fully formed partitions that are not built incrementally. In these cases, the only option is to iterate over the full constraint set and check individually for every constraint whether it is violated by the partition. This task is performed efficiently iterating over $CL$, which is $\mathcal{O}(|CS|)$, in contrast with $CM$, which requires an $\mathcal{O}(n^2)$ computation of the number of violated constraints. However, checking for specific constraint violations in iterative partition building processes can be done in $\mathcal{O}(1)$ with $CM$, as the indexes of the constrained instances are known and matrices support random access. The same task can be performed over $CL$, but with the much higher computation cost of $\mathcal{O}(|CS|)$.

\paragraph{\textbf{The Constraint Graph}} It is a weighted, undirected graph with a one vertex per instance in the dataset and one edge per constraint. An edge connects two instances if they are involved in a constraint, with the weight of the edge indicating the type of constraints, using 1 for ML and -1 for CL. Formally, let an undirected weighted graph $G(V,E,W)$ be a finite set of vertices $V$, a set of edges $E$ over $V \times V$ and a set of weights $W$ for every edge in $E$. In the constraint graph $CG(V,E,W)$, $V$ is the set of instances in $X$, and edges $e(x_i,x_j)$ from $E$ are equivalent to constraints in $CS$, using the weight of the edge $w_{[i,j]}$ as indicator for the type of constraint, i.e., for edge $e(x_i,x_j)$, if $C_=(x_i,x_j) \in CS$ (ML) then $w_{[i,j]} = 1$, and if $C_{\neq}(x_i,x_j) \in CS$ (CL) then $w_{[i,j]} = -1$~\cite{yoshida2010performance}.

\begin{figure}[!ht]
	\centering
	\includegraphics[width=0.6\linewidth]{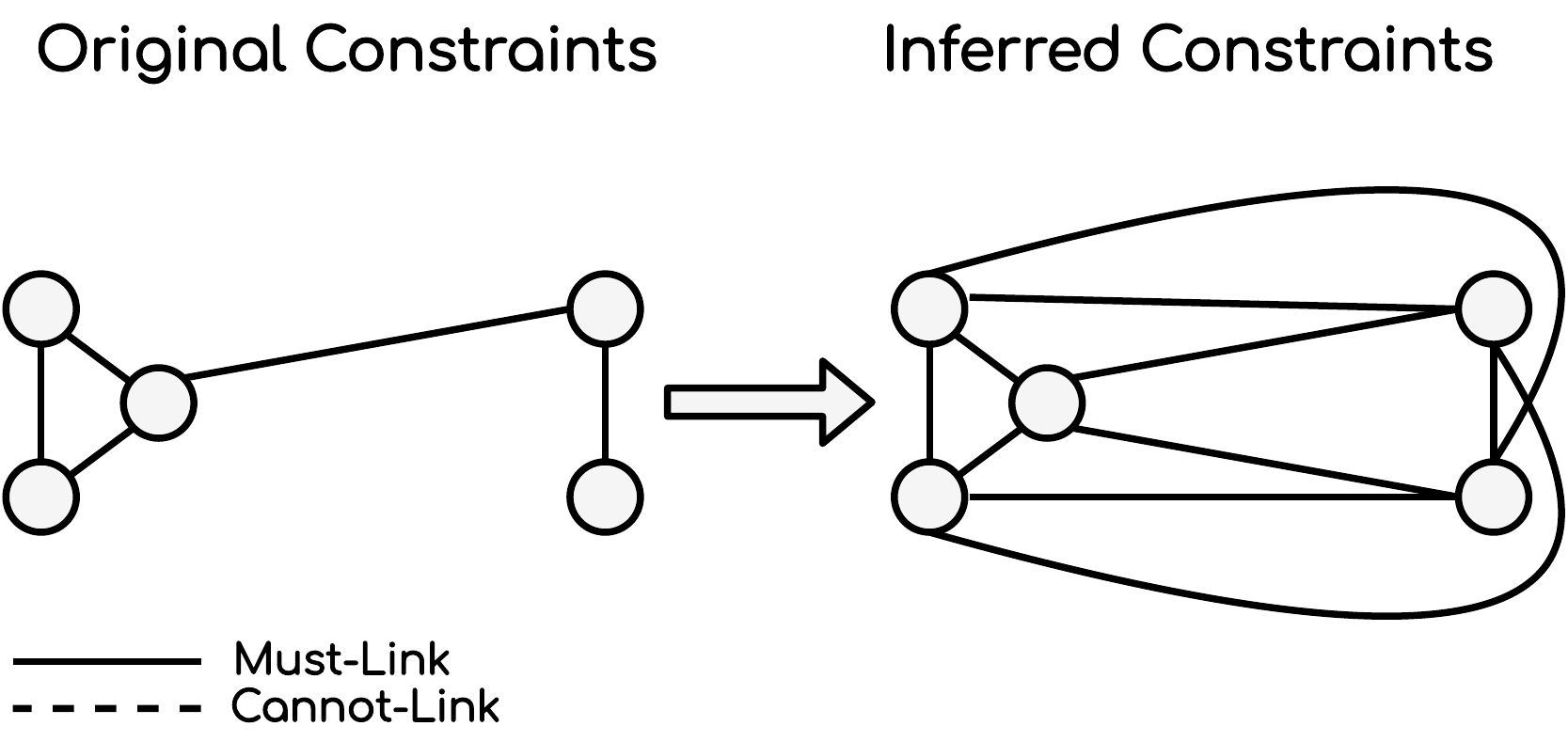}
	\caption{Example of transitive inference of ML constraints.}
	\label{fig:inference_ml}
\end{figure}

\begin{figure}[!ht]
	\centering
	\includegraphics[width=0.6\linewidth]{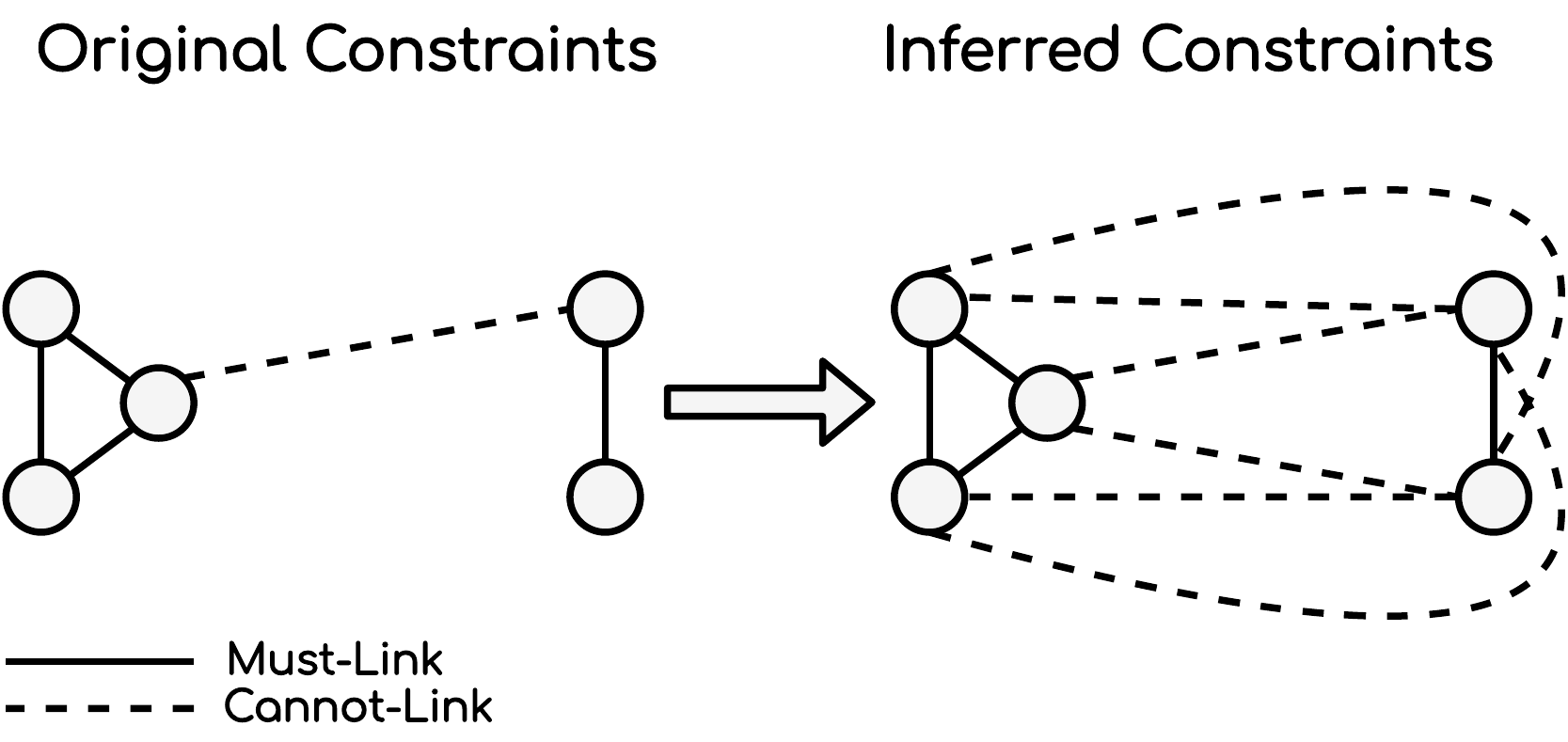}
	\caption{Example of transitive inference of CL constraints.}
	\label{fig:inference_cl}
\end{figure}

\paragraph{\textbf{The Transitive Closure of the constraint set}} It is an augmented set of constraints which can be obtained on the basis of the information contained in the original constraint set, by applying two of its properties which have been introduced in Section~\ref{subsec:pairwise_constraints}, and are formally defined here on the basis of the constraint graph as in Properties~\ref{prprty:tr_inference_ML} and~\ref{prprty:tr_inference_CL}. Graphical examples of these two properties are given in Figure~\ref{fig:inference_ml} and Figure~\ref{fig:inference_cl}, respectively. These two properties can be applied over $CG$ to obtain the transitive closure of the constraint set, which cannot be further augmented without new information.

\begin{property}

	\textbf{Transitive inference of ML}: Let $cc_1$ and $cc_2$ be two connected components in $CG$ with only positive edges in it (only ML constraints). Then, if there is a constraint $C_=(x_i,x_j)$ with $x_i \in cc_1$ and $x_j \in cc_2$, then the new constraints $C_=(a,b)$ can be inferred for all $a \in cc_1$ and $b \in cc_2$~\cite{basu2008constrained}.
	\label{prprty:tr_inference_ML}

\end{property}

\begin{property}

	\textbf{Transitive inference of CL}: Let $cc_1$ and $cc_2$ be two connected components in $CG$ with only positive edges in it (only ML constraints). Then, if there is a constraint $C_{\neq}(x_i,x_j)$ with $x_i \in cc_1$ and $x_j \in cc_2$, then the new constraints $C_{\neq}(a,b)$ can be inferred for all $a \in cc_1$ and $b \in cc_2$~\cite{basu2008constrained}.
	\label{prprty:tr_inference_CL}

\end{property}

\paragraph{\textbf{The Chunklet Graph}} This graph structure can be derived from the definition of the constraint graph and the concept of chunklet. Chunklets are defined in~\cite{bar2003learning,bar2005learning} as ``subsets of points that are known to belong to the same, although unknown, class''. With this definition, it is clear that ML connected components can be compared to chunklets, thus the chunklet graph can be obtained on the basis of the constraint graph. This is done by replacing ML connected components in $CG$ by a single vertex which is adjacent to all former neighbors of the connected components~\cite{cheng2008constrained}. In the case that vertices in $CG$ also store position-related information (as in CC), the position of the new vertex is computed as the average of the nodes in the connected component~\cite{yoshida2010performance}.

\paragraph{\textbf{The Cluster Skeleton}} It is a reduced constraint set which defines the true basic clustering structure of the data. It is obtained by applying the farthest-first scheme to query an oracle about the constraint relating selected instances from the dataset. This is done within an iterative scheme in which membership neighborhoods are created and the farthest instance from all of them is always selected to be queried against at least one instance from every existing neighborhood. If it is constrained to any of those instances by ML, then it is added to that neighborhood and a new ML is created, whereas if it is constrained by CL to all of them, then a new neighborhood is created and related to the other one by CL. The goal of this procedure is to build a constraint set which defines as many disjoint clusters as possible, aiding later CC algorithms determine the number of clusters a feasible partition must have~\cite{basu2004active}. 

\paragraph{\textbf{The Infeasibility}} The concept of infeasibility refers to the number of constraints violated by a given partition. It is one of the most used concepts in CC, as many objective/fitness functions include penalty terms that are directly proportional to the number of violated constraints. Given a partition $C$ (an its associated list of labels $L^C$) and a constraint set $CS$, the infeasibility can be defined as in Equation~\ref{eq:infs}, with $\mathds{1}\llbracket\cdot\rrbracket$ being the indicator function (returns 1 if the input is true and 0 otherwise)~\cite{gonzalez20213shacc}.

\begin{equation}
\text{Infs}(C, CS) = \sum_{C_=(x_i, x_j) \in CS} \mathds{1} \llbracket l_i^C \neq l_j^C \rrbracket + \sum_{C_{\neq}(x_i, x_j) \in CS} \mathds{1} \llbracket l_i^C = l_j^C \rrbracket
\label{eq:infs}
\end{equation}

\paragraph{\textbf{The k-NN Graph}} Also called k-NNG. It is not an exclusive concept from CC. It has been widely used in classic clustering literature and k-NN based classification. However, it is a very useful tool for CC research, therefore many CC approaches are built based on its definition. The k-NNG is a weighted undirected graph in which vertices represent instances from the dataset and every vertex is adjacent to at most $k$ vertices. An edge is created between vertices $u$ and $v$ if and only if instances associated to $u$ and $v$ have each other in their k-nearest neighbors set. The weight $w(v,u)$ for the edge connecting $u$ and $v$ is defined as the number of common neighbors shared by $u$ and $v$: $w(v,u) = |NN(u) \cup NN(v)|$, with $NN(\cdot)$ denoting the set of neighbors of the vertex given as argument~\cite{vu2012improving}.

\section{Statistical Analysis of Experimental Elements} \label{sec:exp_setup}

In this section, a general view on how CC methods are evaluated and compared is presented. Most studies in CC present one or various new methods that need to be evaluated and proved to be competitive with respect to the state-of-the-art at the time they were proposed. In this section, the three experimental elements used to do so are analyzed: the datasets, the validity indices, and the competing methods. Table~\ref{tab:most_frequent_setup} introduces the 15 most frequently used instances of these elements among all the papers analyzed in this study. All statistics presented in this section have been obtained by analyzing 270 studies, which propose a total of 307 methods. Some studies propose more than one method, and some methods are proposed in more than one study, hence the discordance between the number of papers analyzed and the number of proposed methods. Special cases of the experimental elements have not been taken into consideration to obtain the statistics presented in this section. In other words, if a paper uses the Iris dataset for its experiments but removes one of the three classes in the dataset, it is then considered as a single use of the classic Iris dataset, and not listed as a separate dataset. The same can be said for the other two experimental elements. For example, uses of the Pairwise F-measure (PF-measure) are included in the count of the F-measure, and variations on the initialization methods of COP-K-Means are included in the count of the basic COP-K-Means. This is done to obtain more representative and general statistics.

\begin{table}[!h]
	\centering
	\setlength{\tabcolsep}{7pt}
	\renewcommand{\arraystretch}{1.3}
	\resizebox{\textwidth}{!}{
		\begin{tabular}{l c c l c c l c}
		\hline
		\multicolumn{2}{c}{Datasets} && \multicolumn{2}{c}{Competing Methods} && \multicolumn{2}{c}{Validity Indices} \\
		\cline{1-2} \cline{4-5} \cline{7-8}
		Name & No. of Uses && Name & No. of Uses && Name & No. of Uses \\
		\hline
		Iris & 134 && COP-K-Means & 64 && Normalized Mutual Information (NMI) & 89 \\
        Wine & 105 && K-Means & 57 && Clustering Error (CE) & 60 \\
        Ionosphere & 72 && MPCK-Means & 34 && Rand Index (RI) & 55 \\
        Synthetic & 69 && SSKK & 26 && Adjusted Rand Index (ARI) & 48 \\
        Glass & 58 && PCK-Means & 22 && F-measure & 38 \\
        Breast & 51 && KKM & 21 && Time & 25 \\
        Soybean & 47 && FFQS & 17 && Purity & 11 \\
        Balance & 43 && Random & 14 && Unsat & 11 \\
        Sonar & 41 && RCA & 14 && Non Standard (NS) & 10 \\
        Heart & 39 && E$^2$CP & 13 && Jaccard Coefficient (JC) & 8 \\
        Digits & 35 && NCuts & 13 && Visual & 7 \\
        Ecoli & 28 && CSI & 13 && V-measure & 4 \\
        MNIST & 28 && CCSR & 12 && Precision & 4 \\
        Protein & 28 && Constrained EM & 12 && Folkes-Mallows Index (FMI) & 3 \\
        20Newsgroup & 27 && HMRF-K-Means & 11 && Constrained Rand Index (CRI) & 3 \\
		\hline
            
		\end{tabular}}
	\caption{Most frequently used dataset in CC experimental setups.}
	\label{tab:most_frequent_setup}
\end{table}

Sections~\ref{subsec:frec_datasets}, ~\ref{subsec:frec_methods} and~\ref{subsec:frec_VI} dive into the statistics of the frequently used experimental setups regarding datasets, validity indices and competing methods, respectively. Section~\ref{subsec:generation_methods} presents the most used procedure to artificially generate constraints for benchmarking purposes, and Section~\ref{subsec:statistical_tests} gives a quick note on the use of statistical testing to support conclusions in CC literature.

\subsection{Analysis of Datasets} \label{subsec:frec_datasets}

\begin{figure}[!h]
	\centering
	\subfloat[No. of times each dataset in Table~\ref{tab:most_frequent_setup} is used in experiments]{\includegraphics[width=0.5\linewidth]{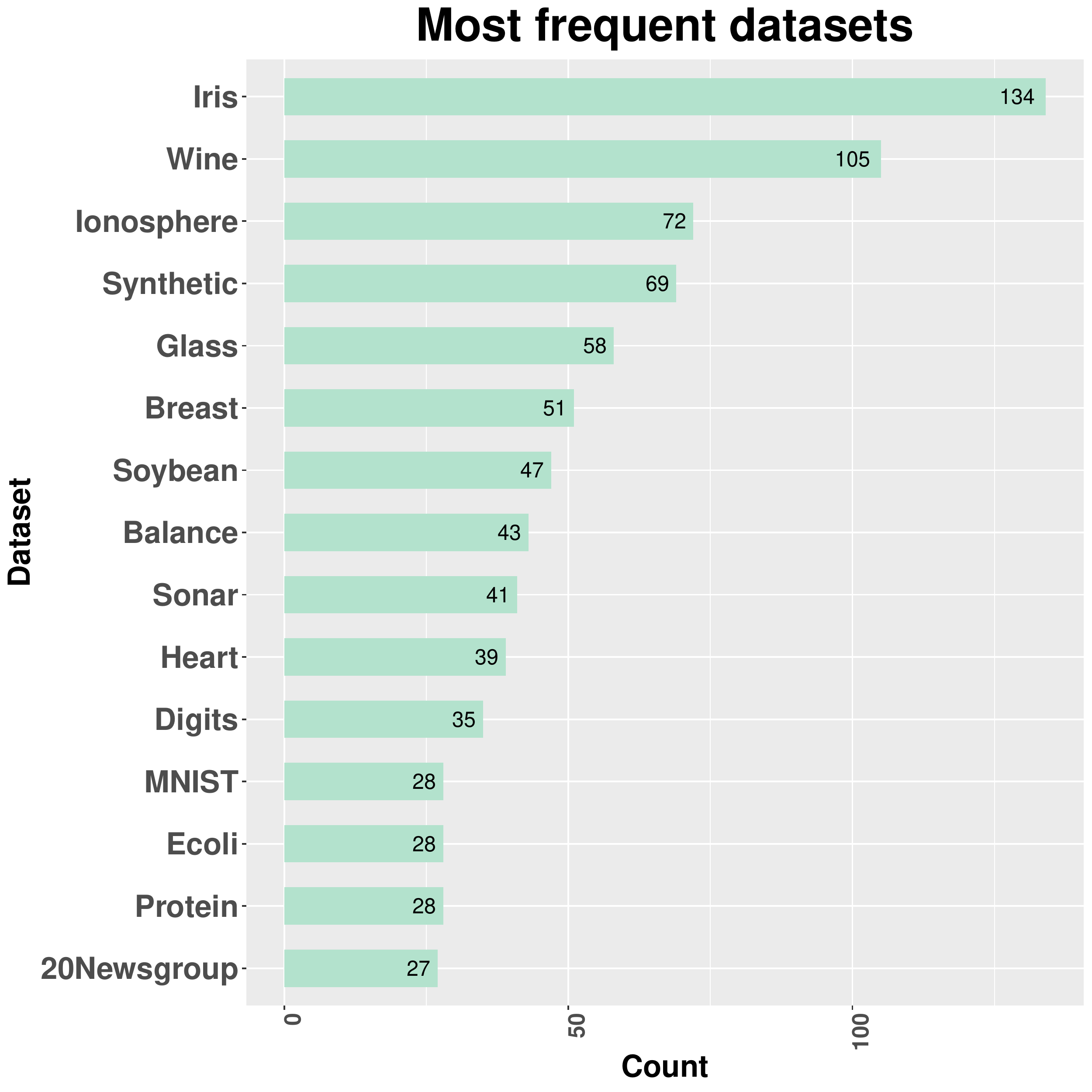}
		\label{fig:datasets_histogram}}
	\subfloat[Distribution of the number of datasets used in experiments]{\includegraphics[width=0.5\linewidth]{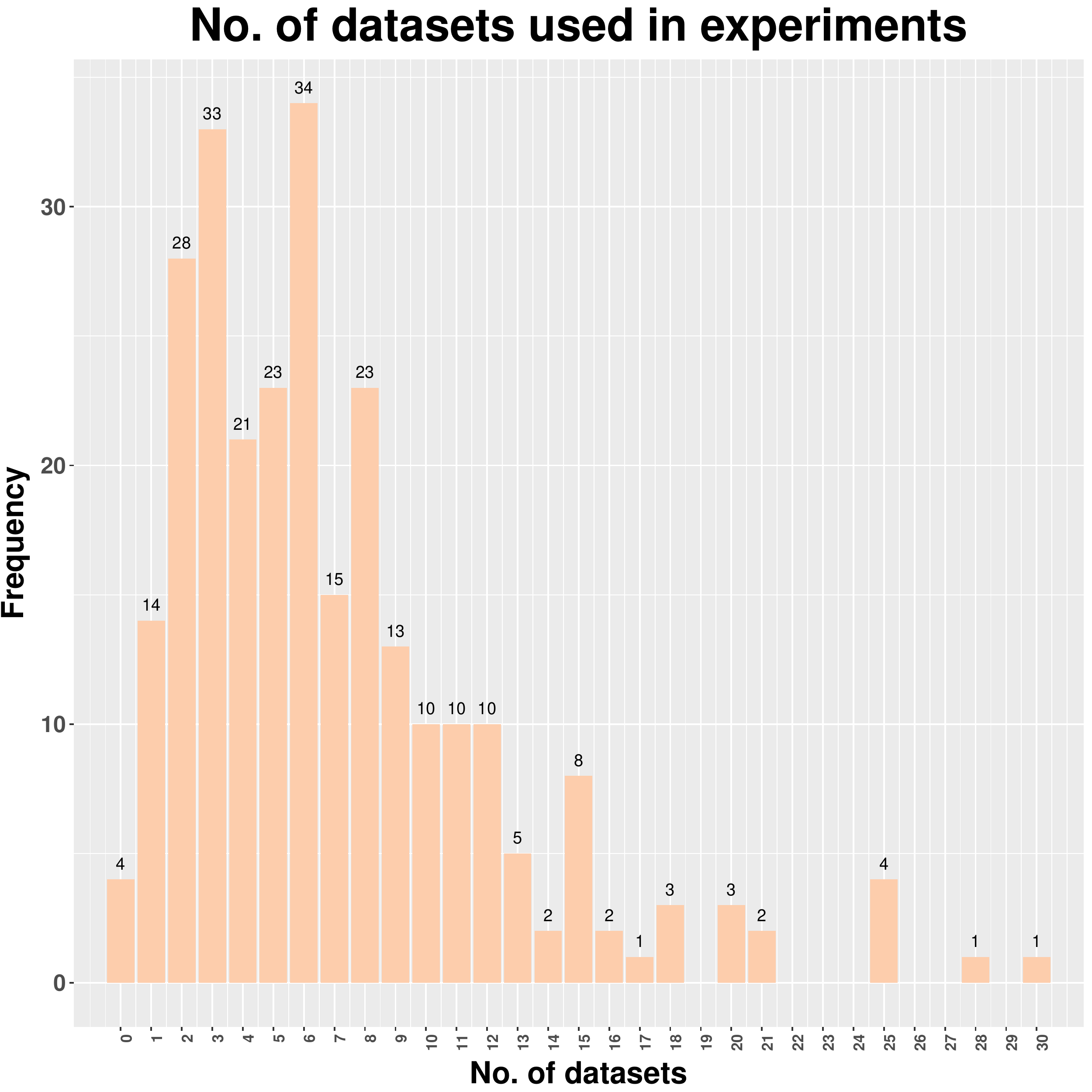}
		\label{fig:datasets_frequencies}}
		
	\vspace{\baselineskip}
	
	\subfloat[Variability of the number of datasets used in every year \\ (the dashed line represents the overall average)]{\includegraphics[width=\linewidth]{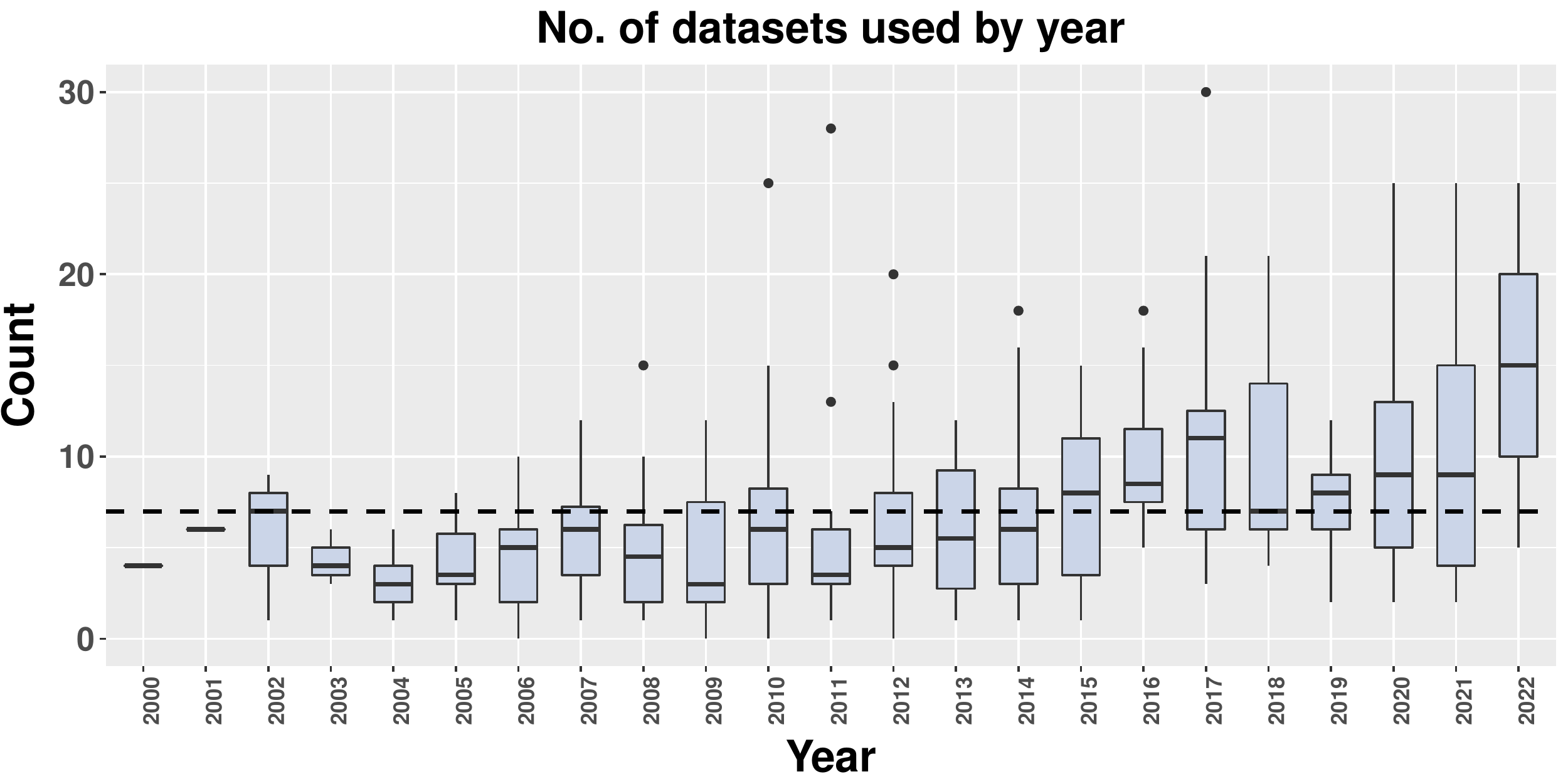}
		\label{fig:datasets_boxplot}}

	\caption{Statistics about the datasets used in the experimental setups of all papers reviewed}
	\label{fig:datasets_statistics}
\end{figure}

A total of 389 different datasets can be identified in the experimental sections of the literature in CC. Figure~\ref{fig:datasets_statistics} displays three different statistical measures about the use of these datasets. Figure~\ref{fig:datasets_histogram} depicts the same information contained in Table~\ref{tab:most_frequent_setup}, presenting it visually for the sake of ease of understanding. Figure~\ref{fig:datasets_frequencies} gives a histogram of the number of datasets used in experiments. Lastly, Figure~\ref{fig:datasets_boxplot} introduces boxplots featuring the variability on the number of datasets used in different years. 

From~\ref{fig:datasets_histogram}, it is clear that classification datasets are used as benchmarks for CC methods. The reason for this lies in the lack of specific benchmarks, as very few have been proposed since the inception of the research topic. Classic classification datasets have to be used in order to generate the constraint sets needed by CC methods (see Section~\ref{subsec:generation_methods}). In these cases, labels are never provided to the CC method, but used as the oracle to generate the constraint sets.

Looking at Figure~\ref{fig:datasets_frequencies}, it can be concluded that the most frequent number of datasets used in experiment is 6, and the study which uses the most datasets, analyzes up to 30 of them. Most papers use between 1 and 9 datasets. Note how some papers do not use any datasets, therefore they don't carry out any experiments to prove the efficacy of their proposal. Figure~\ref{fig:datasets_boxplot} shows a consistent increase over the years in the number of datasets used in experiments, probably due to the general growth in computing power, and to the increasing availability of datasets. It also shows how, except for the first few years, there is no consensus on the number of datasets to be used to demonstrate the capabilities of a new method, as boxplots show high variability within each year.

\subsection{Analysis of Competing Methods} \label{subsec:frec_methods}

\begin{figure}[!h]
	\centering
	\subfloat[No. of times each method in Table~\ref{tab:most_frequent_setup} is used in experiments]{\includegraphics[width=0.5\linewidth]{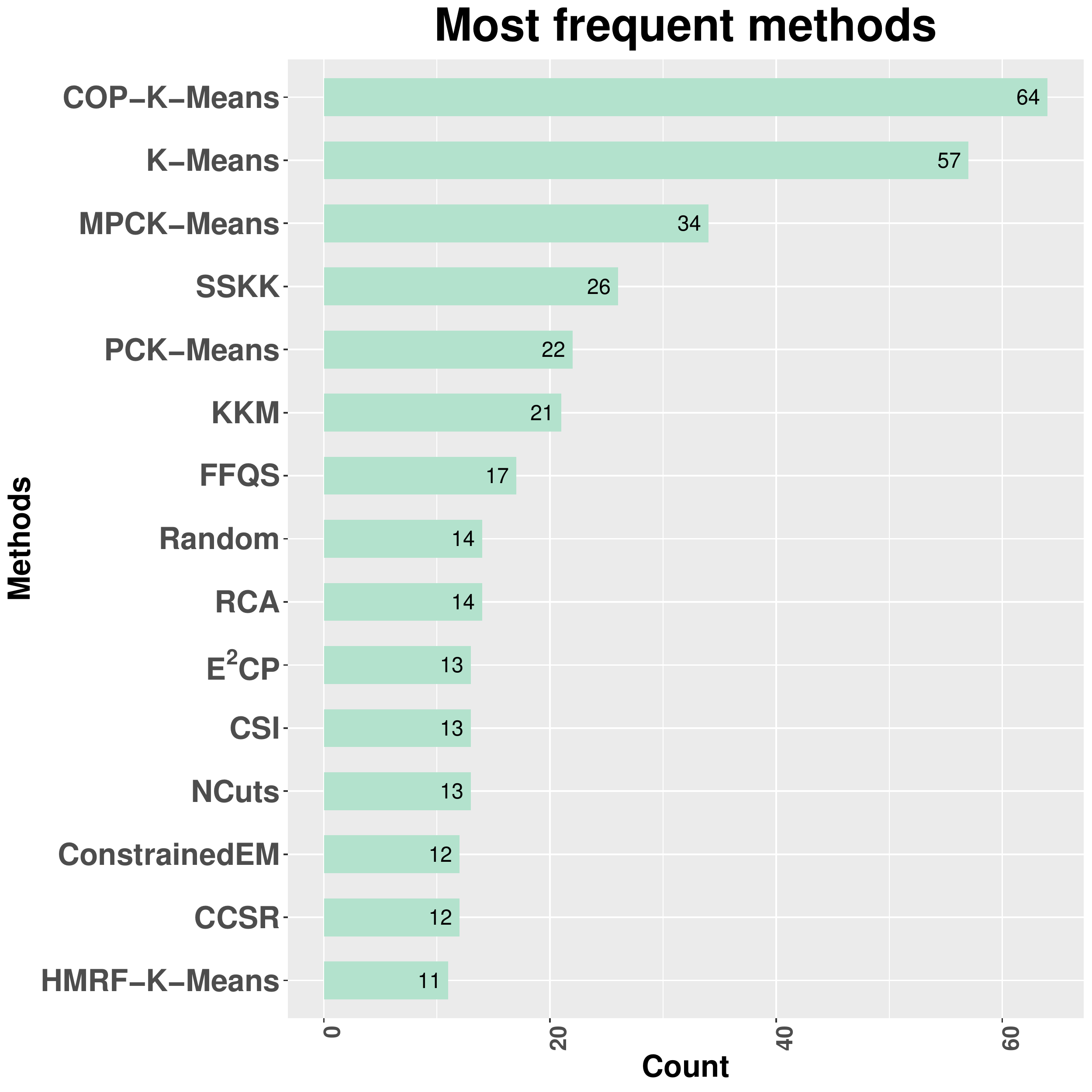}
		\label{fig:methods_histogram}}
	\subfloat[Distribution of the number of methods used in experiments]{\includegraphics[width=0.5\linewidth]{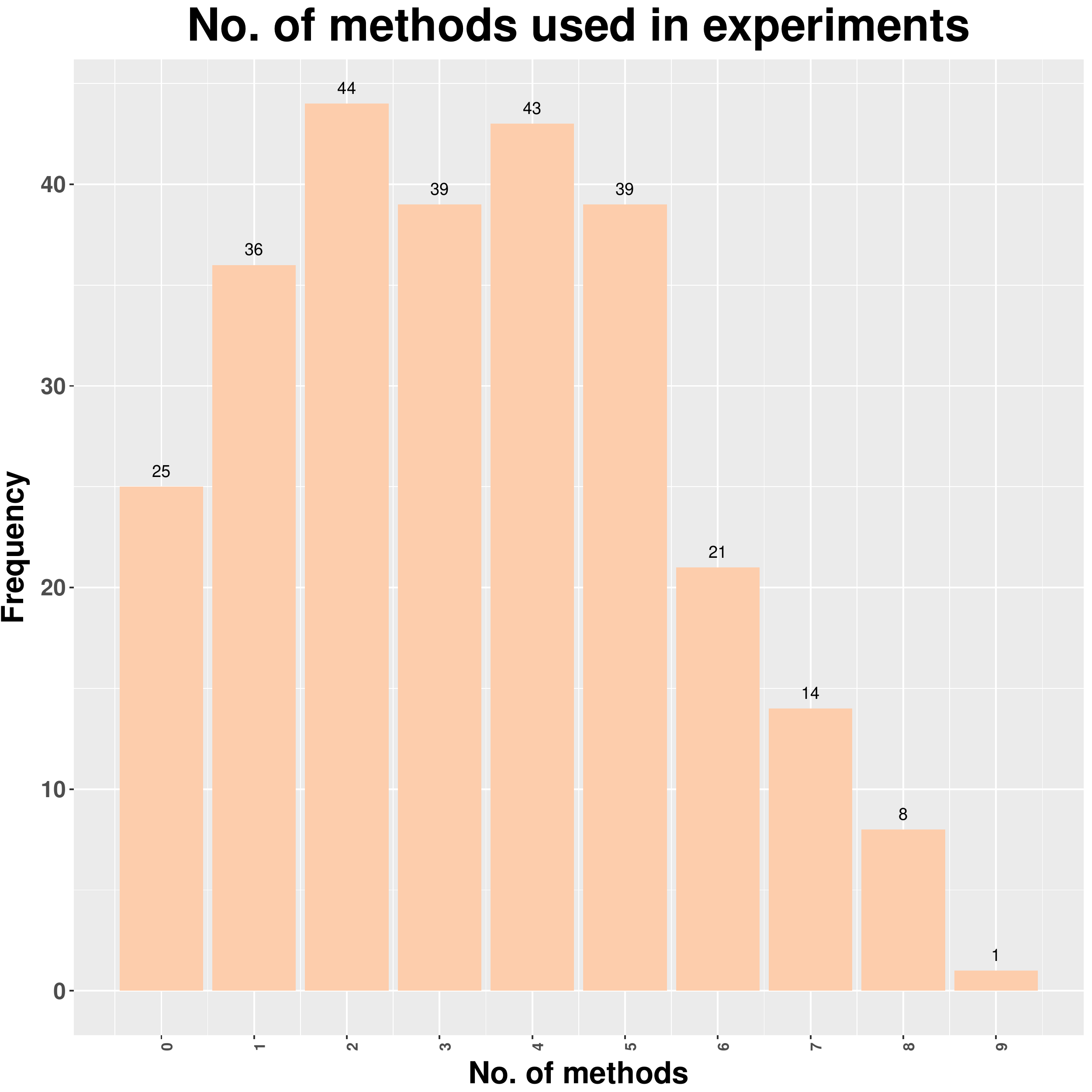}
		\label{fig:methods_frequencies}}
		
	\vspace{\baselineskip}
	
	\subfloat[Variability of the number of methods used in every year \\ (the dashed line represents the overall average)]{\includegraphics[width=\linewidth]{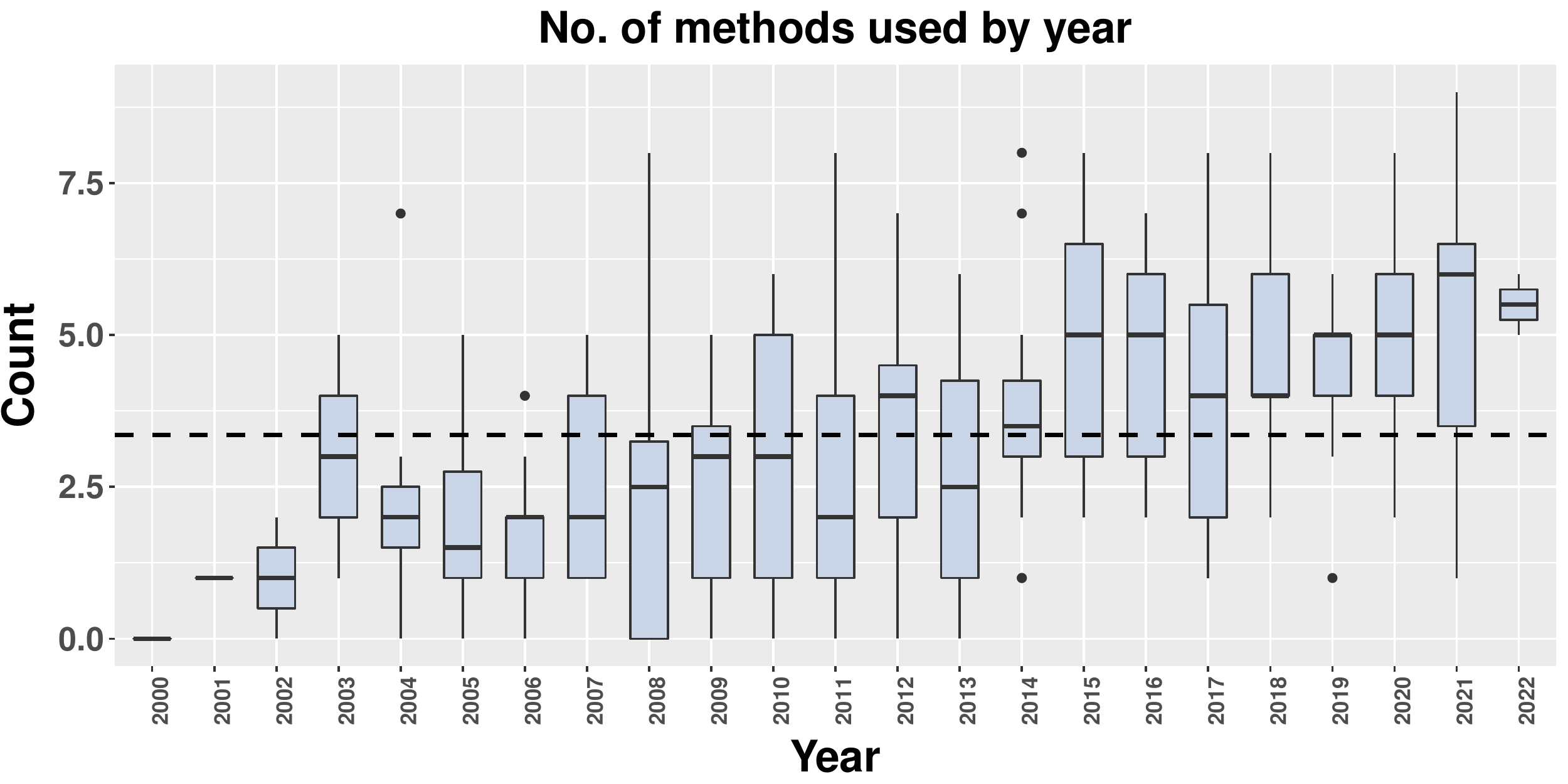}
		\label{fig:methods_boxplot}}

	\caption{Statistics about the methods used in the experimental setups of all papers reviewed}
	\label{fig:methods_statistics}
\end{figure}

Any new CC method has to be proven to be competitive with the methods belonging to the state-of-the-art. Methods belonging to this category change over the years. Nonetheless, Figure~\ref{fig:methods_histogram} presents a set of methods which are used very frequently, and can be subsequently understood as baseline methods. In fact, most of them correspond to the first proposals in different CC categories, e.g.: COP-K-Means is the first CC method ever proposed, PCK-Means is the first penalty-based CC method, KKM is the first constrained spectral clustering method, FFQS is the first active constrained clustering method, etc. Section~\ref{sec:Taxonomy} presents all these methods within the context of their specific CC category. Algorithms such as K-means or NCuts stand out as well, as they are not CC algorithms but classic clustering algorithms. When the experiments carried out aim to prove not only the capabilities of a new CC method, but also the viability of CC itself (as in the first proposals) or the viability of any new constraint generation method, then comparing with classic clustering algorithm is justified.  

Figure~\ref{fig:methods_frequencies} shows the distribution of the amount of methods used in experimental setups in CC literature. The most frequent comparison uses only two methods, which is a very low number taking into account the plethora of methods available to compare with (307 particularly). However, comparisons using between 4 and 6 methods are also reasonably frequent, with said frequency decaying from 6 methods to 9, which are used only in a single study. There is a particular fact that may catch our attention: 25 studies chose not to compare against any previous proposals. Given how well established baselines methods are, this should never be allowed in new CC studies. An increasing tendency can be observed in the number of methods used over the years. Likewise with the number of datasets (see Figure~\ref{fig:datasets_boxplot}). Accordingly, this can be caused by an increase in computing power over the year and by the increase in the number of available methods to compare against.

\begin{figure}[!h]
	\centering
	\subfloat[Proportion of CC methods that are used in later \\ studies to compare with.]{\includegraphics[width=0.5\linewidth]{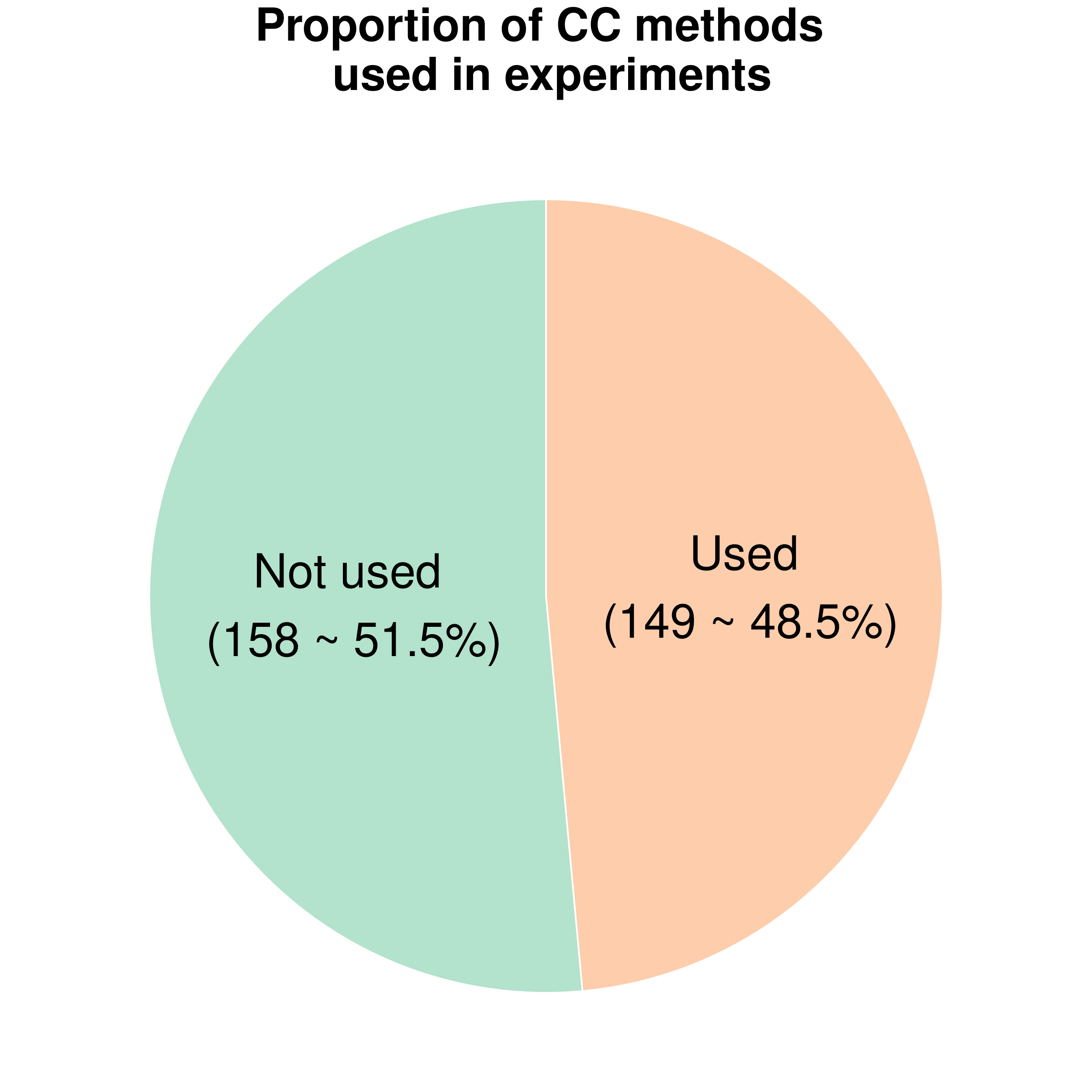}
		\label{fig:pieplot_1}}
	\subfloat[Proportion of methods used in experiments \\ which are CC or other type of clustering methods.]{\includegraphics[width=0.5\linewidth]{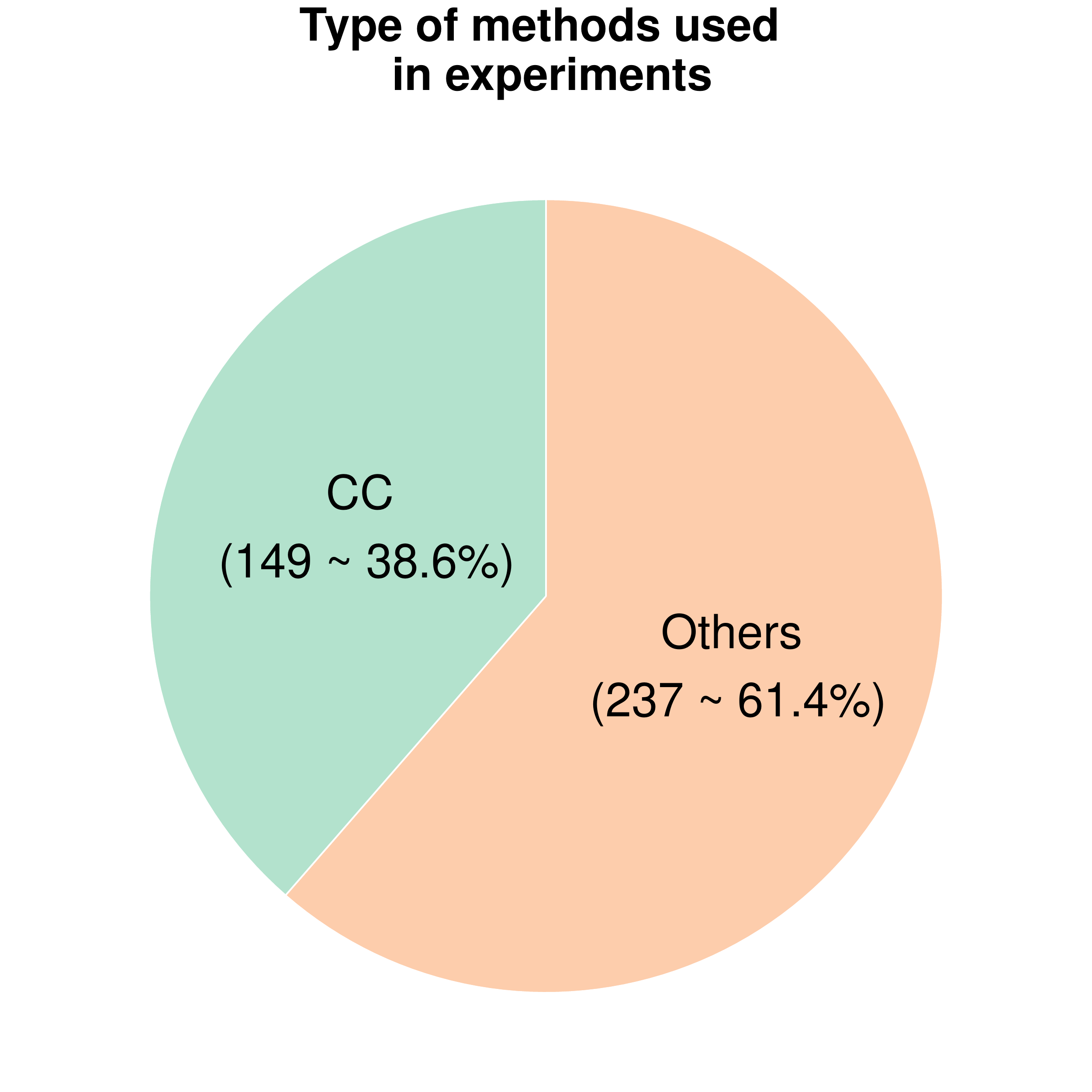}
		\label{fig:pieplot_2}}

	\caption{Piecharts depicting the usability of all CC methods reviewed in experimental setups}
	\label{fig:pieplots_methods}
\end{figure}

Piecharts in Figure~\ref{fig:pieplots_methods} show further statistics about the proportions of methods used in experiments in CC studies. For example, it may be interesting to answer the following question: from all methods used in experiments to compare with, how many of them are CC methods? Figure~\ref{fig:pieplot_2} answers this question. From all methods compared with (386), only 38.8\% (147) are CC methods. The rest of the methods are not necessarily classic clustering methods, they can belong to other fields of SSL or use different types of constraints. This may seem contradictory with respect to what Figure~\ref{fig:datasets_histogram} shows. However, this is not the case. In conjunction, Figures~\ref{fig:pieplots_methods} and~\ref{fig:datasets_histogram} evidence that the most frequently used methods are CC methods, even if the number of different classic clustering methods used to compare against is higher than the number of different CC methods.

Another interesting question is: from all CC methods proposed over the years, how many of them are used to compare with in later studies? Figure~\ref{fig:pieplot_1} provides now the answer to this question. From all CC methods proposed (307) in the reviewed studies, 48.5\% of them (149) are used in the experimental section of other studies. This indicates that more than half of the proposed methods have never been considered to be compared with by other authors. Of course, this statistic does not take into account the number of years any given method has been available to be used, only the absolute number of uses. However, this should not have a great impact in the proportions.

\subsection{Analysis of Validity Indices} \label{subsec:frec_VI}

\begin{figure}[!h]
	\centering
	\subfloat[No. of times each validity index in Table~\ref{tab:most_frequent_setup} \\ is used in experiments]{\includegraphics[width=0.5\linewidth]{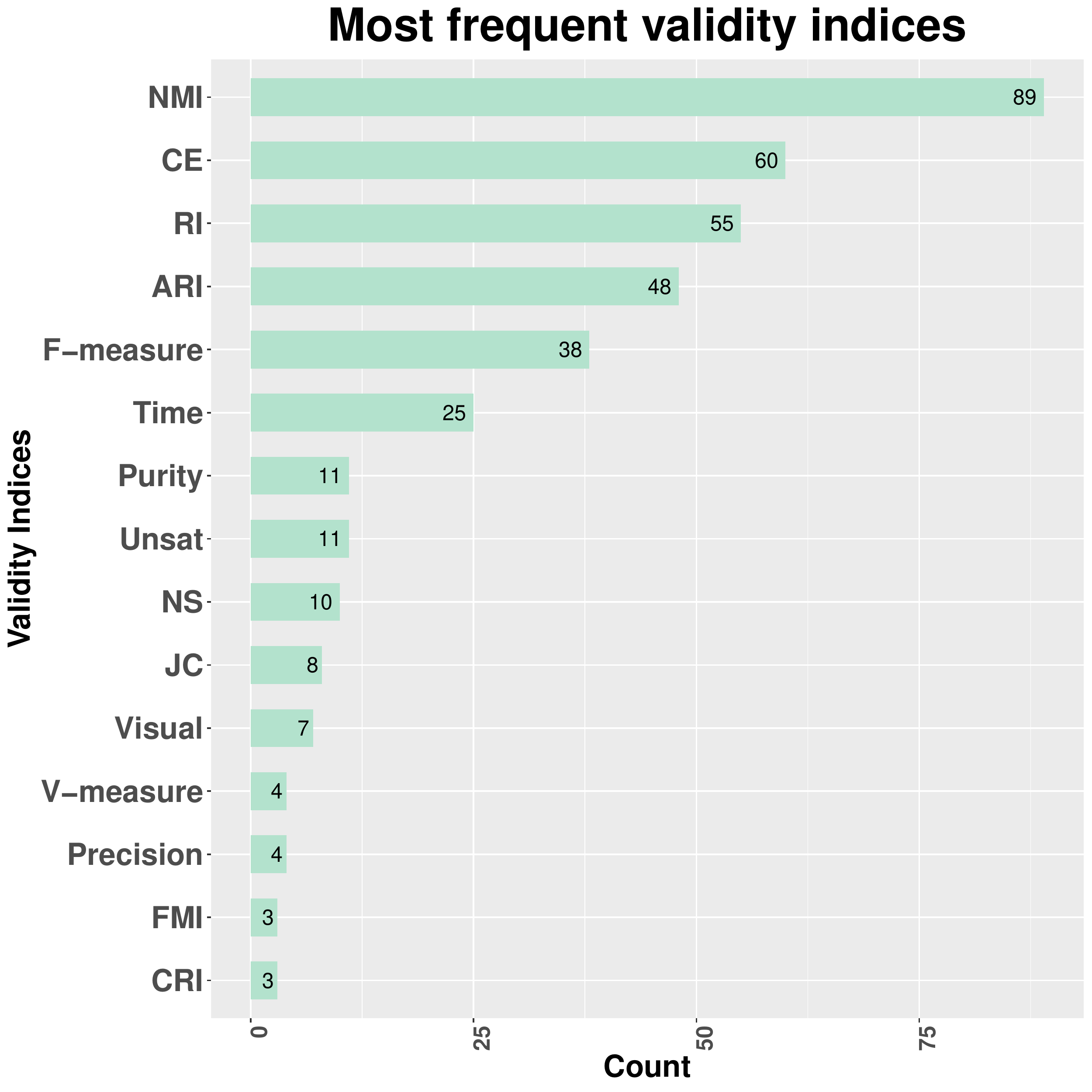}
		\label{fig:indices_histogram}}
	\subfloat[Distribution of the number of validity indices \\ used in experiments]{\includegraphics[width=0.5\linewidth]{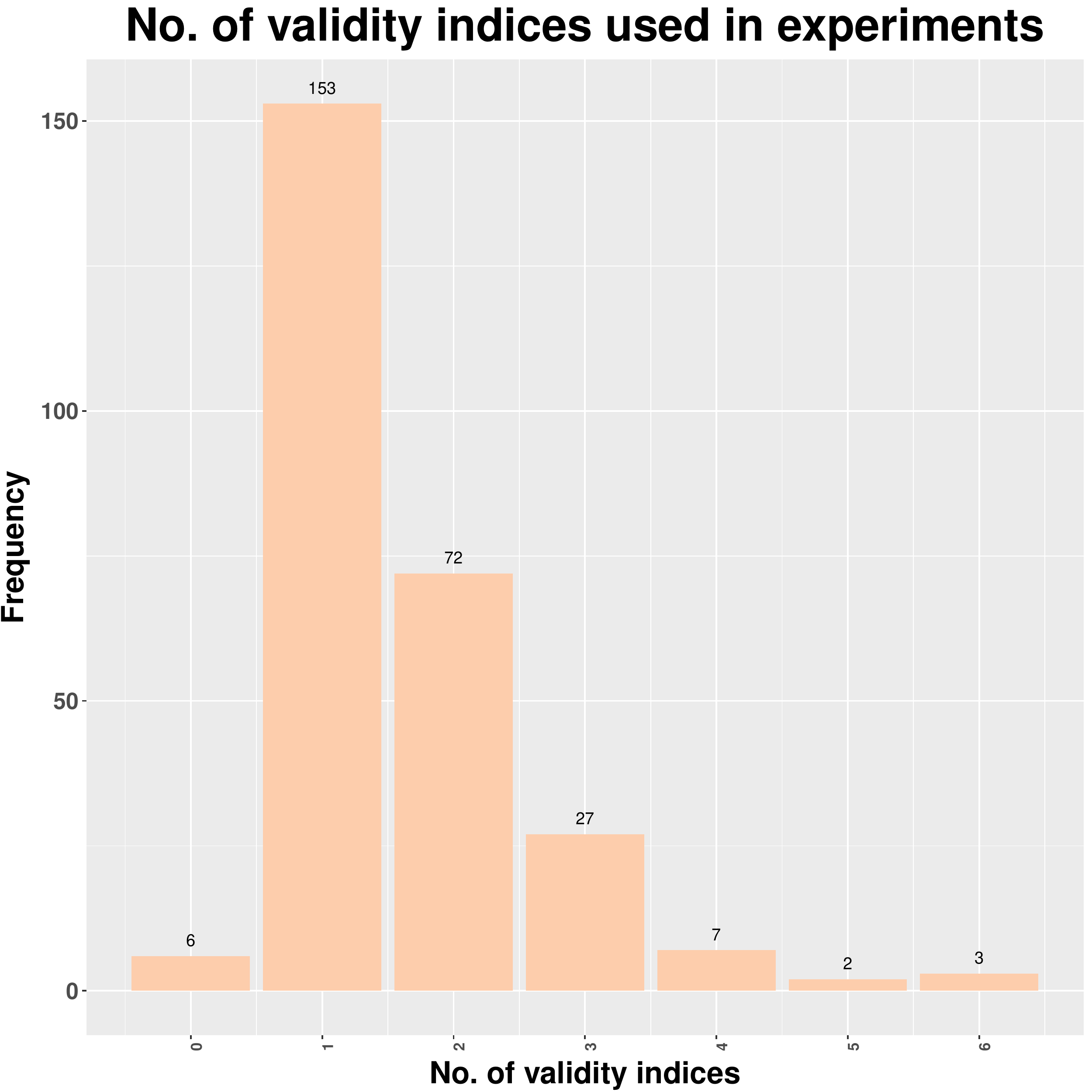}
		\label{fig:indices_frequencies}}
		
	\vspace{\baselineskip}
	
	\subfloat[Variability of the number of validity indices used in every year \\ (the dashed line represents the overall average)]{\includegraphics[width=\linewidth]{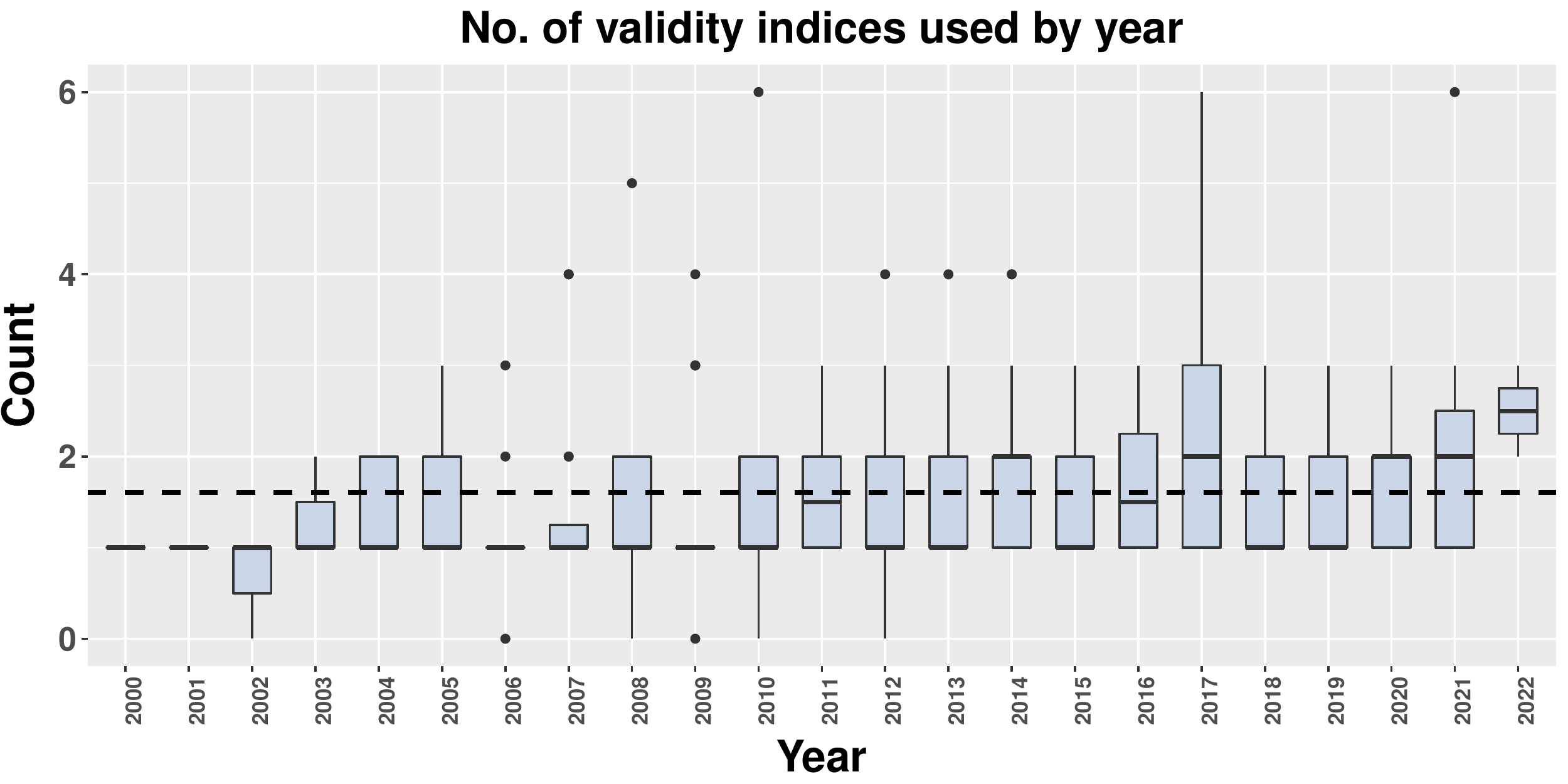}
		\label{fig:indices_boxplot}}

	\caption{Statistics about the validity indices used in the experimental setups of all papers reviewed}
	\label{fig:indices_statistics}
\end{figure}

Validity indices are used to objectively evaluate the performance of a given method independently of the benchmarks it is tested in. This means that the output value of the validity indices is independent from the features of the benchmarks datasets, such as their size of their number of features in the case of classification datasets. The same analysis performed over the datasets (Section~\ref{subsec:frec_datasets}) is performed over the validity indices. Figure~\ref{fig:indices_statistics} shows the statistical summary on the usage of validity indices in CC literature (as it was performed for the datasets). From Figure~\ref{fig:indices_histogram} it is clear that the most used validity index is the Normalized Mutual Information (NMI), followed by the Clustering Error (EC), the Rand Index (RI), the Adjusted Rand Index (ARI) and the F-measure. Time is used to compare methods a total of 25 times, which represents a very low percentage over the total number of comparisons. Note how Visual validation makes it to the top 15, despite not being an objective and reliable comparison method. Non Standard (NS) measures are used in 10 studies, meaning that the used measure is proposed specifically in the same paper for that specific case or that it is never referred to again in CC literature. Among the 15 most used measures, there is only one specifically designed to compare CC methods: the Unsat. Unsat measures the proportion of constraints violated by the output partition of any given method, and therefore can be used to measure scalability with respect to the number of constraints. In this study, authors want to draw two validity indices to the attention of the reader: the Constrained Rand Index (CRI), proposed in~\cite{wagstaff2000clustering}, and the Constrained F-measure (CF-measure), proposed in~\cite{he2014semi}. These two validity indices are versions of RI and F-measure, respectively, corrected by the number of constraints available. They assume that the higher the number of constraints available, the easier it is to score a high value in classic clustering validity indices, therefore they correct (lower) those values with the size of the constraint set. These two measures are used in very few cases, while they are specifically designed to benchmark CC methods. This fact is particularly remarkable in the case of the CRI, as it was proposed along with the first CC study ever in~\cite{wagstaff2000clustering}.

Figure~\ref{fig:indices_frequencies} shows that the most common number of validity indices used in CC literature is 1. Using more than one validity index is a healthy practice in any study, as demonstrating the capabilities of a new method in more than one dimension reinforces positive conclusions about it. With respect to Figure~\ref{fig:indices_boxplot}, the variability observed in the other cases (Figures~\ref{fig:datasets_boxplot} and~\ref{fig:methods_boxplot}) is not present here, as the number of validity indices used is not related to the computation power, and most of them were proposed before the inception of CC.

Some of the validity indices in Table~\ref{tab:most_frequent_setup} do not need to be specifically defined, as it is the case of the Time or the Visual indices, which are self-explanatory. Others are incidentally defined, such as the Precision, which is a by-product of the F-measure. Lastly, no general definition can be given for the Non Standard indices. For the rest of them, both a formal and intuitive definition can be found here. From now on, in this section, $C$ refers to the partition generated by any given CC method, and $C^*$ refers to the ground-truth partition. Please note the influence of the use of classification datasets in the selection of validity indices used to evaluate CC methods. All of the validity indices take two partitions as their input, and produce a measure according to their similarity or dissimilarity. Therefore, these validity indices can be used to evaluate the performance of a clustering algorithm only when one of the partitions given as input is the ground-truth partition, which can be obtained for labeled datasets only.

\paragraph{\textbf{Normalized Mutual Information (NMI)}} The NMI is an external validity index  that estimates the quality of a partition with respect to a given underlying labeling of the data. In other words, NMI measures how closely a clustering algorithm could reconstruct the underlying label distribution. Taking $C$ as the random variable denoting the cluster assignments of instances (the partition), and $C^*$ as the random variable denoting the underlying class labels, the NMI can be formulated in terms of information theory as in Equation~\ref{eq:vi:NMI}~\cite{li2019ascent,basu2004probabilistic,basu2004active}.

\begin{equation}
\text{NMI} = 2 \frac{I(C;{C^*})}{H(C) + H({C^*})},
\label{eq:vi:NMI}
\end{equation}

\noindent where $I(X;Y) = H(X) - H(X|Y)$ is the mutual information between the random variables $X$ and $Y$, $H(X)$ is the Shannon entropy of $X$ and $H(X|Y)$ is the conditional entropy of $X$ given $Y$. For more details on NMI please see~\cite{dom2002information}. The output value range for the NMI is $[0,1]$, with high values indicating a high level of similarity between the two partitions, and a low value indicating a low level of similarity.

\paragraph{\textbf{The Clustering Error (CE)}} The CE is the negative, unsupervised version of the classic classification accuracy. It measures the proportion of correctly clustered instances by best matching the cluster labels to the ground-truth labels. Given the permutation mapping function $\text{map}(\cdot)$ over the cluster labels, the CE with respect to $\text{map}(\cdot)$ can be computed as in Equation~\ref{eq:vi:CE}~\cite{hu2014exploiting,li2009constrained,li2009constrained2}. The best mapping function that permutes clustering labels to match the ground truth labels can be computed by the Kuhn-Munkres algorithm (the Hungarian method)~\cite{hu2014exploiting,kuhn1955hungarian}. Please note that the CE validity index is sometimes used in its positive form, which is the clustering accuracy. It can be computed by just changing the condition in the indicator function $\mathds{1} \llbracket \cdot \rrbracket$ to be negative (replace $=$ by $\neq$). The output value range for the CE is $[0,1]$, with high values indicating a low level of accuracy, and a low value indicating a high level of accuracy.

\begin{equation}
\text{CE} = 1 - \frac{1}{n} \sum_{i=1}^n \mathds{1} \llbracket \text{map}(l_i^C) = l_i^{C^*} \rrbracket
\label{eq:vi:CE}
\end{equation}
\\
\paragraph{\textbf{The Rand Index (RI)}} The RI measures the degree of agreement between two partitions. It can be used to measure the quality of a partition obtained by any CC algorithm by giving the ground-truth partition as one of them. Therefore, the two compared partitions are $C$ and $C^*$. The RI views $C$ and $C^*$ as collections of $n(n - 1)/2$ pairwise decisions. For each $x_i$ and $x_j$ in $X$, they are assigned to the same cluster or to different clusters by a partition. The number of pairings where $x_i$ is in the same cluster as $x_j$ in both $C$ and $C^*$ is taken as $a$; conversely, $b$ represents the number of pairings where $x_i$ and $x_j$ are in different clusters. The degree of similarity between $C$ and $C^*$ is computed as in Equation~\ref{eq:vi:RI}~\cite{rand1971objective}, where $n$ is the number of instances in $X$. The output value range for the RI is $[0,1]$, with high values indicating a high level of agreement between the two partitions, and a low value indicating a low level of agreement.

\begin{equation}
\text{RI} = \frac{a + b}{n(n - 1)/2}
\label{eq:vi:RI}
\end{equation}
\\
The RI can be conveniently formulated in terms of the elements of a confusion matrix as well~\cite{zhong2019active}. Equation~\ref{eq:vi:tp_notation} defines these elements in terms of cluster memberships in a partition, which can be referred to as: True Positives (TP), False Positives (FP), True Negatives (TN), and False Negatives (FN). Equation~\ref{eq:vi:RI_TP} makes use of these elements to give a new definition for the RI.

\begin{equation}
\begin{array}{ll}
\text{TP} & = \{(x_i, x_j) | l_i^{C^*} = l_j^{C^*}, \; l_i^C = l_j^C, \; i \neq j \} \\
\text{FP} & = \{(x_i, x_j) | l_i^{C^*} = l_j^{C^*}, \; l_i^C \neq l_j^C, \; i \neq j \} \\
\text{TN} & = \{(x_i, x_j) | l_i^{C^*} \neq l_j^{C^*}, \; l_i^C \neq l_j^C, \; i \neq j \} \\
\text{FN} & = \{(x_i, x_j) | l_i^{C^*} \neq l_j^{C^*}, \; l_i^C = l_j^C, \; i \neq j \} \\
\end{array}.
\label{eq:vi:tp_notation}
\end{equation}
\\
\begin{equation}
\text{RI} = \frac{|\text{TP}|+|\text{TN}|}{|\text{TP}|+|\text{FP}|+|\text{TN}|+|\text{FN}|}
\label{eq:vi:RI_TP}
\end{equation}

\paragraph{\textbf{The Adjusted Rand Index (ARI)}} The ARI is the corrected-for-chance version of the RI. This correction is done by taking into account the expected similarity of all comparisons between partitions specified by the random model to establish a baseline. This modifies the output value range of the original RI, transforming it into $[-1,1]$ and slightly changing its interpretation. In ARI, a high output value still means a high level of agreement between the two partitions, and a low value means a low level of agreement. However, a value lower than $0$ means that the results obtained are worse than those expected from the average random model. Equation~\ref{eq:vi:ARI} gives the formalization for the ARI~\cite{hubert1985comparing}.

\begin{equation}
\text{ARI} = \frac{\text{RI} - \text{Expected Index}}{\text{Maximum Index} - \text{Expected Index}},
\label{eq:vi:ARI}
\end{equation}

\noindent where Expected Index is the degree of similarity with a random model, Maximum Index is assumed to be 1, and RI is the RI value computed for partitions $C$ and $C^*$.

\paragraph{\textbf{Pairwise F-measure (PF-measure)}} The PF-measure is defined as the harmonic mean of pairwise precision and recall, which are classic validity indices adapted to evaluate pairs of instances. For every pair of instances, the decision to cluster this pair into the same or different clusters is considered to be correct if it matches with the underlying class labeling. In other words, the PF-measure gives the matching degree between the obtained partition $C$ and the ground-truth class labels ${C^*}$. It can be formalized as in Equation~\ref{eq:vi:PFM}~\cite{basu2004active,li2019ascent}, where Precision and Recall are defined as in Equation~\ref{eq:vi:prec_recall}, following the notation introduced in Equation~\ref{eq:vi:tp_notation}~\cite{zhong2019active}. For more details on the PF-measure please see~\cite{hripcsak2005agreement}. The output value range for the PF-measure is $[0,1]$, with high values indicating a high level of agreement between the two partitions, and a low value indicating a low level of agreement.

\begin{equation}
\text{Precision} = \frac{|\text{TP}|}{|\text{TP}| + |\text{FP}|}, \;\;\; \text{Recall} = \frac{|\text{TP}|}{|\text{TP}| + |\text{FN}|}.
\label{eq:vi:prec_recall}
\end{equation}

\begin{equation}
\text{PF-measure} = 2 \frac{\text{Precision} \times \text{Recall}}{\text{Precision} + \text{Recall}} = \frac{2 |\text{TP}|}{2|\text{TP}| + |\text{FP}| + |\text{FN}|}.
\label{eq:vi:PFM}
\end{equation}
\\
\paragraph{\textbf{The Constrained Pairwise F-measure (CPF-measure)}} The CPF-measure is a version of the classic PF-measures that takes constraints into account. It does so by including the number of ML constraints in the computation of the Precision and Recall terms as in Equation~\ref{eq:vi:prec_recall_prime}. This way, the number of correctly clustered instances is penalized by the number of ML constraints. Subsequently, the higher the number of ML constraints available to perform clustering, the less credit the term TP is given. The final CPF-measure can be computed as in Equation~\ref{eq:vi:CPFM}. The output value range for the CPF-measure is $[0,1]$, and the value is interpreted as in the PF-measure.

\begin{equation}
\text{Precision}' = \frac{|\text{TP}| - |C_=|}{|\text{TP}| + |\text{FP}| - |C_=|}, \;\;\; \text{Recall}' = \frac{|\text{TP}| - |C_=|}{|\text{TP}| + |\text{FN}| - |C_=|}.
\label{eq:vi:prec_recall_prime}
\end{equation}

\begin{equation}
\text{CPF-measure} = 2 \frac{\text{Precision}' \times \text{Recall}'}{\text{Precision}' + \text{Recall}'}.
\label{eq:vi:CPFM}
\end{equation}

\paragraph{\textbf{The Purity}} This is a classic validity index used to evaluate the performance of clustering methods. It measures the homogeneity of the generated partition, i.e.: the extent to which clusters contain a single class~\cite{zhao2005mixture,halkidi2012semi,xu2011constrained}. It can be computed by determining the most common class of each cluster $c_i$ (with respect to the true labels ${C^*}$), which can be done my determining the greatest intersection with respect to the ground-truth partition. The sum of all intersection is then divided by the total number of instances $n$ in the partition $C$ to obtain the Purity value of said partition. Equation~\ref{eq:vi:purity} formalizes this concept. The output value range for the Purity is $[0,1]$, with high values indicating high level of resemblance between the two partitions, and a low value indicating a low level of resemblance.

\begin{equation}
\text{Purity} = \frac{1}{n} \sum_{c_i \in C} \max_{c^*_i \in {C^*}}|c_i \cap c^*_i|
\label{eq:vi:purity}
\end{equation}

\paragraph{\textbf{The Unsat}} The Unsat measures the ability of any given CC method to produce partitions satisfying as many constraints as possible. It is computed as the ratio of satisfied constraints as in Equation~\ref{eq:vi:unsat}~\cite{davidson2005clustering,gonzalez2020dils}. It produces a value in the range $[0,1]$, with a high value indicating a high number of violated constraint, and a low value indicating the contrary.

\begin{equation}
\text{Unsat} = \frac{\text{Infs}(C, CS)}{|CS|}.
\label{eq:vi:unsat}
\end{equation}

\paragraph{\textbf{The Jaccard Index (JC)}} The JC  measures similarity between finite sample sets. It is defined as the size of the intersection divided by the size of the union of the sample sets. However, this definition is inconvenient when JC is applied to measure the quality of a partition. Subsequently, a more useful definition can be given in terms of Equation~\ref{eq:vi:tp_notation} as in Equation~\ref{eq:vi:jc}~\cite{bae2006coala,meilua2003comparing,halkidi2001clustering}. Please note that a high value of the CJ in the range $[0,1]$ indicates high dissimilarity between the two compared partitions, while a low value indicates high similarity.

\begin{equation}
\text{JC} = \frac{|\text{TP}|}{|\text{TP}|+|\text{FP}|+|\text{FN}|}
\label{eq:vi:jc}
\end{equation}

\paragraph{\textbf{The V-measure}}. This measure is closely related to the NMI, as it can be viewed as a version of it that computes the normalization of the denominator in Equation~\ref{eq:vi:NMI} with an arithmetic mean instead of a geometric mean. The V-measure is defined as the harmonic mean of Homogeneity and Completeness, which evaluate a partition in a complementary way~\cite{rosenberg2007v,vlachos2008dirichlet}. Homogeneity measures the degree to which each cluster contains instances from a single class of ${C^*}$. This value can be computed as in Equation~\ref{eq:vi:homogeneity}, where $H(X|Y)$ is the conditional entropy of the class distribution of partition $X$ with respect to partition $Y$, and $H(X)$ is the Shannon entropy of $X$. Following the same notation, the Completeness can be defined as in Equation~\ref{eq:vi:completeness}. This can be intuitively interpreted as the degree to which each class is contained in a single cluster. Subsequently, the V-measure is computed as in~\ref{eq:vi:VM}~\cite{rosenberg2007v}. Please note that another aspect to which the V-measure and the NMI are closely related is that the mutual information between two random variables $I(X;Y)$ can always be expressed in terms of the conditional distribution of said variables $H(X|Y)$ as follows: $I(X;Y) = H(X) - H(X|Y)$. The output value range for the V-measure is $[0,1]$, with high values indicating a high level of similarity between the two partitions, and a low value indicating a low level of similarity.

\begin{equation}
\text{Homogeneity} = 1 - \frac{H({C^*}|C)}{H({C^*})}.
\label{eq:vi:homogeneity}
\end{equation}

\begin{equation}
\text{Completeness} = 1 - \frac{H(C|{C^*})}{H(C)}.
\label{eq:vi:completeness}
\end{equation}

\begin{equation}
\text{V-measure} = 2 \frac{\text{Homogeneity} \times \text{Completeness}}{\text{Homogeneity} + \text{Completeness}}.
\label{eq:vi:VM}
\end{equation}

\paragraph{\textbf{The Folkes-Mallows Index (FMI)}} The FMI is another classic external validity index used to measure the similarity between two partitions. It is defined as the geometric mean of the Precision and the Recall~\cite{halkidi2001clustering}. It can be formulated as in Equation~\ref{eq:vi:fmi}. The output value range for the FMI is $[0,1]$, with high values indicating a high level of agreement between the two partitions, and a low value indicating a low level of agreement.

\begin{equation}
\text{FMI} = \sqrt{\frac{|\text{TP}|}{|\text{TP}|+|\text{FP}|}\times \frac{|\text{TP}|}{|\text{TP}|+|\text{FN}|}} = \sqrt{\text{Precision} \times \text{Recall}}
\label{eq:vi:fmi}
\end{equation}

\paragraph{\textbf{The Constrained Rand Index (CRI)}} The CRI is a revised version of the RI which includes constraints specifically in its definition. It introduces the concept of free decisions, which are defined as decisions not influenced by constraints. The CRI subtracts the number of available constraints from the numerator and the denominator of the classic RI~\cite{wagstaff2000clustering,ding2018semi}. As a result, it only evaluates the performance of the CC methods in the free decisions. Equation~\ref{eq:vi:CRI} formalizes CRI, following the same notation as Equation~\ref{eq:vi:RI} (RI). Its results are interpreted as those of RI, but taking into account that the difficulty to obtain values close to 1 increases with the size of the constraint set $|CS|$.

\begin{equation}
\text{CRI} = \frac{a + b - |CS|}{n(n - 1)/2 - |CS|}
\label{eq:vi:CRI}
\end{equation}

\subsection{Constraint Generation Methods} \label{subsec:generation_methods}

The most frequently used procedure to generate constraints is the one proposed in~\cite{wagstaff2001constrained}. It is a simple yet effective method to generate a set of constraints based on a set of labels, hence the generalized use of classification datasets as benchmarks in CC literature. It consists of randomly choosing pairs of instances and setting a constraint between them depending on whether their labels are the same (ML constraint) or different (CL constraint). 

The way pairs of instances are chosen from the dataset may differ from one study to another. However, two common trends are observed. One first decides the percentage of labeled data the oracle has access to and then generates the complete constraints graph based on those labels. The other one first decides the size of the constraint set, and then extracts random pairs of instances from the complete dataset. On the one hand, the first method is more realistic in the sense that it has limited access to labeled data, although it may bias the solution towards poor local optima if the selected labeled instances are not representative enough of the whole dataset. On the other hand, the second constraint generation method has virtual access to the complete set of labels, as pairs of instances are randomly chosen, and the constraint set may end up involving all instances in the dataset in at least one constraint, which might not be a realistic scenario. Nevertheless, it is less likely to bias the solution towards local optima.

There is no consensus on how many constraints need to be generated in order to evaluate the capabilities of a given CC method. However, some general guidelines can be given. Based on Observation~\ref{obv:adverse_effects}, it is clear that proper empiric evaluation of CC methods must include an averaging process on the results obtained for different constraint sets, in order to reduce the effects of specific adverse constraint sets.

\begin{observation}
    
    \textbf{Specific constraint sets can have adverse effects.} Even if constraint sets are generated on the basis of the true labels, some constraint sets may decrease accuracy when predicting those very labels~\cite{davidson2007survey}. \label{obv:adverse_effects}

\end{observation}

Given Observation~\ref{obv:scalability}, testing CC methods should include different levels of constraint-based information. This must be done in order to study the scaling capabilities of the proposed method. If a method does not scale the quality of the solutions with the size of the constraint set, any improvement over the solutions obtained with an empty constraint set may be due to random effects.

\begin{observation}
    
    \textbf{The accuracy of the predictions scales with the amount of constraint-based information.} The quality of the solution should scale with the size of the constraint set: the more constraint are available, the better the results obtained are~\cite{davidson2007survey}. \label{obv:scalability}

\end{observation}

\subsection{On the Use of Statistical Tests} \label{subsec:statistical_tests}

\begin{figure}[!ht]
	\centering
	\includegraphics[width=0.7\linewidth]{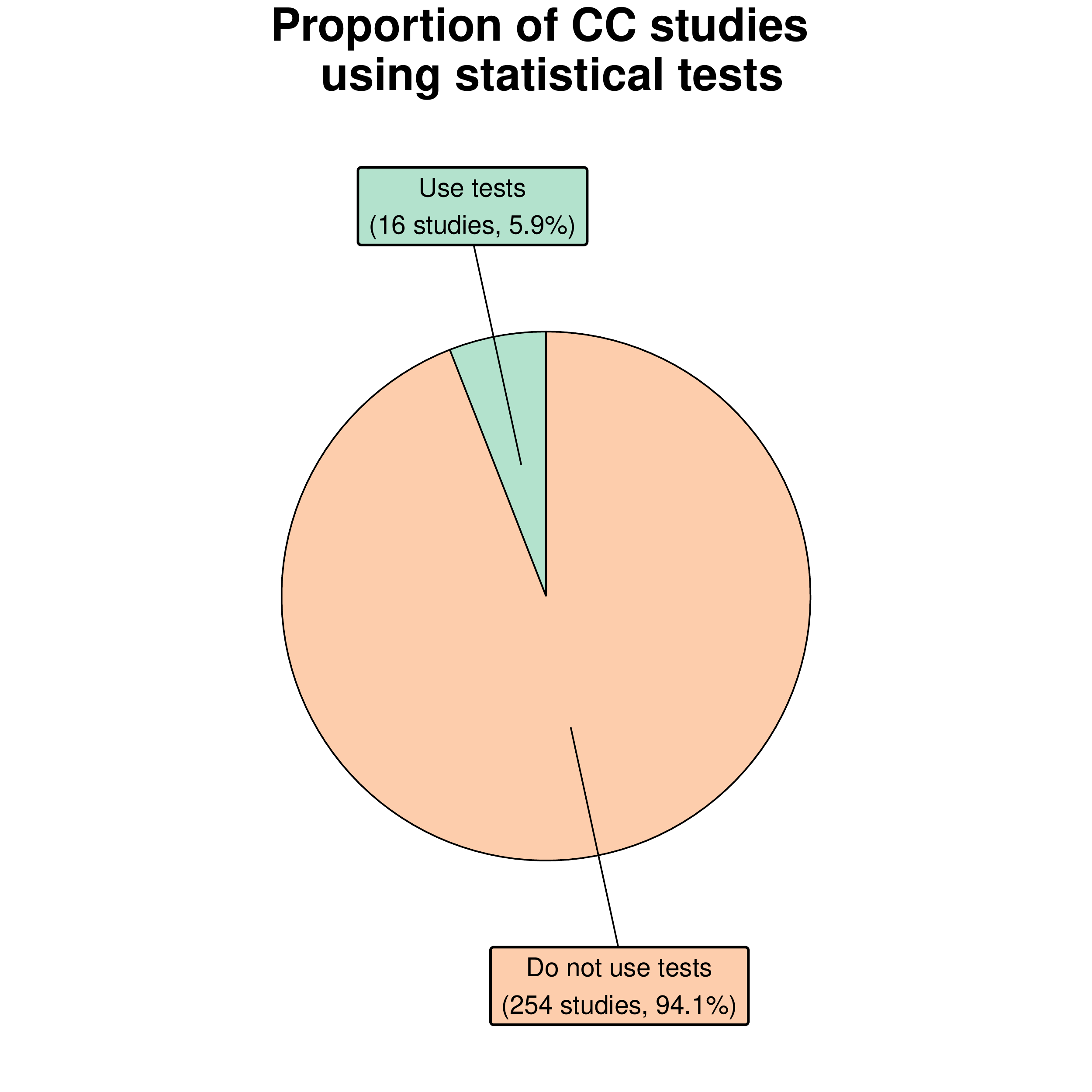}
	\caption{Piechart featuring the proportion of CC studies which use statistical tests.}
	\label{fig:pieplot_tests}
\end{figure}

Statistical testing is a settled practice in Computer Science. It provides objective evidence of the results of a study, supporting its conclusions, either if the used tests are Null Hypothesis Statistical Tests (NHST)~\citep{derrac2011practical} or the more recent Bayesian Tests~\cite{benavoli2017time}. However, this does not seem to be the case in the CC area. As shown in Figure~\ref{fig:pieplot_tests}, only 5,6\% of the studies (16 out of 270) analyzed in this review use statistical testing to support their conclusions. Authors consider this to be one of the major criticisms of the area of CC. Studies supporting their conclusions on mere average results values for any validity index/indices (as it is the case for most of them) should be encouraged to use statistical testing to further objectively prove their hypotheses.

\section{Scoring System} \label{sec:scoring_system}

The aim of this study is not only to give a taxonomy of constrained clustering methods, but also to provide researchers with tools to decide which methods to use. This section proposes an scoring system that is designed to indicate the potential and popularity of every reviewed method. This system assigns a numerical value to every CC method, which will be later used to rank all 307 of them. This value can be interpreted as a measure for the quality of the method. Three semantically different aspects of every method $\mathcal{A}$ are analyzed to decide its score: the quality of the experimental setup they are tested in ($EQ_{\mathcal{A}}$), the confidence in the results obtained in the experiments ($VQ_{\mathcal{A}}$), and the influence of the method in later studies ($I_{\mathcal{A}}$). As for the formalization of these concepts, it is necessary to define the basic quantifiable elements that can be obtained from a method, which are shown in Table~\ref{tab:method_functions}. All of them are lists that contain a value or a set of values associated to every method. Therefore, the length of these lists is always equal to the number of methods reviewed in this study. These lists can be accessed in a more precise way, by method or by year. For example: $D_{\mathcal{A}}$ is a single value which refers to the number of datasets used to test method $\mathcal{A}$, and $D^{Y}$ is a list of values referring to the number of datasets used to test methods from year $Y$. Note that $M_{\mathcal{A}}$ is the list of methods used to compare $\mathcal{A}$, therefore $M^{Y}$ is a list of lists.

\begin{table}[!h]
	\centering
	\setlength{\tabcolsep}{7pt}
	\renewcommand{\arraystretch}{1.8}
		\begin{tabular}{c l}
			\hline
			Function & Meaning\\
			\hline
			$Y$ & The list of publication years for all methods. \\
			
			$M$ & The lists of sets of methods used in comparisons. \\
			
			$C$ & The list of number of times a method is used to be compared with in later studies.\\
			
			$T$ & {The list of indicators for the use of statistical tests for every method}\\
			
			$D$ & The list of number of datasets used to test every method. \\
			
			$V$ & The list of number of validity indices used to evaluate every method. \\
		
			\hline

		\end{tabular}
	\caption{Functions to get basic features of methods.}
	\label{tab:method_functions}
\end{table}

In this study, authors have decided to evaluate each method within its time context, i.e. the year of publication of the method is taken into account to compute its score. This is done to remove the computational capacity component from the scoring system, as the number of datasets or the number of methods used to test new proposal is highly dependent of said parameter (see Figure~\ref{fig:datasets_statistics} and Figure~\ref{fig:methods_statistics}). Moreover, publication requirements and standards change over the years, and tend to become more rigid. Not taking the year of publication into account would greatly benefit recent methods, as their studies have to meet harder publication requirements which are usually related to their novelty and their experimental quality.

\subsection{Scoring of the Experimental Quality}

The quality of the experimental setup $EQ$ used to test a method $\mathcal{A}$ can be computed with information that is fully contained in the study which proposes it. Two of the experimental elements introduced in Section~\ref{sec:exp_setup} take part in this procedure: the number of datasets used to test the scoring method $\mathcal{A}$ (in list $D$), and the methods that are used to compare it (in list $M$). Equation~\ref{eq:scr:exp_quiality} gathers these two basic measures and gives the expression to compute the experimental quality of a scoring method $EQ_{\mathcal{A}}$.

\begin{equation}
EQ_{\mathcal{A}} = \frac{\alpha_1^{Y_{\mathcal{A}}}{D'}_{\mathcal{A}} + \alpha_2^{Y_{\mathcal{A}}}{MS'}_{\mathcal{A}}}{\alpha_1^{Y_{\mathcal{A}}} + \alpha_2^{Y_{\mathcal{A}}}},
\label{eq:scr:exp_quiality}
\end{equation}

\noindent where the $MS_{\mathcal{A}}$ term is computed on the basis of $M_{\mathcal{A}}$, but taking the publication year of both the scoring method $\mathcal{A}$ and the compared method $m$ into account, as shown in Formula~\ref{eq:scr:methods_score}. As a result, every compared method $m$ contributes in an inversely proportional way to $MS$ with respect to the difference between the years of publication of the two methods \footnote{This difference is considered to be 1 for methods published in the same year, in order to avoid divisions by 0.}. This way, methods published in years close to the year of publication of $\mathcal{A}$ contribute more to $MS_{\mathcal{A}}$ that methods published long time before $\mathcal{A}$. In other words: the contribution of every method is proportional to its novelty in the year it is used to make comparisons. Please note that non CC methods are always considered to be published one year before the first CC was published (1999). Subsequently, the contribution of non CC methods to $MS_{\mathcal{A}}$ decays invariably with the years. By doing this, the first CC methods comparing with classic clustering methods are given credit by the comparison, as no CC baseline methods could have been established by that time. However, this is not the case for modern CC methods, which must be compared to other CC methods for said comparison to be meaningful.

\begin{equation}
MS_{\mathcal{A}} = \frac{1}{|M_{\mathcal{A}}|} \sum_{m \in M_{\mathcal{A}}} \frac{1}{Y_{\mathcal{A}} - Y_m}.
\label{eq:scr:methods_score}
\end{equation}
\\
Both the values of $D$ and $MS$ are normalized within each year following the normalization procedure described in Equation~\ref{eq:scr:exp_quiality_minmax} (min-max normalization), which results in $D'$ and $MS'$. This is done to lessen the effects that the computation capability context can have in $EQ_{\mathcal{A}}$. Please note that, only with respect to the year grouping aspects, methods published in years 2000-2003 are considered to belong to the same time context, hence they are treated as if they were published in the same year. With this in mind, neither Equation~\ref{eq:scr:methods_score} nor Equation~\ref{eq:scr:influence} are affected. This is done to enable withing-groups normalization and comparisons, as only 1 method was published in 2000 and 2001, and only 3 were published in 2002 and 2003. These were the years in which the CC research topic was conceived and it was starting to grow in interest (see Section~\ref{subsec:cc_history}). Subsequently, authors consider this exception to be justified.

\begin{equation}
{D'}_{\mathcal{A}} = \frac{D_{\mathcal{A}} - \text{min}(D^{Y_{\mathcal{A}}})}{ \text{max}(D^{Y_{\mathcal{A}}}) -  \text{min}(D^{Y_{\mathcal{A}}})}, \;\;\;
{MS'}_{\mathcal{A}} = \frac{MS_{\mathcal{A}} - \text{min}(MS^{Y_{\mathcal{A}}})}{ \text{max}(MS^{Y_{\mathcal{A}}}) -  \text{min}(MS^{Y_{\mathcal{A}}})}.
\label{eq:scr:exp_quiality_minmax}
\end{equation}
\\
The last elements to be introduced from Equation~\ref{eq:scr:methods_score} are the $\alpha_1^{Y_{\mathcal{A}}}$ and $\alpha_2^{Y_{\mathcal{A}}}$ values, which are different for every year. These values are used to determine the influence of the datasets score and the compared methods score in the computation of the experimental quality score. They are computed as in Equation~\ref{eq:scr:alphas}, where $\sigma(\cdot)$ and $\mu(\cdot)$ are functions which return the standard deviation and the mean of the list of values given as argument, respectively. Subsequently, $\alpha_1^{Y_{\mathcal{A}}}$ and $\alpha_2^{Y_{\mathcal{A}}}$ are directly proportional to the standard deviation of the datasets scores and the compared methods scores, respectively. In other words, $\alpha_1^{Y_{\mathcal{A}}}$ and $\alpha_2^{Y_{\mathcal{A}}}$ are used to give more importance to disperse measures, which are usually good discriminators, and therefore are better suited to be used in a scoring system.

\begin{equation}
\alpha_1^{Y_{\mathcal{A}}} = \frac{\sigma({D'}^{Y_{\mathcal{A}}})}{\mu({D'}^{Y_{\mathcal{A}}})}, \;\;\; \alpha_2^{Y_{\mathcal{A}}} = \frac{\sigma({MS'}^{Y_{\mathcal{A}}})}{\mu({MS'}^{Y_{\mathcal{A}}})}.
\label{eq:scr:alphas}
\end{equation}

\subsection{Scoring of the Validation Procedure Quality}

Once again, the information needed to determine the quality of the validation procedures $VQ$ used to evaluate the results obtained with a method $\mathcal{A}$ is fully contained in the study which proposes it. The two experimental elements (introduced in Section~\ref{sec:exp_setup}) that take part in this procedure are: the number of validity indices used to quantify the results obtained by the scoring method $\mathcal{A}$ (in list $V$), and the indicator of the use of statistical testing procedures (in list $T$). The list $T$ indicates which methods use statistical tests by giving them a value of 1, whereas the 0 value is assigned to method that do not support their conclusions with statistical tests. Equation~\ref{eq:scr:validation_score} shows the expression to compute the validation procedures quality of a scoring method $VQ_{\mathcal{A}}$.

\begin{equation}
VQ_{\mathcal{A}} = V'_{\mathcal{A}} + T_{\mathcal{A}},
\label{eq:scr:validation_score}
\end{equation}

\noindent where $V'_{\mathcal{A}}$ is the normalized value of $V'_{\mathcal{A}}$, which is computed following formula~\ref{eq:scr:valid_quiality_minmax}. Please note that in this case the min-max normalization does not take the publication year into account (in contrast to Equation~\ref{eq:scr:exp_quiality_minmax}), as the number of validity indices used to quantify the results of the proposed methods does not show any tendency with respect to the publication year (see Figure~\ref{fig:indices_statistics}). Authors consider studies which use statistical tests to have a significantly higher confidence rate in their results, hence the strength of the second term in Equation~\ref{eq:scr:validation_score}.

\begin{equation}
{V'}_{\mathcal{A}} = \frac{V_{\mathcal{A}} - \text{min}(V)}{ \text{max}(V) -  \text{min}(V)}.
\label{eq:scr:valid_quiality_minmax}
\end{equation}

\subsection{Scoring of the Influence}

The influence $I$ of a given method $\mathcal{A}$ cannot be computed with just the information contained in the study which proposes the method. This aspect of the method refers to how influential it has been in later literature, i.e. how many times method $\mathcal{A}$ has been used to make experimental comparisons. This number differs from the total number of times it has been cited, as a citation does not guarantee that the method is being used to make comparisons. In fact, this is one of the hardest aspects to evaluate, and requires experimental comparisons carried out in a corpus of papers to be self-contained. This means that no method referred in the experimental section of any paper is left out of the corpus. As will be explained in Section~\ref{sec:Taxonomy}, authors have made sure that this is the case for the taxonomy presented in this study. However, once this information has been obtained, an index for the influence of any given method can be computed as simply as in Equation~\ref{eq:scr:influence}, where CY refers to the current year, therefore $\text{CY} = 2022$. This is, the number of times a method is used in experimental comparisons divided by the number of years it has been available.

\begin{equation}
{I}_{\mathcal{A}} = \frac{C_{\mathcal{A}}}{\text{CY} - Y_{\mathcal{A}}}.
\label{eq:scr:influence}
\end{equation}

\subsection{Final Scoring}

The final scoring $S$ of any given method $\mathcal{A}$ can be computed by normalizing and adding up the three partial scores presented in previous sections, and scaling the output range to $[0,100]$. Equation~\ref{eq:scr:final_score} gives the expression to compute $S_{\mathcal{A}}$. Please note that none of the partial scores are bounded, hence the need of the min-max normalization step in Equation~\ref{eq:scr:final_minmax}.

\begin{equation}
{S}_{\mathcal{A}} = \frac{({EQ'}_{\mathcal{A}} + {VQ'}_{\mathcal{A}} + {I'}_{\mathcal{A}})\times 100}{3}.
\label{eq:scr:final_score}
\end{equation}

\begin{equation}
{EQ'}_{\mathcal{A}} = \frac{EQ_{\mathcal{A}} - \text{min}(EQ)}{ \text{max}(EQ) -  \text{min}(EQ)},\;\;\;
{VQ'}_{\mathcal{A}} = \frac{VQ_{\mathcal{A}} - \text{min}(VQ)}{ \text{max}(VQ) -  \text{min}(VQ)},\;\;\;
{I'}_{\mathcal{A}} = \frac{I_{\mathcal{A}} - \text{min}(I)}{ \text{max}(I) -  \text{min}(I)}.
\label{eq:scr:final_minmax}
\end{equation}
\\
Finally, authors want to remark that no hand-tuned parameter is needed to compute $S_{\mathcal{A}}$. Consequently, the probability of introducing any human bias in the scoring system is reduced.

\section{Taxonomic Review of Constrained Clustering Methods} \label{sec:Taxonomy}

In this section, a ranked taxonomic classification for a total of 307 CC methods is presented. The starting point to obtain the corpus of CC studies to be reviewed was to run Query~\ref{query:scopus_query} in the Scopus scientific database.

\begin{query}
    \textbf{Scopus Query}: ( TITLE-ABS-KEY ( "constrained clustering" )  OR  TITLE-ABS-KEY ( "semi-supervised clustering" )  AND  TITLE-ABS-KEY ( "constraint"  OR  "constraints"  OR  "constrained" ) )
    \label{query:scopus_query}
\end{query}

This is a very general and wide query, which was conceived to make sure that the most of the CC research area was contained in its output. This search outputted 1162 indexed scientific papers in 24/3/2022. Authors briefly reviewed and evaluated all of these papers to remove those which did not belong to the CC research area. Afterwards, a recursive procedure was used to obtain the final corpus to be reviewed: if a study compares its proposal with a CC proposal not included in the corpus, then the newly identified study is included and applied this procedure over. This is done with the aim of producing a self-contained comparison.

\begin{figure}[!ht]
	\centering
	\includegraphics[width=\linewidth]{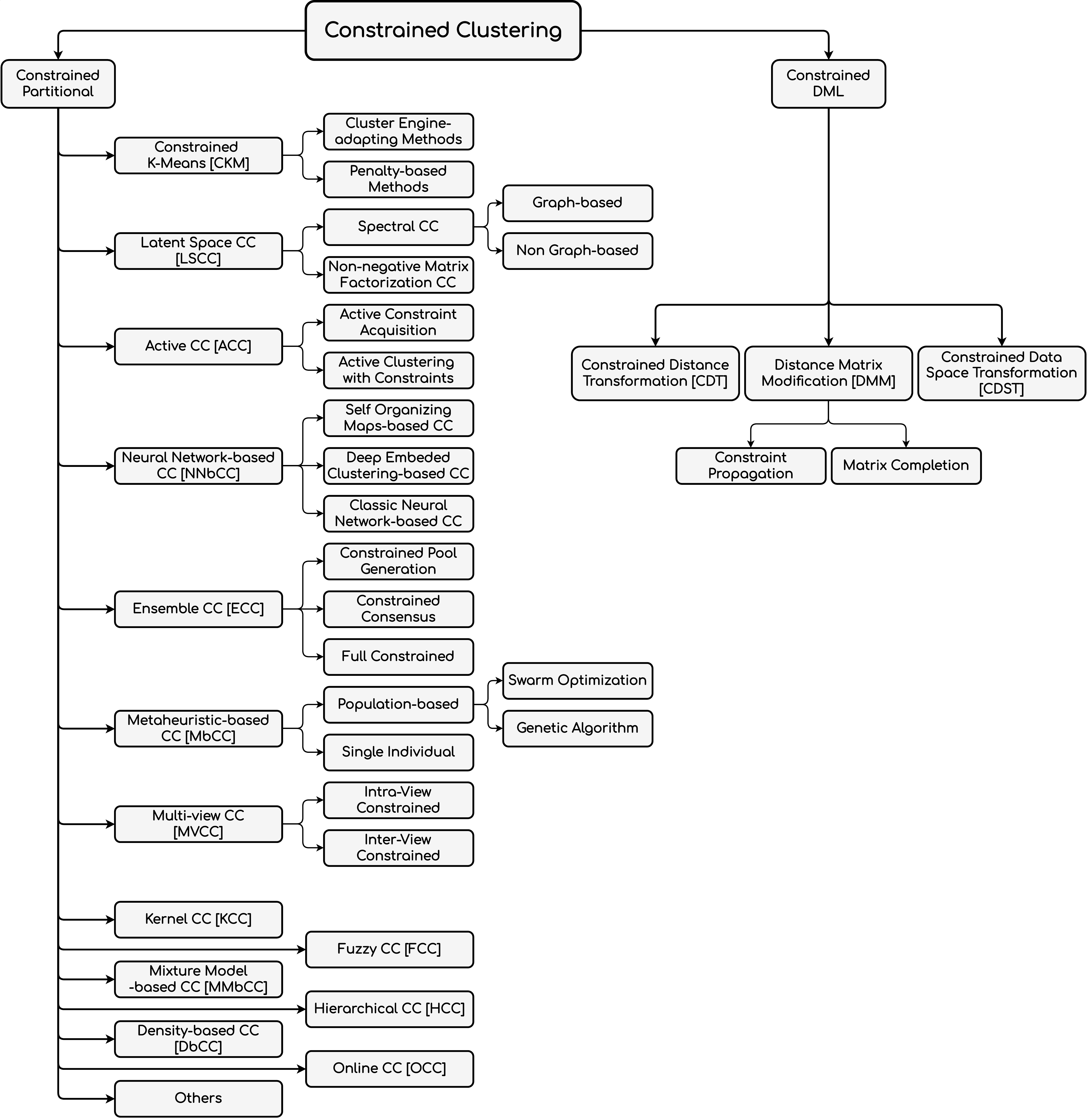}
	\caption{Taxonomic tree for the CC landscape.}
	\label{fig:taxonomic_tree}
\end{figure}

Figure~\ref{fig:taxonomic_tree} presents a taxonomic tree, organizing the categories in which the CC landscape may be divided. Particular methods are introduced and discussed in Sections~\ref{subsec:tax:constrained_kmeans} to~\ref{subsec:tax:cddt}, where tables detailing the features of every method can be found. 

As Figure~\ref{fig:taxonomic_tree} shows, a high-level dichotomy can be made within the CC area: constrained partitional methods versus constrained distance metric learning (DML) methods~\cite{davidson2007survey, basu2008constrained}. The main difference lies in their approach to CC and in their output. In constrained partitional methods, constraints are included into a procedure that progressively builds a partition for the dataset. This is typically done by designing a clustering engine which can deal with constraints or by including constraints in the objective function of a given method, for example, by means of a penalty term. Generally, constrained partitional methods produce a partition of the dataset, which may be accompanied by other by-products of the CC method, such as new constraints or feature weights. On the other hand, constrained DML methods aim to learn a distance metric that reflects the information contained in the constraint set. In general, the learned distances will try to bring ML instances together in the output space, while trying to maximize the distance between CL instances. Generally, constrained DML method do not produce a partition of the dataset, but a new metric, data space or distance matrix. This output can be used to later produce a partition by means of classic clustering algorithms, or even by constrained clustering algorithms. Please note that the difference found between the tree classes of constrained DML methods is merely conceptual, as the results of all the three of them (new metric/data space/distance matrix) can always be derived from each other using classic DML methods. However, the distinction between the three classes is useful from the point of view of CC, as their approach to the problem is different. The vast majority of CC methods are constrained partitional methods. There are hybrid methods, which combine features inherited from both approaches.

Feature tables in Sections~\ref{subsec:tax:constrained_kmeans} to~\ref{subsec:tax:cddt} generally include 8 columns:

\begin{itemize}

    \item The \textbf{$S_{\mathcal{A}}$} column gives the quality score assigned to each method. It is computed following the scoring system introduced in Section~\ref{sec:scoring_system}.

    \item The \textbf{Acronym} column provides the acronym of the method. Bearing in mind that some authors do not name their methods, we have decided to refer to these methods by the initials of their authors' names. However, there are exceptions for this rule, such as methods that are not named by their author but are consistently referred by later literature with a given name. In cases in which two methods have the same name, the year it was proposed in is added at the end of the name to differentiate them.
    
    \item The \textbf{Full Name} column gives the full name of the method. Methods named after the initials of their authors' names are marked with the word \textbf{names}. Methods whose name does not refer to neither any acronym nor authors' initials are marked with a hyphen dash (\textbf{"-"}).
    
    \item The \textbf{Penalty} column takes two values: "\checkmark" or "$\times$". This indicates whether constraints are included in the method by means of a penalty term in its objective function ("\checkmark") or by other means ("$\times$").
    
    \item The \textbf{ML} and \textbf{CL} columns refer to the type of constraints the method can handle. \textbf{Soft} is used for soft constraints, \textbf{Hard} is used for hard constraints, \textbf{Hybrid} is used for method that can use both hard and soft constraints. If a method cannot handle ML or CL constraints it is indicated with \textbf{"-"}.
    
    \item The \textbf{Hybrid} column indicates if the method belongs to more than one class, specifying the classes it belongs to. The \textbf{"-"} character is used if the method belongs to only one class.
    
    \item The \textbf{Year} and \textbf{Ref} columns provide the year of publication and the reference of the method respectively.
\end{itemize}

\subsection{Constrained K-Means} \label{subsec:tax:constrained_kmeans}

The Constrained K-Means (CKM) category gathers methods that can be considered modifications over the classic K-Means algorithm to include constraints. Their common feature is that all of them use an expectation-minimization (EM) optimization scheme. In the expectation step of an EM scheme, instances are assigned to clusters minimizing the error according to an objective function. In the minimization step, centroids are reestimated according to the assignations made in the expectation step. A plethora of objective functions and centroid update rules has emerged to approach the CC problem, although methods belonging to these category can be divided into two main categories. 

\subsubsection{Cluster Engine-adapting Methods}

 Cluster engine-adapting methods modify one of the two steps (or both) from the EM scheme in order to include constraints. Methods belonging to this category are presented in Table~\ref{tab:tax:cluster_engine_adapting_methods}. The first and most basic method performing CC this way is COP-K-Means. It modifies the instance to clusters assignation rule from the expectation step (the clustering engine) so that an instance is assigned to the cluster associated to its closer centroid whose assignation does not violate any constraint. Another popular technique in this category consists of performing clustering over the previously computed chunklet graph (which enforces ML), considering only CL in the expectation step. This is how methods like CLAC, CLWC, PCCK-Means, PCBK-Means or SSKMP perform CC. All of them consider hard ML. They differ from each other in the way in which they build their particular chunklet graph, which may contain weighted chunklets (as in CLAC and CLWC), or may rank chunklets in order for them to be examined more efficiently (as in PCCK-Means). Other methods use basic chunklet graph (like PCBK-Means and SSKMP). Some methods include constraints in EM scheme that are not basic K-Means, like SSKMP, which is a constrained version of the K-Medoids algorithm. Based on the COP-K-Means, methods like CLC-K-Means or ICOP-K-Means are designed to solve the dead-ends problem found in the basic algorithm.

\begin{table}[!h]
	\centering
	\setlength{\tabcolsep}{7pt}
	\renewcommand{\arraystretch}{1.3}
	\resizebox{\textwidth}{!}{
		\begin{tabular}{c l l c c c c c c}
		\hline
		$S_{\mathcal{A}}$ & Acronym & Full Name & Penalty & ML & CL & Hybrid & Year & Ref. \\
        \hline
        \cellcolor{scorecolor!100.0}33.33 & COP-K-Means & \makecell[l]{COnstrained Partitional - K-Means} & $\times$ & Hard & Hard & \makecell{---} & 2001 & \cite{wagstaff2001constrained} \\
\hhline{--}
\cellcolor{scorecolor!50.80859628234199}16.94 & PCSK-Means & \makecell[l]{Pairwise Constrained Spherical K-Means} & $\times$ & Hard & Soft & \makecell{---} & 2007 & \cite{tang2007enhancing} \\
\hhline{--}
\cellcolor{scorecolor!22.869704426325352}7.62 & CLAC & \makecell[l]{Constrained Locally Adaptive Clustering} & $\times$ & Hard & Hard & \makecell{Graph-based} & 2008 & \cite{domeniconi2008penta} \\
\hhline{--}
\cellcolor{scorecolor!22.689685347204236}7.56 & MLC-K-Means & \makecell[l]{Must-Link Constrained - K-Means} & $\times$ & Soft & Soft & \makecell{---} & 2008 & \cite{huang2008semi} \\
\hhline{--}
\cellcolor{scorecolor!37.630709537463666}12.54 & CLWC & \makecell[l]{Constrained Locally Weighted Clustering} & $\times$ & Hard & Soft & \makecell{---} & 2008 & \cite{cheng2008constrained} \\
\hhline{--}
\cellcolor{scorecolor!23.702528040839972}7.90 & COPGB-K-Means & \makecell[l]{COnstrained Partitional Graph-Based - K-Means} & $\times$ & Soft & Soft & \makecell{Graph-based} & 2008 & \cite{rothaus2008constrained} \\
\hhline{--}
\cellcolor{scorecolor!27.433839019791307}9.14 & PCCK-Means & \makecell[l]{Partial Closure-based Constrained K-Means} & $\times$ & Hard & Soft & \makecell{---} & 2008 & \cite{zhang2008partial} \\
\hhline{--}
\cellcolor{scorecolor!11.384815210086703}3.79 & SCK-Means & \makecell[l]{Soft Constrained K-Means} & $\times$ & Hybrid & Hybrid & \makecell{---} & 2009 & \cite{ares2009avoiding} \\
\hhline{--}
\cellcolor{scorecolor!15.482906262020935}5.16 & CMSC & \makecell[l]{Constrained Mean Shift Clustering} & $\times$ & Soft & Soft & \makecell{---} & 2009 & \cite{tuzel2009kernel} \\
\hhline{--}
\cellcolor{scorecolor!19.528429686706517}6.51 & SCKMM & \makecell[l]{Semi-supervised Clustering Kernel Method based on Metric learning} & $\times$ & Soft & Soft & \makecell{Constrained Distance Transformation} & 2010 & \cite{yin2010semi} \\
\hhline{--}
\cellcolor{scorecolor!18.12654760225858}6.04 & PCBK-Means & \makecell[l]{Pairwise Constrained Based K-Means} & $\times$ & Hard & Soft & \makecell{---} & 2010 & \cite{yin2010semi} \\
\hhline{--}
\cellcolor{scorecolor!17.471006109505012}5.82 & ICOP-K-Means & \makecell[l]{Improved COP-K-Means} & $\times$ & Hard & Hard & \makecell{---} & 2010 & \cite{tan2010improved} \\
\hhline{--}
\cellcolor{scorecolor!30.733450396604294}10.24 & CLC-K-Means & \makecell[l]{Cannot-Link Constrained – K-Means} & $\times$ & Hard & Hard & \makecell{---} & 2011 & \cite{rutayisire2011modified} \\
\hhline{--}
\cellcolor{scorecolor!15.72035203679389}5.24 & SKMS & \makecell[l]{Semi-supervised Kernel Mean Shift} & $\times$ & Soft & Soft & \makecell{---} & 2014 & \cite{anand2013semi} \\
\hhline{--}
\cellcolor{scorecolor!14.807202693642047}4.94 & BCK-Means & \makecell[l]{Binary Constrained K-Means} & $\times$ & Hard & Hard & \makecell{---} & 2019 & \cite{le2018binary} \\
\hhline{--}
\cellcolor{scorecolor!26.198038305348387}8.73 & SSKMP & \makecell[l]{Semi-Supervised K-Medioids Problem} & $\times$ & Hard & Hard & \makecell{---} & 2019 & \cite{randel2018k}  \\
        \hline
		\end{tabular}}
	\caption{Feature table for CKM - Cluster Engine-adapting Methods.}
	\label{tab:tax:cluster_engine_adapting_methods}
\end{table}

\subsubsection{Penalty-based Methods}

Penalty-based methods include constraints by means of a penalty term in the objective function of an EM scheme. These methods are presented in Table~\ref{tab:tax:penalty_based_methods}. Some of them simply modify previous CC or classic clustering algorithms to include a plain penalty term, such as SCOP-K-Means, PCK-Means, S-SCAD. Other methods, like MPCK-Means, also include a metric learning step in the EM scheme, which estimates a cluster-local distance measure for every cluster, and allows them to find clusters with arbitrary shapes. Besides, there are methods which use variable penalty terms, such as HMRF-K-Means, CVQE, LCVQE or CVQE+ that include the distance between constrained instances in it. This is, more relevance is assigned to ML relating distant instances and CL relating close instances. Methods which combine pairwise constraints and other types of constraints have also emerged, like PCS, which includes cluster-size constraints in its EM scheme too. Other methods like GPK-Means use a Gaussian function and the current cluster centroids to infer new constraints in the neighborhood of the original constraints. These new constraints are added to the constraint set and used in subsequent iterations of the EM scheme.

\begin{table}[!h]
	\centering
	\setlength{\tabcolsep}{7pt}
	\renewcommand{\arraystretch}{1.3}
	\resizebox{\textwidth}{!}{
		\begin{tabular}{c l l c c c c c c}
		\hline
		$S_{\mathcal{A}}$ & Acronym & Full Name & Penalty & ML & CL & Hybrid & Year & Ref. \\
		\hline
        \cellcolor{scorecolor!0.0}0.00 & SCOP-K-Means & \makecell[l]{Soft COnstrained Partitional - K-Means} & \checkmark & Soft & Soft & \makecell{---} & 2002 & \cite{wagstaff2002intelligent} \\
\hhline{--}
\cellcolor{scorecolor!78.50329531615233}16.18 & PCK-Means & \makecell[l]{Pairwise Constrained K-Means} & \checkmark & Soft & Soft & \makecell{---} & 2003 & \cite{basu2003comparing,basu2004active} \\
\hhline{--}
\cellcolor{scorecolor!100.0}20.61 & MPCK-Means & \makecell[l]{Metric Pairwise Constrained K-Means} & \checkmark & Soft & Soft & \makecell{Constrained Distance Transformation} & 2003 & \cite{basu2003comparing,bilenko2004integrating} \\
\hhline{--}
\cellcolor{scorecolor!43.542942383036376}8.97 & HMRF-K-Means & \makecell[l]{Hidden Markov Random Fields - K-Means} & \checkmark & Soft & Soft & \makecell{Constrained Distance Transformation} & 2004 & \cite{basu2004probabilistic} \\
\hhline{--}
\cellcolor{scorecolor!67.8666179337723}13.99 & GPK-Means & \makecell[l]{Gaussian Propagated K-Means} & \checkmark & Soft & Soft & \makecell{---} & 2005 & \cite{eaton2005clustering} \\
\hhline{--}
\cellcolor{scorecolor!42.66162108754658}8.79 & CVQE & \makecell[l]{Constrained Vector Quantization Error} & \checkmark & Soft & Soft & \makecell{---} & 2005 & \cite{davidson2005clustering} \\
\hhline{--}
\cellcolor{scorecolor!43.27710046384552}8.92 & LCVQE & \makecell[l]{Linear Constrained Vector Quantization Error} & \checkmark & Soft & Soft & \makecell{---} & 2007 & \cite{pelleg2007k} \\
\hhline{--}
\cellcolor{scorecolor!29.88489304658422}6.16 & S-SCAD & \makecell[l]{Semi-Supervised Clustering and Attribute Discrimination} & \checkmark & Soft & Soft & \makecell{---} & 2007 & \cite{frigui2007semi} \\
\hhline{--}
\cellcolor{scorecolor!21.222449932257963}4.37 & SemiStream & \makecell[l]{---} & \checkmark & Soft & Soft & \makecell{Online CC} & 2012 & \cite{halkidi2012semi} \\
\hhline{--}
\cellcolor{scorecolor!85.58980494450188}17.64 & PC-HCM-NM & \makecell[l]{Pairwise Constrained - Hard C-Means - Non Metric} & \checkmark & Soft & Soft & \makecell{---} & 2014 & \cite{endo2014hard} \\
\hhline{--}
\cellcolor{scorecolor!12.200868832395491}2.51 & AC-CF-tree & \makecell[l]{Active Constrained - Clustering Feature - tree} & \checkmark & Soft & Soft & \makecell{Active Clustering with Constraints} & 2014 & \cite{lai2014new} \\
\hhline{--}
\cellcolor{scorecolor!22.70858564100711}4.68 & PCS & \makecell[l]{PCK-Means with Size constraints} & \checkmark & Soft & Soft & \makecell{---} & 2014 & \cite{zhang2014semi} \\
\hhline{--}
\cellcolor{scorecolor!21.053302543655796}4.34 & TDCK-Means & \makecell[l]{Temporal-Driven Constrained K-Means} & \checkmark & Soft & Soft & \makecell{Online CC} & 2014 & \cite{rizoiu2014use} \\
\hhline{--}
\cellcolor{scorecolor!60.485054265916624}12.47 & HSCE & \makecell[l]{Hybrid Semi-supervised Clustering Ensemble} & \checkmark & Soft & Soft & \makecell{Non Graph-based \& Constrained Pool Generation} & 2015 & \cite{wei2015semi,wei2018combined} \\
\hhline{--}
\cellcolor{scorecolor!11.970569311648113}2.47 & CVQE+ & \makecell[l]{Constrained Vector Quantization Error +} & \checkmark & Soft & Soft & \makecell{Active Clustering with Constraints} & 2018 & \cite{mai2018scalable} \\
\hhline{--}
\cellcolor{scorecolor!14.543991565125246}3.00 & PCSK-Means(21) & \makecell[l]{Pairwise Constrained Sparse K-Means} & \checkmark & Soft & Soft & \makecell{---} & 2021 & \cite{vouros2021semi} \\
\hhline{--}
\cellcolor{scorecolor!81.50578850811277}16.80 & fssK-Means & \makecell[l]{fast semi-supervised K-Means} & \checkmark & Soft & Soft & \makecell{Time Series} & 2021 & \cite{he2019fast}  \\
        \hline
		\end{tabular}}
	\caption{Feature table for CKM - Penalty-based Methods.}
	\label{tab:tax:penalty_based_methods}
\end{table}

\subsection{Latent Space CC}

Latent space clustering performs clustering in a space which is different from the input space and which is computed on the basis of the dataset, and also the constraint set in Latent Space CC (LSCC). The input to these algorithms is an adjacency matrix defining the topology of the network (or graph) over which clustering needs to be performed. Each row or column may be regarded as the feature or property representation of the corresponding node. Latent space clustering methods first obtain new property representations in a latent space for each node by optimizing different objective functions, and then clusters nodes in that latent space~\cite{yang2014unified}.

\subsubsection{Spectral CC}

Classic spectral clustering algorithms try to obtain the latent space by finding the most meaningful eigenvectors of the adjacency matrix, which are used to define the embedding in which clusters are eventually obtained~\cite{yang2014unified}. A dichotomy can be made within this category: in graph-based methods the input data is always given in the form of a graph, while in non-graph based the input is an adjacency matrix or a regular dataset, which can be transformed into an adjacency matrix. Please note that this distinction only affects the conceptual level of the spectral CC category, as a graph can always be converted to and adjacency matrix and vice versa, all methods from one category may also be applied in the other category. However, the authors have decided to make this distinction, since the terminology and concepts used in the studies referring to each of them differ greatly and can be misleading if interpreted together.

\paragraph{Graph-based spectral clustering} In graph-based spectral clustering, the input is assumed to be a graph. The goal is to partition the set of vertices of the graph, taking into account the information contained in the vertices themselves and in the edges of the graph. Edges may carry similarity or dissimilarity information regarding the vertices they connect. Some common strategies to perform graph clustering try to maximize the similarity of vertices within a cluster, normalizing the contribution of each to the objective by the size of the cluster in order to balance the size of the clusters. Other methods try to minimize the total cost of the edges crossing the cluster boundaries~\cite{domeniconi2011composite}. Graph-based methods are particularly suitable to perform CC, as constraints can be naturally represented in their graph form, which is the constraint graph and the chunklet graph (introduced in Section~\ref{sec:cc_concepts_structures}).

Table~\ref{tab:tax:spectral_graph} gathers graph-based spectral clustering methods. COP-b-coloring and CLAC exemplify the use of chunklets to enforce ML, while including CL by other means. Other methods modify the input graph to include the information contained in the constraint set. For example, PCOG modifies affinities so that ML instances are always placed in the same connected components and removes edges which connect CL instances. CCHAMELEON modifies affinities between constrained instances, making them larger if instances are related by ML and lower in the case of CL. The all-pairs-shortest-path algorithm is used to propagate changes. PAST-Toss uses a spanning tree based technique to perform CC directly over the constraint graph. SCRAWL is the only non-spectral graph-based CC algorithm, as it does not need pairwise similarity/dissimilarity information to perform CC, but graph-related measures instead.

\begin{table}[!h]
	\centering
	\setlength{\tabcolsep}{7pt}
	\renewcommand{\arraystretch}{1.3}
	\resizebox{\textwidth}{!}{
		\begin{tabular}{c l l c c c c c c}
		\hline
		$S_{\mathcal{A}}$ & Acronym & Full Name & Penalty & ML & CL & Hybrid & Year & Ref. \\
		\hline
        \cellcolor{scorecolor!38.614877959889306}4.35 & COP-b-coloring & \makecell[l]{COnstrained Partitional - b-coloring} & $\times$ & Hard & Hard & \makecell{---} & 2007 & \cite{elghazel2007constrained} \\
\hhline{--}
\cellcolor{scorecolor!31.610280737755993}3.56 & PAST-Toss & \makecell[l]{Pick A Spanning Tree – Toss} & $\times$ & Soft & Soft & \makecell{Single Individual} & 2008 & \cite{coleman2008local} \\
\hhline{--}
\cellcolor{scorecolor!67.65692935006804}7.62 & CLAC & \makecell[l]{Constrained Locally Adaptive Clustering} & $\times$ & Hard & Hard & \makecell{Cluster Engine-adapting Methods} & 2008 & \cite{domeniconi2008penta} \\
\hhline{--}
\cellcolor{scorecolor!70.12072544458266}7.90 & COPGB-K-Means & \makecell[l]{COnstrained Partitional Graph-Based - K-Means} & $\times$ & Soft & Soft & \makecell{Cluster Engine-adapting Methods} & 2008 & \cite{rothaus2008constrained} \\
\hhline{--}
\cellcolor{scorecolor!73.5923272118147}8.29 & GBSSC & \makecell[l]{Graph-Based Semi-Supervised Clustering} & $\times$ & Hard & Soft & \makecell{Dimensionality Reduction} & 2010 & \cite{yoshida2010performance,yoshida2010graph,yoshida2011pairwise,yoshida2014graph} \\
\hhline{--}
\cellcolor{scorecolor!40.258403658319644}4.54 & CCHAMELEON & \makecell[l]{Constrained CHAMELEON} & $\times$ & Soft & Soft & \makecell{---} & 2011 & \cite{anand2011graph} \\
\hhline{--}
\cellcolor{scorecolor!100.0}11.27 & SCRAWL & \makecell[l]{Semi-supervised Clustering via RAndom WaLk} & $\times$ & Soft & Soft & \makecell{---} & 2014 & \cite{he2014semi} \\
\hhline{--}
\cellcolor{scorecolor!47.618055989682745}5.37 & PCOG & \makecell[l]{Pairwise Constrained Optimal Graph} & $\times$ & Hard & Hard & \makecell{Non Graph-based} & 2020 & \cite{nie2021semi}  \\
        \hline
		\end{tabular}}
	\caption{Feature table for the LSCC - Spectral CC - Graph-based methods.}
	\label{tab:tax:spectral_graph}
\end{table}

\paragraph{Non graph-based spectral clustering} In these methods, the input is given in the form of an adjacency matrix, or a dataset whose adjacency matrix can be easily obtained. Two techniques are commonly used to include pairwise constraints in these methods. (1) Modifying similarities/dissimilarities in the original adjacency matrix, computing eigenvectors and eigenvalues to obtain the spectral embedding. (2) Using the constraints to directly modify the embedding, obtained on the basis of the original adjacency matrix. Any classic clustering method can be used to obtain the final partition in the new embedding, which can always be mapped to the original data~\cite{yang2014unified}. 

Table~\ref{tab:tax:spectral_nongraph} shows a list of non-graph-based spectral CC methods. The first spectral CC method is found in KKM/SL, known in the literature by these two acronyms (respectively obtained from the name of its authors and the title of the study which proposes it). It is based on HMRF, performing CC by modifying the transition probabilities of the field based on the constraints. KKM/SL and AHMRF constitute the only two HMRF-based approaches to spectral CC. Another common technique in spectral CC is learning a kernel matrix based on the dataset over which spectral clustering is later conducted, as in RSCPC, CCSR, CCSKL, LSE or SSCA. This kernel matrix is usually built taking both pairwise distances and constraints into account. With respect to the methods which modify the original adjacency matrix, two strategies are the most used ones: some methods, such as ACCESS or CSC, simply set entries which relate constrained instances to specific fixed values, while other methods, such as NSDR-NCuts or LCPN, use constraint propagation techniques to propagate changes in the affinity matrix once it has been modified.

\begin{table}[!h]
	\centering
	\setlength{\tabcolsep}{7pt}
	\renewcommand{\arraystretch}{1.3}
	\resizebox{\textwidth}{!}{
		\begin{tabular}{c l l c c c c c c}
		\hline
		$S_{\mathcal{A}}$ & Acronym & Full Name & Penalty & ML & CL & Hybrid & Year & Ref. \\
		\hline
        \cellcolor{scorecolor!79.25107994658325}14.40 & KKM & \makecell[l]{names} & $\times$ & Soft & Soft & \makecell{---} & 2003 & \cite{kamvar2003spectral} \\
\hhline{--}
\cellcolor{scorecolor!61.15512289418545}11.11 & TBJSBM & \makecell[l]{names} & $\times$ & Hybrid & Hybrid & \makecell{---} & 2004 & \cite{bie2004learning} \\
\hhline{--}
\cellcolor{scorecolor!70.05875284946647}12.73 & ACCESS & \makecell[l]{Active Constrained Clustering by Examining Spectral eigenvectorS} & $\times$ & Soft & Soft & \makecell{Active Constraint Acquisition} & 2005 & \cite{xu2005active} \\
\hhline{--}
\cellcolor{scorecolor!37.882778040836904}6.88 & CSC & \makecell[l]{Constrained Spectral Clustering} & $\times$ & Soft & Soft & \makecell{---} & 2005 & \cite{xu2005constrained} \\
\hhline{--}
\cellcolor{scorecolor!68.64235677131302}12.47 & LCPN & \makecell[l]{names} & $\times$ & Soft & Soft & \makecell{Constraint Propagation} & 2008 & \cite{lu2008constrained} \\
\hhline{--}
\cellcolor{scorecolor!46.230130427098324}8.40 & S3-K-Means & \makecell[l]{Semi-Supervised Spectral K-Means} & \checkmark & Soft & Soft & \makecell{---} & 2008 & \cite{hu2008towards} \\
\hhline{--}
\cellcolor{scorecolor!20.224596629784113}3.68 & RSCPC & \makecell[l]{Regularized Spectral Clustering with Pairwise Constraints} & $\times$ & Soft & Soft & \makecell{---} & 2009 & \cite{alzate2009regularized} \\
\hhline{--}
\cellcolor{scorecolor!84.67356990698508}15.39 & CCSR & \makecell[l]{Constrained Clustering with Spectral Regularization} & \checkmark & Soft & Soft & \makecell{---} & 2009 & \cite{li2009constrained} \\
\hhline{--}
\cellcolor{scorecolor!37.528830081393544}6.82 & SCLC & \makecell[l]{Spectral Clustering with Linear Constraints} & $\times$ & Soft & Soft & \makecell{---} & 2009 & \cite{xu2009fast} \\
\hhline{--}
\cellcolor{scorecolor!87.02504210114202}15.81 & CCSKL & \makecell[l]{Constrained Clustering by Spectral Kernel Learning} & $\times$ & Soft & Soft & \makecell{Kernel CC} & 2009 & \cite{li2009constrained2} \\
\hhline{--}
\cellcolor{scorecolor!48.221389260206415}8.76 & CSP & \makecell[l]{Constrained SPectral clustering} & $\times$ & Hybrid & Hybrid & \makecell{---} & 2010 & \cite{wang2010flexible} \\
\hhline{--}
\cellcolor{scorecolor!74.14911708415039}13.47 & ASC & \makecell[l]{Active Spectral Clustering} & $\times$ & Any & Any & \makecell{Active Clustering with Constraints} & 2010 & \cite{wang2010active} \\
\hhline{--}
\cellcolor{scorecolor!47.180263283347585}8.57 & SSC-ESE & \makecell[l]{Semi-Supervised Clustering with Enhanced Spectral Embedding} & \checkmark & Soft & Soft & \makecell{---} & 2012 & \cite{jiao2012fast} \\
\hhline{--}
\cellcolor{scorecolor!21.467901221797376}3.90 & IU-Red & \makecell[l]{Iterative Uncertainty Reduction} & $\times$ & Soft & Soft & \makecell{Active Clustering with Constraints} & 2012 & \cite{wauthier2012active} \\
\hhline{--}
\cellcolor{scorecolor!31.06913405815657}5.65 & LSE & \makecell[l]{Learned Spectral Embedding} & $\times$ & Soft & Soft & \makecell{---} & 2012 & \cite{shang2012learning} \\
\hhline{--}
\cellcolor{scorecolor!42.52497144965676}7.73 & NSDR-NCuts & \makecell[l]{Near Stranger or Distant Relatives – NCuts} & $\times$ & Soft & Soft & \makecell{---} & 2012 & \cite{chen2012spectral} \\
\hhline{--}
\cellcolor{scorecolor!73.64692813673126}13.38 & COSC & \makecell[l]{Constrained One Spectral Clustering} & $\times$ & Hybrid & Hybrid & \makecell{---} & 2012 & \cite{rangapuram2012constrained} \\
\hhline{--}
\cellcolor{scorecolor!19.907523118070714}3.62 & SSCA & \makecell[l]{Semi-supervised Spectral Clustering Algorithm} & $\times$ & Soft & Soft & \makecell{---} & 2014 & \cite{ding2014research} \\
\hhline{--}
\cellcolor{scorecolor!37.30723251103294}6.78 & FHCSC & \makecell[l]{Flexible Highly Constrained Spectral Clustering} & $\times$ & Hybrid & Hybrid & \makecell{---} & 2014 & \cite{wang2014constrained} \\
\hhline{--}
\cellcolor{scorecolor!38.142024301829004}6.93 & CNP-K-Means & \makecell[l]{Constraint Neighborhood Projections - K-Means} & $\times$ & Soft & Soft & \makecell{Dimensionality Reduction} & 2014 & \cite{wang2014constraint} \\
\hhline{--}
\cellcolor{scorecolor!24.534514939937125}4.46 & STSC & \makecell[l]{Self-Taught Spectral Clustering} & $\times$ & Soft & Soft & \makecell{Matrix Completion} & 2014 & \cite{wang2014self} \\
\hhline{--}
\cellcolor{scorecolor!32.91793271530336}5.98 & LXDXD & \makecell[l]{names} & $\times$ & Soft & Soft & \makecell{Non-negative Matrix Factorization CC} & 2015 & \cite{yang2014unified} \\
\hhline{--}
\cellcolor{scorecolor!68.60290830974076}12.47 & HSCE & \makecell[l]{Hybrid Semi-supervised Clustering Ensemble} & $\times$ & Soft & Soft & \makecell{Constrained Pool Generation \& Penalty-based Methods} & 2015 & \cite{wei2015semi,wei2018combined} \\
\hhline{--}
\cellcolor{scorecolor!48.54726294834813}8.82 & FAST-GE & \makecell[l]{Fast-Generalized Spectral Clustering} & $\times$ & Soft & Soft & \makecell{---} & 2016 & \cite{cucuringu2016simple} \\
\hhline{--}
\cellcolor{scorecolor!36.6201962580703}6.65 & URASC & \makecell[l]{Uncertainty Reducing Active Spectral Clustering} & $\times$ & Soft & Soft & \makecell{Active Clustering with Constraints} & 2017 & \cite{xiong2017active} \\
\hhline{--}
\cellcolor{scorecolor!91.04287350104431}16.54 & FAST-GE2.0 & \makecell[l]{Fast-Generalized Spectral Clustering 2.0} & $\times$ & Soft & Soft & \makecell{---} & 2017 & \cite{jiang2017robust} \\
\hhline{--}
\cellcolor{scorecolor!66.65562047017063}12.11 & TI-APJCF & \makecell[l]{Type-I Affinity and Penalty Jointly Constrained Spectral Clustering} & \checkmark & Soft & Soft & \makecell{---} & 2017 & \cite{qian2016affinity} \\
\hhline{--}
\cellcolor{scorecolor!66.65562047017063}12.11 & TII-APJCF & \makecell[l]{Type-II Affinity and Penalty Jointly Constrained Spectral Clustering} & \checkmark & Soft & Soft & \makecell{---} & 2017 & \cite{qian2016affinity} \\
\hhline{--}
\cellcolor{scorecolor!10.375480822717241}1.89 & AHMRF & \makecell[l]{A - Hidden Markov Random Field} & $\times$ & Soft & Soft & \makecell{---} & 2018 & \cite{ding2018semi} \\
\hhline{--}
\cellcolor{scorecolor!100.0}18.17 & MVCSC & \makecell[l]{Multi-View Constrained Spectral Clustering} & $\times$ & Soft & Soft & \makecell{Intra-View Constrained} & 2019 & \cite{chen2019auto} \\
\hhline{--}
\cellcolor{scorecolor!82.74659585362798}15.04 & SFS$^3$EC & \makecell[l]{Stratified Feature Sampling for Semi-Supervised Ensemble Clustering} & $\times$ & Soft & Soft & \makecell{Full Constrained} & 2019 & \cite{tian2019stratified} \\
\hhline{--}
\cellcolor{scorecolor!29.526217306580666}5.37 & PCOG & \makecell[l]{Pairwise Constrained Optimal Graph} & $\times$ & Hard & Hard & \makecell{Graph-based} & 2020 & \cite{nie2021semi}  \\
        \hline
		\end{tabular}}
	\caption{Feature table for the LSCC - Spectral CC - Non Graph-based methods.}
	\label{tab:tax:spectral_nongraph}
\end{table}

\subsubsection{Non-negative Matrix Factorization CC}

Non-negative Matrix Factorization (NMF) clustering algorithms obtain the new representation of the data by factorizing the adjacency matrix into two non-negative matrices~\cite{yang2014unified}. These two matrices can be interpreted as the centroids of the partition and the membership degree of each instances to each cluster. By doing this, all instances can be obtained as a linear combination of each column of the centroids matrix, parameterized by its corresponding membership, found in its associated row from the membership matrix. It can be proven that minimizing the difference between the original dataset matrix and the product matrix (computed usually as the Frobenius norm) is equivalent to performing K-Means clustering over the dataset~\cite{zhang2015constrained}.

Table~\ref{tab:tax:NMF_CC} presents a list of NMF-based CC methods. One of the most common strategies to include constraints into the classic NMF-based methods is to modify its objective function. This can be done by means of a penalty term accounting for the number of violated constraints, as in PNMF, SSCsNMF and SS-NMF(08), or by more complex techniques, as in CPSNMF or NMFS. Another popular strategy is forcing affinities between ML instances to be 0 and affinities between CL instances to be 1, as in NMFCC and SymNMFCC.

\begin{table}[!h]
	\centering
	\setlength{\tabcolsep}{7pt}
	\renewcommand{\arraystretch}{1.3}
	\resizebox{\textwidth}{!}{
		\begin{tabular}{c l l c c c c c c}
		\hline
		$S_{\mathcal{A}}$ & Acronym & Full Name & Penalty & ML & CL & Hybrid & Year & Ref. \\
		\hline
        \cellcolor{scorecolor!34.8774403474117}5.68 & NMFS & \makecell[l]{Non-negative Matrix Factorization-based Semi-supervised} & $\times$ & Soft & Soft & \makecell{---} & 2007 & \cite{li2007solving} \\
\hhline{--}
\cellcolor{scorecolor!100.0}16.28 & SS-NMF(08) & \makecell[l]{Semi-Supervised - Non-negative Matrix Factorization} & $\times$ & Soft & Soft & \makecell{---} & 2008 & \cite{chen2008non} \\
\hhline{--}
\cellcolor{scorecolor!29.134444428730248}4.74 & OSS-NMF & \makecell[l]{Orthogonal Semi-Supervised - Non-negative Matrix tri-Factorization} & $\times$ & Soft & Soft & \makecell{Co-Clustering} & 2010 & \cite{ma2010orthogonal} \\
\hhline{--}
\cellcolor{scorecolor!73.00359455677273}11.88 & SS-NMF & \makecell[l]{Semi-Supervised Nonnegative Matrix Factorization} & $\times$ & Soft & Soft & \makecell{Co-Clustering} & 2010 & \cite{chen2009non} \\
\hhline{--}
\cellcolor{scorecolor!27.69966994338114}4.51 & PNMF & \makecell[l]{Penalty - Nonnegative Matrix Factorization} & \checkmark & Soft & Soft & \makecell{---} & 2011 & \cite{zhu2011text} \\
\hhline{--}
\cellcolor{scorecolor!39.74671413823961}6.47 & SSCsNMF & \makecell[l]{Semi-supervised symmetriC Non-negative Matrix Factorization} & $\times$ & Soft & Soft & \makecell{---} & 2012 & \cite{jing2012semi} \\
\hhline{--}
\cellcolor{scorecolor!36.75360650740112}5.98 & LXDXD & \makecell[l]{names} & $\times$ & Soft & Soft & \makecell{Non Graph-based} & 2015 & \cite{yang2014unified} \\
\hhline{--}
\cellcolor{scorecolor!27.313013918944307}4.45 & CPSNMF & \makecell[l]{Constrained Propagation for Semi-supervised Nonnegative Matrix Factorization} & $\times$ & Soft & Soft & \makecell{Constraint Propagation} & 2016 & \cite{wang2015semi} \\
\hhline{--}
\cellcolor{scorecolor!50.397879205194606}8.20 & NMFCC & \makecell[l]{Non-negative Matrix Factorization based Constrained Clustering} & $\times$ & Soft & Soft & \makecell{---} & 2016 & \cite{zhang2016constrained} \\
\hhline{--}
\cellcolor{scorecolor!73.3676553478476}11.94 & SymNMFCC & \makecell[l]{Symmetric Non-negative Matrix Factorization based Constrained Clustering} & $\times$ & Soft & Soft & \makecell{---} & 2016 & \cite{zhang2016constrained} \\
\hhline{--}
\cellcolor{scorecolor!30.869498999402246}5.02 & PCPSNMF & \makecell[l]{Pairwise Constraint Propagation-induced Symmetric NMF} & $\times$ & Soft & Soft & \makecell{---} & 2018 & \cite{wu2018pairwise} \\
\hhline{--}
\cellcolor{scorecolor!17.51096925343264}2.85 & CMVNMF & \makecell[l]{Constrained Multi-View NMF} & $\times$ & Soft & Soft & \makecell{Inter-View Constrained \& Active Clustering with Constraints} & 2018 & \cite{zong2018multi,zhang2015constrained}  \\
        \hline
		\end{tabular}}
	\caption{Feature table for the LSCC - Non-negative Matrix Factorization CC methods.}
	\label{tab:tax:NMF_CC}
\end{table}

\subsection{Active CC}

Active learning is a subfield of machine learning in which algorithms are allowed to choose the data from which they learn. The goal of active learning is to reduce the amount of supervisory information needed to learn, an therefore reduce the human effort and implication in machine learning. In the active learning paradigm, learning methods are provided with an oracle, which is capable of answering a limited number of an specific type of query. For example, in traditional classification, active learning is used to select the best instances to be labeled from a dataset, so the oracle provides the label of the specific queried instance~\cite{settles2009active}. In Active CC (ACC), the oracle is queried about the type of constraint relating pairs of instances. The key aspect in any active learning algorithm is how to choose the queries to be presented to the oracle.

Active learning is specially useful in CC. In order to have an explanation for this, we compare the complexity of the answers given by oracles involved in active classification and ACC. In active classification, the oracle is queried about the class of a given instance. This query has a virtually infinite number of answers, as the number of classes in the dataset may be unknown. On the other hand, an oracle involved in CC is queried with two instances and asked about the constraint between them (ML or CL), which is the same as asking whether they belong to the same class. There are only three possible answers to this question: "yes", "no" or "unknown". It is clear that the oracle in CC carries out a far simpler job than the one in classification. Let us remember that the oracle is just an abstraction of a knowledge source, which is generally a human user. Querying a human about the relation between instances instead of about their class requires less effort from them and leads to less variability and noise in the queries, as the extensive literature in active CC shows.

Two subcategories can be found in Active CC. In active constraint acquisition, constraints are actively generated before performing constrained clustering, while in active clustering with constraints, both clustering and active constraint generation are performed iteratively at the same time. This way, in active constraint acquisition queries are generated on the basis of the dataset and the current state of the constraint set, while in active clustering with constraints information about the current partition can also be used.



\subsubsection{Active Constraint Acquisition}

The immediate result of active constraint acquisition methods is a set of constraints, rather than a partition of the dataset. However, the partition can be obtained using any other CC method by just feeding the generated constraints into it, along with the dataset used to generate the constraints. Only the dataset, the constraint set generated so far, and an initial unconstrained partition are available to perform active constraint learning in the active constraint acquisition paradigm. No CC algorithm is involved in the constraint acquisition step. Table~\ref{tab:tax:active_constraint_acquisition} shows a list of active constraint acquisition methods. Columns indicating the type of constraint these methods can handle have been removed, as they are not relevant here. Column ``CC Method'' has been added, indicating the CC method used to produce a partition based on the constraint generated by every active constraint acquisition method in the experimental section of the studies that propose them.

Many strategies to select the best pair of instances to query to the oracle have been proposed. Some methods start by dividing the dataset into preliminary groups and then use the oracle to query constraints which consolidate that information, such as FFQS, MMFFQS, SSL-EC or LCML. Other methods focus on finding the boundaries in the dataset to select pairs of instances from them, such as ACCESS, ASC(10) or SACS. Besides, there are methods that use classic clustering to obtain preliminary information from the dataset, such as the co-association of instances or the compactness of clusters. DGPC, JDFD, WAKL, MICS, AAA(19), AIPC, ALPCS or ASCENT are some of the methods which use this strategy. Other methods focus on specific features from the constraint themselves in order to evaluate them and select the more informative ones, such as AAVV or KAKB. More complex approaches can be found in AAA(18), which solves the active constraint acquisition problem as an instance of the uncapacited k-facility location problem, or RWACS, which uses the commute time from graph theory to select queries.

\begin{table}[!h]
	\centering
	\setlength{\tabcolsep}{7pt}
	\renewcommand{\arraystretch}{1.3}
	\resizebox{\textwidth}{!}{
		\begin{tabular}{c l l c c c c}
		\hline
		$S_{\mathcal{A}}$ & Acronym & Full Name & CC Methods & Hybrid & Year & Ref. \\
		\hline
        \cellcolor{scorecolor!88.566075530866}16.24 & FFQS & \makecell[l]{Farthest First Query Selection} & --- & \makecell{---} & 2004 & \cite{basu2004active} \\
\hhline{--}
\cellcolor{scorecolor!69.40756339092624}12.73 & ACCESS & \makecell[l]{Active Constrained Clustering by Examining Spectral eigenvectorS} & --- & \makecell{Non Graph-based} & 2005 & \cite{xu2005active} \\
\hhline{--}
\cellcolor{scorecolor!38.179726670227474}7.00 & DGPC & \makecell[l]{names} & PCK-Means & \makecell{---} & 2007 & \cite{greene2007constraint} \\
\hhline{--}
\cellcolor{scorecolor!71.9113051974586}13.19 & MMFFQS & \makecell[l]{Min-Max Farthest First Query Selection} & MPCK-Means & \makecell{---} & 2008 & \cite{mallapragada2008active} \\
\hhline{--}
\cellcolor{scorecolor!51.37176905460505}9.42 & ASC(10) & \makecell[l]{Ability to Separate between Clusters} & AHCC, MPCK-Means & \makecell{---} & 2010 & \cite{vu2010efficient,vu2012improving,vu2010boosting} \\
\hhline{--}
\cellcolor{scorecolor!39.479759488010814}7.24 & KAKB & \makecell[l]{names} & S3OM & \makecell{---} & 2011 & \cite{allab2011constraint} \\
\hhline{--}
\cellcolor{scorecolor!25.250106256030445}4.63 & SSL-EC & \makecell[l]{Semi-Supervised Learning based on Exemplar Constraints} & --- & \makecell{Constraint Propagation} & 2012 & \cite{wang2012exemplars} \\
\hhline{--}
\cellcolor{scorecolor!31.271002019876104}5.74 & Cons-DBSCAN & \makecell[l]{Constrained - DBSCAN} & --- & \makecell{Hierarchical CC \& Density-based CC} & 2012 & \cite{zhao2012effective} \\
\hhline{--}
\cellcolor{scorecolor!47.01632678669627}8.62 & JDFD & \makecell[l]{names} & --- & \makecell{---} & 2013 & \cite{duarte2013constraint} \\
\hhline{--}
\cellcolor{scorecolor!47.08400980195582}8.64 & SACS & \makecell[l]{Sequential Approach for Constraint Selection} &  Xiang's, RCA, MPCK-Means & \makecell{---} & 2014 & \cite{abin2014active} \\
\hhline{--}
\cellcolor{scorecolor!34.542102128486505}6.34 & WAKL & \makecell[l]{names} & MPCK-Means & \makecell{---} & 2014 & \cite{atwa2014active} \\
\hhline{--}
\cellcolor{scorecolor!23.26645954589068}4.27 & MICS & \makecell[l]{Most InformativeConStraints} & RCA+K-Means & \makecell{---} & 2015 & \cite{yang2015selective} \\
\hhline{--}
\cellcolor{scorecolor!64.59066448782953}11.85 & CCCPYL & \makecell[l]{names} & DBSCAN, COP-K-Means, KKM, CSI, MPCK-Means & \makecell{---} & 2015 & \cite{chang2015active} \\
\hhline{--}
\cellcolor{scorecolor!36.143927015292896}6.63 & LCML & \makecell[l]{names} & MPCK-Means & \makecell{---} & 2016 & \cite{cai2016active} \\
\hhline{--}
\cellcolor{scorecolor!38.79483273258546}7.12 & AAA(18) & \makecell[l]{names} & MPCK-Means, RCA & \makecell{---} & 2018 & \cite{abin2016querying} \\
\hhline{--}
\cellcolor{scorecolor!32.07753865560921}5.88 & RWACS & \makecell[l]{Random Walk Approach to Constraints Selection} & MPCK-Means, RCA & \makecell{---} & 2018 & \cite{abin2017random} \\
\hhline{--}
\cellcolor{scorecolor!54.03519049540591}9.91 & AAA(19) & \makecell[l]{names} & MPCK-Means, RCA & \makecell{---} & 2019 & \cite{abin2019querying} \\
\hhline{--}
\cellcolor{scorecolor!100.0}18.34 & AIPC & \makecell[l]{Active Informative Pairwise Constraints algorithm} & PCK-Means & \makecell{---} & 2019 & \cite{zhong2019active} \\
\hhline{--}
\cellcolor{scorecolor!30.79822132733939}5.65 & AAVV & \makecell[l]{names} & MPCK-Means, RCA & \makecell{---} & 2020 & \cite{abin2020density} \\
\hhline{--}
\cellcolor{scorecolor!40.10369230091332}7.36 & ALPCS & \makecell[l]{Active Learning Pairwise Constraint based on Skeletons} & --- & \makecell{---} & 2020 & \cite{chen2020active} \\
\hhline{--}
\cellcolor{scorecolor!21.415053718073494}3.93 & ASCENT & \makecell[l]{---} & MPCK-Means & \makecell{---} & 2020 & \cite{li2019ascent}  \\
        \hline
		\end{tabular}}
	\caption{Feature table for the ACC - Active Constraint Acquisition methods.}
	\label{tab:tax:active_constraint_acquisition}
\end{table}

\subsubsection{Active Clustering with Constraints}

In active clustering with constraints, a CC procedure and a constraint generation method are applied alternately. The immediate result of these methods are both a partition of the dataset and a constraint set. These methods usually start by computing an unconstrained partition of the dataset. After this, some criteria are applied to select pairs of instances to query the oracle on the basis of the obtained partition. The answers to these queries are used to generate and save new constraints, which are later used to generate a new partition of the dataset by means of a CC method. Active clustering with constraints methods iterate these steps to produce the final constraint set and the partition. The active constraint generation method can be dependent of the CC method used to produce partitions, in which case they cannot be used separately. On the other hand, some active constraint generation methods are designed to be paired with any CC algorithm.

Table~\ref{tab:tax:active_clustering_with_constraints} gathers a list of active clustering with constraints methods. A major trend in this category is found in the use of the uncertainty of instances to rank and pair them to select the more uncertain ones and query them to the oracle. The uncertainty is always computed based in the current partition and is usually defined as the probability of an instance belonging to different known clusters. Some methods in this category are: RHWL, IU-Red, ALCSSC, CMKIPCM, AAA, URASC, A-COBS, ADP and ADPE. There are as well other criteria to select pairs of instances to query, such as the utility maximization (SRBR), the maximum expected error reduction (ASC), the partition change maximization (Active-HACC), the ensemble consensus (PT), or the classic informativeness and coherence (A-ITML-K-Means). Cluster-related criteria can also be used to select queries, such as the size and distance between the clusters (CAC, COBRA) or how well defined the frontiers are between them (AFCC, CVQE+). Paradigm-specific criteria are used by some methods, such as CMVNMF, which performs multi-view clustering and selects pairs of instances to query, based on intra-view and inter-view criteria, or AC-CF-tree and COBRAS, which use queries to determine the best cluster merge to perform in hierarchical CC. Similarly, the family of active FIECE-EM use concepts related to the population of individuals it maintains to select the best instances to query.

\begin{table}[!h]
	\centering
	\setlength{\tabcolsep}{7pt}
	\renewcommand{\arraystretch}{1.3}
	\resizebox{\textwidth}{!}{
		\begin{tabular}{c l l c c c c c c}
		\hline
		$S_{\mathcal{A}}$ & Acronym & Full Name & Penalty & ML & CL & Hybrid & Year & Ref. \\
		\hline
        \cellcolor{scorecolor!33.72171098676272}7.14 & RHWL & \makecell[l]{names} & $\times$ & Soft & Soft & \makecell{Probabilistic Clustering} & 2007 & \cite{huang2007semi} \\
\hhline{--}
\cellcolor{scorecolor!35.98312227137023}7.62 & PT & \makecell[l]{Penta-Training} & $\times$ & Soft & Soft & \makecell{Constrained Pool Generation} & 2008 & \cite{domeniconi2008penta} \\
\hhline{--}
\cellcolor{scorecolor!44.829047826412975}9.50 & AFCC & \makecell[l]{Active Fuzzy Constrained Clustering} & $\times$ & Soft & Soft & \makecell{Fuzzy CC} & 2008 & \cite{grira2008active} \\
\hhline{--}
\cellcolor{scorecolor!63.59989447810981}13.47 & ASC & \makecell[l]{Active Spectral Clustering} & --- & Any & Any & \makecell{Non Graph-based} & 2010 & \cite{wang2010active} \\
\hhline{--}
\cellcolor{scorecolor!19.67176139851519}4.17 & CAC1 & \makecell[l]{Constrained Active Clustering 1} & --- & Hard & Soft & \makecell{Hierarchical CC} & 2011 & \cite{biswas2011large} \\
\hhline{--}
\cellcolor{scorecolor!18.41365488982534}3.90 & IU-Red & \makecell[l]{Iterative Uncertainty Reduction} & $\times$ & Soft & Soft & \makecell{Non Graph-based} & 2012 & \cite{wauthier2012active} \\
\hhline{--}
\cellcolor{scorecolor!37.1009139019095}7.86 & SRBR & \makecell[l]{names} & $\times$ & Soft & Soft & \makecell{---} & 2013 & \cite{raj2013incremental} \\
\hhline{--}
\cellcolor{scorecolor!26.521648419834737}5.62 & A-ITML-K-Means & \makecell[l]{Active - Information Theoric Metric Learning - K-Means} & $\times$ & Soft & Soft & \makecell{Constraint Propagation} & 2013 & \cite{rao2013semi} \\
\hhline{--}
\cellcolor{scorecolor!42.89190659768211}9.09 & ALCSSC & \makecell[l]{Active Learning of Constraints for Semi-Supervised Clustering} & $\times$ & Soft & Soft & \makecell{---} & 2014 & \cite{xiong2013active} \\
\hhline{--}
\cellcolor{scorecolor!11.869587377737483}2.51 & AC-CF-tree & \makecell[l]{Active Constrained - Clustering Feature - tree} & \checkmark & Soft & Soft & \makecell{Penalty-based Methods} & 2014 & \cite{lai2014new} \\
\hhline{--}
\cellcolor{scorecolor!16.854579594095203}3.57 & Active-HACC & \makecell[l]{Active - Hierarchical Agglomerative Constrained Clustering} & $\times$ & Soft & Soft & \makecell{Hierarchical CC} & 2014 & \cite{biswas2014active} \\
\hhline{--}
\cellcolor{scorecolor!79.13553791008631}16.77 & CMKIPCM & \makecell[l]{Constrained Multiple Kernels Improved Possibilistic C-Means} & $\times$ & Soft & Soft & \makecell{Fuzzy CC} & 2015 & \cite{abin2015active} \\
\hhline{--}
\cellcolor{scorecolor!100.0}21.19 & AAA & \makecell[l]{names} & $\times$ & Soft & Soft & \makecell{Fuzzy CC} & 2016 & \cite{abin2016clustering} \\
\hhline{--}
\cellcolor{scorecolor!19.808044534326434}4.20 & COBRA & \makecell[l]{COnstraint-Based Repeated Aggregation} & $\times$ & Soft & Soft & \makecell{Hierarchical CC} & 2017 & \cite{van2017cobra} \\
\hhline{--}
\cellcolor{scorecolor!31.410227247045437}6.65 & URASC & \makecell[l]{Uncertainty Reducing Active Spectral Clustering} & $\times$ & Soft & Soft & \makecell{Non Graph-based} & 2017 & \cite{xiong2017active} \\
\hhline{--}
\cellcolor{scorecolor!41.34085257081233}8.76 & A-COBS & \makecell[l]{Active - COnstraint-Based Selection} & $\times$ & Soft & Soft & \makecell{Constrained Consensus} & 2017 & \cite{van2017constraint} \\
\hhline{--}
\cellcolor{scorecolor!11.645541014965023}2.47 & CVQE+ & \makecell[l]{Constrained Vector Quantization Error +} & $\times$ & Soft & Soft & \makecell{Penalty-based Methods} & 2018 & \cite{mai2018scalable} \\
\hhline{--}
\cellcolor{scorecolor!92.36871561620573}19.57 & COBRAS & \makecell[l]{COnstraint-Based Repeated Aggregation and Splitting} & $\times$ & Soft & Soft & \makecell{Hierarchical CC} & 2018 & \cite{craenendonck2018cobras} \\
\hhline{--}
\cellcolor{scorecolor!13.452196057867072}2.85 & CMVNMF & \makecell[l]{Constrained Multi-View NMF} & $\times$ & Soft & Soft & \makecell{Non-negative Matrix Factorization CC \& Inter-View Constrained} & 2018 & \cite{zong2018multi,zhang2015constrained} \\
\hhline{--}
\cellcolor{scorecolor!73.44449753658648}15.56 & FIECE-EM+BFCU & \makecell[l]{FIECE-EM + Best Feasible Classification Uncertainty} & $\times$ & Hard & Hard & \makecell{Genetic Algorithm \& Mixture Model-based CC} & 2020 & \cite{fernandes2019active,fernandes2020improving} \\
\hhline{--}
\cellcolor{scorecolor!73.44449753658648}15.56 & FIECE-EM+FCU & \makecell[l]{FIECE-EM + Feasible Classification Uncertainty} & $\times$ & Hard & Hard & \makecell{Genetic Algorithm \& Mixture Model-based CC} & 2020 & \cite{fernandes2019active,fernandes2020improving} \\
\hhline{--}
\cellcolor{scorecolor!73.44449753658648}15.56 & FIECE-EM+DVO & \makecell[l]{FIECE-EM + Distance to Violated Objects} & $\times$ & Hard & Hard & \makecell{Genetic Algorithm \& Mixture Model-based CC} & 2020 & \cite{fernandes2019active,fernandes2020improving} \\
\hhline{--}
\cellcolor{scorecolor!73.44449753658648}15.56 & FIECE-EM+LUC & \makecell[l]{FIECE-EM + Largest Unlabeled Clusters} & $\times$ & Hard & Hard & \makecell{Genetic Algorithm \& Mixture Model-based CC} & 2020 & \cite{fernandes2019active,fernandes2020improving} \\
\hhline{--}
\cellcolor{scorecolor!81.28876072163092}17.22 & ADPE & \makecell[l]{Active Density Peak Ensemble} & $\times$ & Soft & Soft & \makecell{Density-based CC \& Constrained Pool Generation} & 2021 & \cite{shi2021fast} \\
\hhline{--}
\cellcolor{scorecolor!84.22595815896753}17.84 & ADP & \makecell[l]{Active Density Peak} & $\times$ & Soft & Soft & \makecell{Density-based CC} & 2021 & \cite{shi2021fast}  \\
        \hline
		\end{tabular}}
	\caption{Feature table for the ACC - Active Clustering with Constraints methods.}
	\label{tab:tax:active_clustering_with_constraints}
\end{table}

\subsection{Neural Network-based CC}

Neural Networks (NN) are universal approximators which have been applied in many machine learning tasks, and CC is not an exception. Neural Network-based CC (NNbCC) tackles the CC from the NN perspective in three different ways: through self organizing maps, through deep-embeded clustering and through classic neural networks architectures.

\subsubsection{Self Organizing Maps-based CC}

Self organizing maps are NN (usually with fixed topology) whose neurons modify their position in the solution space to organize themselves according to the shape of the clusters. The result is a net whose neurons are grouped in clusters, which can be used to determine the cluster every instance belongs to. Constraints can be included into this process in different ways, as Table~\ref{tab:tax:cc_through_SOM} shows. Some on them, like SS-FKCN, simply use a penalty term accounting for violated constraints in a classic SOM variant. Others such as PrTM and SSGSOM, reformulate the classic SOM problem and use multiple neuron layers, forcing instances to flow through these to layers to be ultimately assigned to the appropiate cluster. In order to do so, PrTM uses constraint-influenced probabilities to decide how the position of the neurons changes, while SSGSOM adjusts the between-layer weights and the number of nodes of the first layer to dynamically correct the violation of constraints. Simpler methods like the S3OM, modify classic SOM for it to carry out only assignations without violating any constraints, similarly to COP-K-Means. 

\begin{table}[!h]
	\centering
	\setlength{\tabcolsep}{7pt}
	\renewcommand{\arraystretch}{1.3}
	\resizebox{\textwidth}{!}{
		\begin{tabular}{c l l c c c c c c}
		\hline
		$S_{\mathcal{A}}$ & Acronym & Full Name & Penalty & ML & CL & Hybrid & Year & Ref. \\
		\hline
        \cellcolor{scorecolor!53.44459371670725}5.74 & SS-FKCN & \makecell[l]{Semi-Supervised Fuzzy Kohonen Clustering Network} & \checkmark & Soft & Soft & \makecell{---} & 2006 & \cite{maraziotis2006semi} \\
\hhline{--}
\cellcolor{scorecolor!94.23660694504457}10.13 & PrTM & \makecell[l]{Probabilistic Topographic Mapping} & $\times$ & Soft & Soft & \makecell{---} & 2009 & \cite{benabdeslem2009probabilistic} \\
\hhline{--}
\cellcolor{scorecolor!71.61896338123647}7.70 & S3OM & \makecell[l]{Semi-Supervised Self Organizing Map} & $\times$ & Hard & Hard & \makecell{---} & 2011 & \cite{allab2011constraint} \\
\hhline{--}
\cellcolor{scorecolor!100.0}10.74 & CS2GS & \makecell[l]{Constrained Semi-Supervised Growing SOM} & $\times$ & Soft & Soft & \makecell{Online CC} & 2015 & \cite{allahyar2015constrained}  \\
        \hline
		\end{tabular}}
	\caption{Feature table for the NNbCC - Self Organizing Maps-based CC methods.}
	\label{tab:tax:cc_through_SOM}
\end{table}

The main difference between SOM-based approaches to CC and the other two approaches (deep embedded clustering and classic neural networks) is that the primary goal of the former is to produce a partition of the dataset, while the latter's is to cast predictions over unseen instances regarding the cluster they belong to. Please note that a partition can be obtained with deep embedded clustering and classic neural networks by feeding the training instances to the trained model. 

\subsubsection{Deep Embedded Clustering-based CC}

In deep embeded clustering-based CC constraints are included into the classic Deep Embeded Clustering (DEC) model. Table~\ref{tab:tax:cc_through_deep_embeded} gathers methods which use this approach. SDEC includes constraints into the classic DEC model by using constraints to influence its distance learning step, DCC does so by simply modifying the loss function of DEC with a penalty term. CDEC uses DEC to initialize its encoder, which is finally retrained to finally assign instances to clusters and satisfy the constraints.

\begin{table}[!h]
	\centering
	\setlength{\tabcolsep}{7pt}
	\renewcommand{\arraystretch}{1.3}
	\resizebox{\textwidth}{!}{
		\begin{tabular}{c l l c c c c c c}
		\hline
		$S_{\mathcal{A}}$ & Acronym & Full Name & Penalty & ML & CL & Hybrid & Year & Ref. \\
		\hline
        \cellcolor{scorecolor!80.73402311466913}8.32 & SDEC & \makecell[l]{Semi-supervised Deep Embedded Clustering} & $\times$ & Soft & Soft & \makecell{---} & 2019 & \cite{ren2019semi} \\
\hhline{--}
\cellcolor{scorecolor!100.0}10.31 & DCC & \makecell[l]{Deep Constrained Clustering} & $\times$ & Soft & Soft & \makecell{---} & 2020 & \cite{zhang2019framework,zhang2021framework} \\
\hhline{--}
\cellcolor{scorecolor!23.674440628196443}2.44 & CDEC & \makecell[l]{Constrained Deep Embedded Clustering} & $\times$ & Soft & Soft & \makecell{---} & 2021 & \cite{amirizadeh2021cdec}  \\
        \hline
		\end{tabular}}
	\caption{Feature table for the NNbCC - Deep Embeded Clustering-based CC methods.}
	\label{tab:tax:cc_through_deep_embeded}
\end{table}

\subsubsection{Classic Neural Network-based CC}

Lastly, classic neural network-based CC methods are presented in Table~\ref{tab:tax:cc_through_classic_nn}. Some of them, such as S$^3$C$^2$ and CDC, use the siamese neural networks, as they are known, to solve the CC problems in two steps. In the case of S$^3$C$^2$, the siamese neural network is used to solve the two steps in which the CC is decomposed into simpler binary problems, while CDC uses the siamese neural network to perform unsupervised clustering and a triple NN to perform CC. NN-EVCLUS simply implements EVCLUS in an NN setup and uses a penalty term in its loss function to include constraints.

SNNs consist of a NN model designed to learn non-linear similarity measures from pairwise constraints and to generalize the learned criterion to new data pairs. A SNN is a feedforward multi-layer perceptron whose learning set is defined as triplets composed of two instances and the constraint set between them, using 1 for ML and 0 for CL. In other words, ML instances have an associated target equal to 1, while CL instances have an associated target equal to 0. From the architectural point of view, the SNN has an input layer which accepts pairs of instances, a single hidden layer which contains an even number of units, and an output neuron with sigmoidal activation. The training of the SNN can be performed using the standard backpropagation scheme. Since the metric learned by an SNN cannot be straightforwardly used by a K-Means style algorithm, as the centroids do not necessarily have to be found in the dataset, centroids computation can be embedded in the SNN by using a classic K-Means minimization scheme based on backpropagation. This scheme keeps the weights and biases of the trained SNN fixed and varies the centroid coordinates (seen as free parameters). This is equivalent to redefining the original SNN model by adding a new layer to the network structure whose neuron activation functions correspond to the identity mapping.

\begin{table}[!h]
	\centering
	\setlength{\tabcolsep}{7pt}
	\renewcommand{\arraystretch}{1.3}
	\resizebox{\textwidth}{!}{
		\begin{tabular}{c l l c c c c c c}
		\hline
		$S_{\mathcal{A}}$ & Acronym & Full Name & Penalty & ML & CL & Hybrid & Year & Ref. \\
		\hline
        \cellcolor{scorecolor!45.09972695393743}7.01 & SNN & \makecell[l]{Similarity Neural Networks} & $\times$ & Soft & Soft & \makecell{Constrained Distance Transformation} & 2012 & \cite{maggini2012learning} \\
\hhline{--}
\cellcolor{scorecolor!100.0}15.54 & S$^3$C$^2$ & \makecell[l]{Semi-Supervised Siamese Classifiers for Clustering} & $\times$ & Soft & Soft & \makecell{---} & 2020 & \cite{smieja2020classification} \\
\hhline{--}
\cellcolor{scorecolor!64.88499075040988}10.09 & CDC & \makecell[l]{Constrained Deep Clustering} & $\times$ & Soft & Soft & \makecell{---} & 2021 & \cite{cui2021maintaining} \\
\hhline{--}
\cellcolor{scorecolor!22.88031246685534}3.56 & NN-EVCLUS & \makecell[l]{Neural Network-based EVidential CLUSstering} & \checkmark & Soft & Soft & \makecell{---} & 2021 & \cite{denoeux2021nn}  \\
        \hline
		\end{tabular}}
	\caption{Feature table for the NNbCC - Classic Neural Network-based CC methods.}
	\label{tab:tax:cc_through_classic_nn}
\end{table}

\subsection{Ensemble CC}

Ensemble clustering methods usually perform clustering in two steps. (1) generating a pool of solutions, whose diversity depends on the method (or methods) used for the generation. (2) taking the pool of solutions as input and producing a single final solution by merging or selecting solutions from the pool. The function in charge of this procedure is called the consensus function. The application of ensemble-based clustering methods on the constrained clustering problem gives place to a new distinction within this category, which classifies Ensemble CC (ECC) methods depending on the step (or steps) in which they consider constraints.

\subsubsection{Constrained Pool Generation}

These ensemble methods use constraints in the pool generation step (the first step), i.e.: the partitions in the pool of solutions are generated with CC methods. Table~\ref{tab:tax:constrained_pool} presents the list of methods belonging to this category. The consensus functions used by these methods do not take constraints into account. Therefore they are not considered in any distinction made within this category. The most commonly used consensus functions are majority voting, NCuts and CSPA.

Many methods use the subspace technique, which consists of performing clustering in a new space with a lower number of dimensions than the original space. The technique used to produce different subspaces introduces variability on the pool of solutions. The most common procedure used to generate the subspaces is simply a random sampling of the original features, such as in SCSC, ISSCE, RSSCE, CESCP, DCECP or ADPE. However, there are subspace generation methods specifically designed for certain algorithms. An example of this is SMCE, which uses the CSI method to project instances and constraints into multiple low-dimensional subspaces and then learning positive semi-definite matrices therein.

Other methods simply use any previous CC algorithm to produce the pool. The most common way to introduce diversity in the pool is by applying different CC methods to produce different partitions. Methods that use this strategy are SCEV, MVSCE, E$^2$CPE, HSCE or FQH. Another (and less used) method to generate diversity is varying the hyperparameters of a single CC method, as in Samarah.

\begin{table}[!h]
	\centering
	\setlength{\tabcolsep}{7pt}
	\renewcommand{\arraystretch}{1.3}
	\resizebox{\textwidth}{!}{
		\begin{tabular}{c l l c c c c c c}
		\hline
		$S_{\mathcal{A}}$ & Acronym & Full Name & Penalty & ML & CL & Hybrid & Year & Ref. \\
		\hline
        \cellcolor{scorecolor!23.129952404858077}5.33 & SMCE & \makecell[l]{Subspace Metric Cluster Ensemble} & $\times$ & Soft & Soft & \makecell{---} & 2006 & \cite{yan2006subspace} \\
\hhline{--}
\cellcolor{scorecolor!33.06805150487164}7.62 & PT & \makecell[l]{Penta-Training} & $\times$ & Soft & Soft & \makecell{Active Clustering with Constraints} & 2008 & \cite{domeniconi2008penta} \\
\hhline{--}
\cellcolor{scorecolor!46.908073507319365}10.81 & Samarah & \makecell[l]{---} & $\times$ & Soft & Soft & \makecell{---} & 2010 & \cite{forestier2010collaborative} \\
\hhline{--}
\cellcolor{scorecolor!0.0}0.00 & SCEV & \makecell[l]{Semi-supervised Clustering Ensemble by Voting} & $\times$ & Soft & Soft & \makecell{---} & 2012 & \cite{iqbal2012semi} \\
\hhline{--}
\cellcolor{scorecolor!47.94654974026017}11.05 & MVSCE & \makecell[l]{Majority Voting Semi-supervised Clustering Ensemble} & $\times$ & Soft & Soft & \makecell{---} & 2013 & \cite{chen2013convergence} \\
\hhline{--}
\cellcolor{scorecolor!5.405405405405404}1.25 & E$^2$CPE & \makecell[l]{Exhaustive and Efficient Constraint Propagation Ensemble} & $\times$ & Soft & Soft & \makecell{---} & 2013 & \cite{lu2013exhaustive} \\
\hhline{--}
\cellcolor{scorecolor!23.785488701359426}5.48 & SCSC & \makecell[l]{Semi-supervised Clustering with Sequential Constraints} & $\times$ & Soft & Soft & \makecell{Online CC} & 2015 & \cite{yi2015efficient} \\
\hhline{--}
\cellcolor{scorecolor!54.07576836941489}12.47 & HSCE & \makecell[l]{Hybrid Semi-supervised Clustering Ensemble} & $\times$ & Soft & Soft & \makecell{Non Graph-based \& Penalty-based Methods} & 2015 & \cite{wei2015semi,wei2018combined} \\
\hhline{--}
\cellcolor{scorecolor!100.0}23.05 & ISSCE & \makecell[l]{Incremental Semi-Supervised Clustering Ensemble} & $\times$ & Soft & Soft & \makecell{Constraint Propagation} & 2016 & \cite{yu2016incremental} \\
\hhline{--}
\cellcolor{scorecolor!87.83783783783782}20.25 & RSSCE & \makecell[l]{Random Subspace based Semi-supervised Clustering Ensemble} & $\times$ & Soft & Soft & \makecell{Constraint Propagation} & 2016 & \cite{yu2016incremental} \\
\hhline{--}
\cellcolor{scorecolor!26.436498131816016}6.09 & FQH & \makecell[l]{names} & $\times$ & Soft & Soft & \makecell{---} & 2017 & \cite{yang2017cluster} \\
\hhline{--}
\cellcolor{scorecolor!33.9815192725249}7.83 & CESCP & \makecell[l]{Clustering Ensemble based on Selected Constraint Projection } & $\times$ & Soft & Soft & \makecell{Constraint Propagation} & 2018 & \cite{yu2018semi} \\
\hhline{--}
\cellcolor{scorecolor!46.14368143468705}10.64 & DCECP & \makecell[l]{Double-weighting Clustering Ensemble with Constraint Projection} & $\times$ & Soft & Soft & \makecell{Constraint Propagation} & 2018 & \cite{yu2018semi} \\
\hhline{--}
\cellcolor{scorecolor!74.70338193661475}17.22 & ADPE & \makecell[l]{Active Density Peak Ensemble} & $\times$ & Soft & Soft & \makecell{Density-based CC \& Active Clustering with Constraints} & 2021 & \cite{shi2021fast}  \\
        \hline
		\end{tabular}}
	\caption{Feature table for the ECC - Constrained Pool Generation methods.}
	\label{tab:tax:constrained_pool}
\end{table}

\subsubsection{Constrained Consensus}

In constrained consensus ensemble methods, constraints are used only in the consensus function to produce a final partition meeting as much constraints as possible. Table~\ref{tab:tax:constrained_consensus} gathers the four methods which belong to this category. All of these methods generate the partitions in the pool by means of classic clustering algorithms. This is why the consensus functions used by these methods usually measure the quality of the generated solutions with respect to the constraints by means of a quality index and select the best ones to be finally merged. For example COBS and A-COBS use the infeasibility to select the best partition in the pool, which is generated by any classic clustering method. WECR K-Means runs classic K-Means multiple times with different hyperparameters to generate the pool, then a weighting procedure is used to automatically assign a weight to every partition depending on their local and global quality, which includes the infeasibility. A weighted co-association matrix based consensus approach is then applied to achieve a final partition. Semi-MultiCons builds a tree-like pool or partitions and then applies a normalized score which measures constraint satisfaction if any given merge or split operation between clusters is performed in the tree.

\begin{table}[!h]
	\centering
	\setlength{\tabcolsep}{7pt}
	\renewcommand{\arraystretch}{1.3}
	\resizebox{\textwidth}{!}{
		\begin{tabular}{c l l c c c c c c}
		\hline
		$S_{\mathcal{A}}$ & Acronym & Full Name & Penalty & ML & CL & Hybrid & Year & Ref. \\
		\hline
        \cellcolor{scorecolor!79.95228381331353}7.00 & COBS & \makecell[l]{COnstraint-Based Selection} & $\times$ & Soft & Soft & \makecell{---} & 2017 & \cite{van2017constraint} \\
\hhline{--}
\cellcolor{scorecolor!100.0}8.76 & A-COBS & \makecell[l]{Active - COnstraint-Based Selection} & $\times$ & Soft & Soft & \makecell{Active Clustering with Constraints} & 2017 & \cite{van2017constraint} \\
\hhline{--}
\cellcolor{scorecolor!76.25813964509804}6.68 & WECR K-Means & \makecell[l]{WEighted Consensus of Random K-Means ensemble} & $\times$ & Soft & Soft & \makecell{---} & 2021 & \cite{lai2019adaptive} \\
\hhline{--}
\cellcolor{scorecolor!61.52508074996228}5.39 & Semi-MultiCons & \makecell[l]{Semi-supervised Multiple Consensus} & $\times$ & Soft & Soft & \makecell{---} & 2022 & \cite{yang2022semi}  \\
        \hline
		\end{tabular}}
	\caption{Feature table for the ECC - Constrained Consensus methods.}
	\label{tab:tax:constrained_consensus}
\end{table}

\subsubsection{Full Constrained}

These methods (in Table~\ref{tab:tax:full_constrained}) include constraints in both the pool generation and the consensus steps. They make use of the formulas described before and combine them. On the one hand, SFS$^3$EC, ARSCE and RSEMICE make use of the subspace technique, although they differ in the consensus function. SFS$^3$EC merges partitions in the pool by building a hypergraph which takes partitions and constraints into account and running METIS over this graph to get the final partition. ARSCE computes the affinity graph for every solution in the pool, and uses regularized ensemble diffusion to fuse the similarity information. Finally, RSEMICE assigns a confidence factor to each solution in the pool to build a consensus matrix, which can be interpreted as a graph over which the NCut algorithm is applied (used as the consensus function). On the other hand, COP-SOM-E and  Cop-EAC-SL use previous CC methods (ICOP-K-Means and  COP-K-Means, respectively) to generate their pool.  COP-SOM-E uses a hard constrained version of SOM as the consensus matrix, and Cop-EAC-SL runs the constrained single-link algorithm over a co-association matrix which counts how many times pairs of instances are placed in the same cluster in different partitions. 

\begin{table}[!h]
	\centering
	\setlength{\tabcolsep}{7pt}
	\renewcommand{\arraystretch}{1.3}
	\resizebox{\textwidth}{!}{
		\begin{tabular}{c l l c c c c c c}
		\hline
		$S_{\mathcal{A}}$ & Acronym & Full Name & Penalty & ML & CL & Hybrid & Year & Ref. \\
		\hline
        \cellcolor{scorecolor!42.28838779284646}6.36 & Cop-EAC-SL & \makecell[l]{Constrained partitional Evicende ACcumulation Single Link} & $\times$ & Soft & Soft & \makecell{---} & 2009 & \cite{abdala2009evidence} \\
\hhline{--}
\cellcolor{scorecolor!19.96227896478599}3.00 & En-Ant & \makecell[l]{Ensemble Ant} & $\times$ & Soft & Soft & \makecell{Swarm Optimization} & 2012 & \cite{yang2012semi} \\
\hhline{--}
\cellcolor{scorecolor!50.15882952363759}7.54 & COP-SOM-E & \makecell[l]{COnstrained Partitional - Self Organazing Map - Ensemble} & $\times$ & Hard & Hard & \makecell{---} & 2012 & \cite{yang2012consensus} \\
\hhline{--}
\cellcolor{scorecolor!71.61536551889023}10.77 & RSEMICE & \makecell[l]{Random subspace based SEMI-supervised Clustering Ensemble framework} & \checkmark & Soft & Soft & \makecell{---} & 2017 & \cite{yu2017adaptive} \\
\hhline{--}
\cellcolor{scorecolor!100.0}15.04 & SFS$^3$EC & \makecell[l]{Stratified Feature Sampling for Semi-Supervised Ensemble Clustering} & $\times$ & Soft & Soft & \makecell{Non Graph-based} & 2019 & \cite{tian2019stratified} \\
\hhline{--}
\cellcolor{scorecolor!45.632433086917274}6.86 & ARSCE & \makecell[l]{Adaptive Regularized Semi-supervised Clustering Ensemble} & $\times$ & Soft & Soft & \makecell{---} & 2020 & \cite{luo2019adaptive}  \\
        \hline
		\end{tabular}}
	\caption{Feature table for the ECC - Full Constrained methods.}
	\label{tab:tax:full_constrained}
\end{table}

\subsection{Metaheuristics-based CC}

Metaheuristics-based CC (MbCC) use metaheuristic algorithms to approach the CC problem. Many distinctions can be made within the metaheuristic algorithms field, in this study the trajectory-based methods versus population-based methods is used to produce to subcategories of CC approaches, as it is the one which results in the more consistent dichotomy.

\subsubsection{Population-based}

A plethora of metaheuristic methods has been applied to the CC problem. Particularly, population-based methods have shown remarkable success, with evolutive algorithms being the most used ones. A further distinction can be made within these methods: swarm optimization algorithms and genetic algorithms. In swarm optimization algorithms, a population of individuals is used to mimic the behavior of a colony of insects in its natural environment, while in genetic algorithms, the population is evolved according to the rules of natural selection, expecting them to generate the best possible individual (solution). 

\paragraph{Swarm Optimization} Table~\ref{tab:tax:MbCC_swarm_opt} gathers swarm optimization algorithms which tackle the CC problem. All of them are based on ant colonies behavior, with the main differences found in the scheme used to include constraints. MCLA, MELA and CELA are all based in the Leader Ant algorithm and use the same integration scheme. They modify the ant-nest assignment rule so that only feasible assignments are taken into account. MCLA and MELA ensure that they do not violate any ML constraints by using chunklets. They only differ in the type of constraints they can handle. CAC is based on the RWAC algorithm, which tries to simulate the behavior of ants in their environment trying to find a place to sleep. Constraints are included in this scheme by modifying attractive and repulsive forces between ants associated to constrained instances. The En-Ant algorithm uses three instances of the Semi-Ant algorithm as the partition generation algorithm in an ensemble setup. Constraints are used both in the generation of the partition pool and in the consensus function. Please note how none of these algorithms modify the fitness function of the ant colony algorithm on which they are based, instead they include constraints by modifying other aspects of the algorithms.

\begin{table}[!h]
	\centering
	\setlength{\tabcolsep}{7pt}
	\renewcommand{\arraystretch}{1.3}
	\resizebox{\textwidth}{!}{
		\begin{tabular}{c l l c c c c c c}
		\hline
		$S_{\mathcal{A}}$ & Acronym & Full Name & Penalty & ML & CL & Hybrid & Year & Ref. \\
		\hline
        \cellcolor{scorecolor!49.466063824232656}6.26 & MCLA & \makecell[l]{Must-link Cannot-link Leader Ant} & $\times$ & Hard & Hard & \makecell{---} & 2009 & \cite{vu2009leader} \\
\hhline{--}
\cellcolor{scorecolor!100.0}12.66 & MELA & \makecell[l]{Must-link $\epsilon$-link Leader Ant} & $\times$ & Hard & --- & \makecell{---} & 2009 & \cite{vu2009leader} \\
\hhline{--}
\cellcolor{scorecolor!100.0}12.66 & CELA & \makecell[l]{Cannot-link $\epsilon$-link Leader Ant} & $\times$ & --- & Hard & \makecell{---} & 2009 & \cite{vu2009leader} \\
\hhline{--}
\cellcolor{scorecolor!60.9139351428599}7.71 & CAC & \makecell[l]{Constrained Ant Clustering} & $\times$ & Soft & Soft & \makecell{---} & 2012 & \cite{xu2011constrained,xu2013improving} \\
\hhline{--}
\cellcolor{scorecolor!23.716936949696414}3.00 & En-Ant & \makecell[l]{Ensemble Ant} & $\times$ & Soft & Soft & \makecell{Full Constrained} & 2012 & \cite{yang2012semi}  \\
        \hline
		\end{tabular}}
	\caption{Feature table for the MbCC - Population-based - Swarm Optimization methods.}
	\label{tab:tax:MbCC_swarm_opt}
\end{table}

\paragraph{Genetic Algorithm} Genetic strategies used to tackle the CC problems are presented in Table~\ref{tab:tax:MbCC_genetic}. A low-level dichotomy can be made within these methods, they can be either single-objective or multi-objective genetic algorithms. On the one hand, multi-objective algorithms optimize a set of fitness functions all at the same time. Constraints can be naturally included in this paradigm by simply adding the infeasibility as one of the functions to be optimized. This is the case for MOCK (which includes constraints into the classic PESA-II algorithm), PCS (which is based on NSGAII) and ME-MOEA/D$_{CC}$ (which modifies classic MOEA/D). These three proposals also modify a basic aspect of the algorithm they are based on for it to fit better to the CC problem. MOCK implements a constraint-oriented initialization scheme. PCS features an improving procedure applied to the population after the classic operators have been applied. Lastly, ME-MOEA/D$_{CC}$ uses memetic elitism with controlled feedback. On the other hand, single objective genetic algorithms usually optimize a combination of any classic clustering related measure and the infeasibility included as a penalty term. Methods such as COP-HGA, BRKGA+LS or SHADE$_{CC}$ use this strategy. Other proposals, such as Cop-CGA and FIECE-EM, evolve separate populations or subpopulations which have individuals with different solutions qualities and make them interact to generate new individuals. FIECE-EM+BFCU, FIECE-EM+FCU, FIECE-EM+DVO and FIECE-EM+LUC are all active variants of FIECE-EM.

\begin{table}[!h]
	\centering
	\setlength{\tabcolsep}{7pt}
	\renewcommand{\arraystretch}{1.3}
	\resizebox{\textwidth}{!}{
		\begin{tabular}{c l l c c c c c c}
		\hline
		$S_{\mathcal{A}}$ & Acronym & Full Name & Penalty & ML & CL & Hybrid & Year & Ref. \\
		\hline
        \cellcolor{scorecolor!54.27622801955153}15.92 & MOCK & \makecell[l]{Multi Objective Clustering with automatic K-determination} & $\times$ & Soft & Soft & \makecell{---} & 2006 & \cite{handl2006semi} \\
\hhline{--}
\cellcolor{scorecolor!31.049593032279116}9.10 & COP-CGA & \makecell[l]{COnstrained Partitional - Clustering Genetic Algorithm} & $\times$ & Soft & Soft & \makecell{---} & 2008 & \cite{hong2008probabilistic} \\
\hhline{--}
\cellcolor{scorecolor!36.809840775526396}10.79 & COP-HGA & \makecell[l]{COnstrained Partitional - Hybrid Genetic Algorithm} & \checkmark & Soft & Soft & \makecell{---} & 2008 & \cite{hong2008genetic} \\
\hhline{--}
\cellcolor{scorecolor!14.705295235748522}4.31 & PSC & \makecell[l]{Pareto based multi objective algorithm for Semi-supervised Clustering} & $\times$ & Soft & Soft & \makecell{---} & 2012 & \cite{ebrahimi2012semi} \\
\hhline{--}
\cellcolor{scorecolor!14.937518509811607}4.38 & CEAC & \makecell[l]{Constrained Evolutionary Algorithm for Clustering} & $\times$ & Soft & Soft & \makecell{---} & 2016 & \cite{he2016evolutionary} \\
\hhline{--}
\cellcolor{scorecolor!46.87868891035089}13.75 & BRKGA+LS & \makecell[l]{Biased Random Key Genetic Algorithm + Local Search} & \checkmark & Soft & Soft & \makecell{---} & 2017 & \cite{de2017comparison} \\
\hhline{--}
\cellcolor{scorecolor!47.65650214535175}13.97 & FIECE-EM & \makecell[l]{Feasible-Infeasible Evolutionary Create \& Eliminate - Expectation Maximization} & $\times$ & Hard & Hard & \makecell{Mixture Model-based CC} & 2018 & \cite{covoes2018classification} \\
\hhline{--}
\cellcolor{scorecolor!53.063833214184605}15.56 & FIECE-EM+BFCU & \makecell[l]{FIECE-EM + Best Feasible Classification Uncertainty} & $\times$ & Hard & Hard & \makecell{Active Clustering with Constraints \& Mixture Model-based CC} & 2020 & \cite{fernandes2019active,fernandes2020improving} \\
\hhline{--}
\cellcolor{scorecolor!53.063833214184605}15.56 & FIECE-EM+FCU & \makecell[l]{FIECE-EM + Feasible Classification Uncertainty} & $\times$ & Hard & Hard & \makecell{Active Clustering with Constraints \& Mixture Model-based CC} & 2020 & \cite{fernandes2019active,fernandes2020improving} \\
\hhline{--}
\cellcolor{scorecolor!53.063833214184605}15.56 & FIECE-EM+DVO & \makecell[l]{FIECE-EM + Distance to Violated Objects} & $\times$ & Hard & Hard & \makecell{Active Clustering with Constraints \& Mixture Model-based CC} & 2020 & \cite{fernandes2019active,fernandes2020improving} \\
\hhline{--}
\cellcolor{scorecolor!53.063833214184605}15.56 & FIECE-EM+LUC & \makecell[l]{FIECE-EM + Largest Unlabeled Clusters} & $\times$ & Hard & Hard & \makecell{Active Clustering with Constraints \& Mixture Model-based CC} & 2020 & \cite{fernandes2019active,fernandes2020improving} \\
\hhline{--}
\cellcolor{scorecolor!100.0}29.32 & SHADE$_{CC}$ & \makecell[l]{Succes History-based Adaptive Differential Evolution - Constrained Clustering} & \checkmark & Soft & Soft & \makecell{---} & 2021 & \cite{gonzalez2021enhancing} \\
\hhline{--}
\cellcolor{scorecolor!96.25002720897776}28.22 & ME-MOEA/D$_{CC}$ & \makecell[l]{Memetic Elitist - Multiobjective Optimization Evolutionary Algorithm based on Decomposition - Constrained Clustering } & \checkmark & Soft & Soft & \makecell{---} & 2021 & \cite{gonzalez2021me}  \\
        \hline
		\end{tabular}}
	\caption{Feature table for the MbCC - Population-based - Genetic Algorithm methods.}
	\label{tab:tax:MbCC_genetic}
\end{table}

\subsubsection{Single Individual}

Single individual methods focus on modifying and improving a single candidate solution. They start with a single individual which is improved with respect to the fitness function. Simulated annealing, local search, iterated local search or guided local search are examples of single solution metaheuristics. Table~\ref{tab:tax:MbCC_single} lists methods which belong to this category. CCLS and DILS$_{\text{CC}}$ use both variants of the classic LS algorithm to find solutions for the CC problem. They use a combination of the intra-cluster mean distance and the infeasibility to build their fitness function. The SemiSync algorithm is a nature-inspired non-evolutive based on regarding instances as a set of constrained phase oscillators, whose dynamics can be simulated to build a partition. The local interaction of every oscillator with respect to its neighborhood can be computed over time. Therefore similar instances will synchronize together in groups that can be interpreted as clusters. ML and CL are included by introducing an additional global interaction term.

\begin{table}[!h]
	\centering
	\setlength{\tabcolsep}{7pt}
	\renewcommand{\arraystretch}{1.3}
	\resizebox{\textwidth}{!}{
		\begin{tabular}{c l l c c c c c c}
		\hline
		$S_{\mathcal{A}}$ & Acronym & Full Name & Penalty & ML & CL & Hybrid & Year & Ref. \\
		\hline
        \cellcolor{scorecolor!17.724302172210407}3.56 & PAST-Toss & \makecell[l]{Pick A Spanning Tree – Toss} & $\times$ & Soft & Soft & \makecell{Graph-based} & 2008 & \cite{coleman2008local} \\
\hhline{--}
\cellcolor{scorecolor!19.83248300361143}3.99 & CCLS & \makecell[l]{Constrained Clustering by Local Search} & \checkmark & Soft & Soft & \makecell{---} & 2016 & \cite{hiep2016local} \\
\hhline{--}
\cellcolor{scorecolor!57.24330852293574}11.50 & CG+PR+LS & \makecell[l]{Column Generation + Path Relinking + Local Search} & $\times$ & Soft & Soft & \makecell{Column Generation} & 2017 & \cite{de2017comparison} \\
\hhline{--}
\cellcolor{scorecolor!70.17675190827586}14.10 & SemiSync & \makecell[l]{---} & $\times$ & Soft & Soft & \makecell{---} & 2019 & \cite{zhang2019semisync} \\
\hhline{--}
\cellcolor{scorecolor!100.0}20.09 & DILS$_{CC}$ & \makecell[l]{Dual Iterative Local Search - Constrained Clustering} & \checkmark & Soft & Soft & \makecell{---} & 2020 & \cite{gonzalez2020dils}  \\
        \hline
		\end{tabular}}
	\caption{Feature table for the MbCC - Single individual methods.}
	\label{tab:tax:MbCC_single}
\end{table}

\subsection{Multi-View CC}

In many applications, data is collected from different sources in diverse domains, usually involving multiple feature collectors. This data refers to the same reality, although it exhibits heterogeneous properties, which translates into every instance being described by different sets of features. These are the called views, and the problem of performing clustering over instances described by different sets of features is known as multi-view (or multi-source) clustering~\cite{yang2018multi}. In multi-view clustering, different sources of the data are used to produce a single partition. Constraints can be included into multi-view clustering in different ways and levels, giving place to Multi-View Constrained Clustering (MVCC). With respecto to the level in which constraints can be used, two options are available: intra-view constraints and inter-view constraints. Intra-view constraints relate instances which belong to the same view of the data (similarly to constraints in any non MVCC algorithm), while inter-view constraints relate instances which belong to different views, hence encouraging collaboration between the clustering processes applied to them.

\subsubsection{Intra-View Constrained}

Intra-view CC usually performs clustering separately in each view and then tries to find a consensus between the obtained partitions (similarly to what ensemble clustering does with the consensus function). Methods which belong to this category are gathered in Table~\ref{tab:tax:MVCC_intra}. SMVC models clustering views via multivariate Bayesian mixture distributions located in subspace projections. It includes constraints in the Bayesian learning processes. TVClust and RDPM are both very particular methods, as they view the dataset and the constraint set as different sources of information for the same data, thus performing multi-view clustering with only two views. The dataset is modeled by a Dirichlet Process Mixture model and the constraint set is modeled by a random graph. They aggregate information from the two views through a Bayesian framework and they reach a consensus about the cluster structure though a Gibbs sampler. MVMC independently builds a pairwise similarity matrix for every view and casts the clustering task into a matrix completion problem based on the constraints and the feature information from multiple views. The final pairwise similarity matrix is built iteratively by approaching the independent pairwise similarity matrices in different views to each other. The final partition is obtained by performing spectral clustering of the final similarity matrix. SSCARD is based on classic CARD, which is able to combine multiple weighted sources of relational information (some may be more relevant than others) to produce a partition of the dataset. SSCARD simply includes the PCCA penalty term (see Section~\ref{subsec:tax:hcc}) into the classic CARD objective function. Lastly, MVCC is the only intra-view CC method that performs clustering in the different views in a collaborative way. It performs constrained clustering in each view separately, inferring new constraints and transferring them between views using partial mapping. For constraint inference and transfer, a variant of the co-EM algorithm~\cite{nigam2000analyzing} is used, which is an iterative EM based algorithm that learns a model from multiple views of the data.

\begin{table}[!h]
	\centering
	\setlength{\tabcolsep}{7pt}
	\renewcommand{\arraystretch}{1.3}
	\resizebox{\textwidth}{!}{
		\begin{tabular}{c l l c c c c c c}
		\hline
		$S_{\mathcal{A}}$ & Acronym & Full Name & Penalty & ML & CL & Hybrid & Year & Ref. \\
		\hline
        \cellcolor{scorecolor!35.43886552933381}7.09 & SSCARD & \makecell[l]{Semi-Supervised Clustering and Aggregating Relational Data} & \checkmark & Soft & Soft & \makecell{Fuzzy CC} & 2007 & \cite{frigui2007adaptive,frigui2008fuzzy} \\
\hhline{--}
\cellcolor{scorecolor!17.221466717657798}3.45 & MVCC & \makecell[l]{Multi-View Constrained Clustering} & $\times$ & Soft & Soft & \makecell{---} & 2014 & \cite{eaton2014multi} \\
\hhline{--}
\cellcolor{scorecolor!18.104389542244075}3.62 & SMVC & \makecell[l]{Semi-supervised Multi-View Clustering} & $\times$ & Soft & Soft & \makecell{---} & 2014 & \cite{gunnemann2014smvc} \\
\hhline{--}
\cellcolor{scorecolor!95.99778866921953}19.21 & TVClust & \makecell[l]{Two-Views Clustering} & $\times$ & Soft & Soft & \makecell{---} & 2015 & \cite{khashabi2015clustering} \\
\hhline{--}
\cellcolor{scorecolor!100.0}20.02 & RDPM & \makecell[l]{Relational Diritchlet Process Means} & $\times$ & Soft & Soft & \makecell{---} & 2015 & \cite{khashabi2015clustering} \\
\hhline{--}
\cellcolor{scorecolor!55.138427338904414}11.04 & MVMC & \makecell[l]{Multi-View Matrix Completion} & $\times$ & Soft & Soft & \makecell{Matrix Completion \& Spectral CC} & 2017 & \cite{zhao2017multi} \\
\hhline{--}
\cellcolor{scorecolor!90.78569938739375}18.17 & MVCSC & \makecell[l]{Multi-View Constrained Spectral Clustering} & $\times$ & Soft & Soft & \makecell{Non Graph-based} & 2019 & \cite{chen2019auto}  \\
        \hline
		\end{tabular}}
	\caption{Feature table for the MVCC - Intra-View Constrained methods.}
	\label{tab:tax:MVCC_intra}
\end{table}

\subsubsection{Inter-View Constrained}

Inter-View CC methods can handle both intra-view and inter-view constraints. Methods which belong to this category are presented in Table~\ref{tab:tax:MVCC_inter}. UCP uses the results of intra-view constraint propagation to adjust the similarity matrix of each view, and then performs inter-view constraint propagation with the adjusted similarity matrices. Its main drawback is that it is limited to two views. CMVNMF minimizes the loss function of NMF in each view, as well as the disagreement between each pair of views. The disagreement is defined as the difference between feature vectors associated to the same instance in the same view. It should be high if they are CL instances and low if they are ML instances. MSCP can propagate constraints across different data sources by dividing the problem into a series of two-source constraint propagation subproblems, which can be transformed into solving a Sylvester matrix equation, viewed as a generalization of the Lyapunov matrix equation. MCPCP uses a low-rank relation matrix to represent the pairwise constraints between instances from different views. Afterwards it learns the full relation matrix by using a matrix completion algorithm and derives an indicator matrix from it with an iterative optimization process.

\begin{table}[!h]
	\centering
	\setlength{\tabcolsep}{7pt}
	\renewcommand{\arraystretch}{1.3}
	\resizebox{\textwidth}{!}{
		\begin{tabular}{c l l c c c c c c}
		\hline
		$S_{\mathcal{A}}$ & Acronym & Full Name & Penalty & ML & CL & Hybrid & Year & Ref. \\
		\hline
        \cellcolor{scorecolor!58.91710829047315}5.27 & UCP & \makecell[l]{Unified Constraint Propagation} & $\times$ & Soft & Soft & \makecell{Constraint Propagation} & 2013 & \cite{lu2013unified} \\
\hhline{--}
\cellcolor{scorecolor!44.994234676837586}4.03 & MSCP & \makecell[l]{Multi-Source Constraint Propagation} & $\times$ & Soft & Soft & \makecell{Constraint Propagation} & 2013 & \cite{lu2013exhaustive} \\
\hhline{--}
\cellcolor{scorecolor!100.0}8.95 & MCPCP & \makecell[l]{Matrix Completion - Pairwise Constraint Propagation} & $\times$ & Soft & Soft & \makecell{Matrix Completion} & 2015 & \cite{yang2014matrix} \\
\hhline{--}
\cellcolor{scorecolor!31.842236108920353}2.85 & CMVNMF & \makecell[l]{Constrained Multi-View NMF} & $\times$ & Soft & Soft & \makecell{Non-negative Matrix Factorization CC \& Active Clustering with Constraints} & 2018 & \cite{zong2018multi,zhang2015constrained}  \\
        \hline
		\end{tabular}}
	\caption{Feature table for the MVCC - Inter-View Constrained methods.}
	\label{tab:tax:MVCC_inter}
\end{table}

\subsection{Kernel CC}

Kernel methods perform clustering by mapping the data from the original input space to a new feature space, which is usually of higher dimensionality. The key aspect of kernel-based methods is the avoidance of an explicit knowledge of the mapping function, which is achieved by computing dot products in the feature space via a kernel function. A critical aspect for kernel-based methods is the selection of the optimal kernel and its parameters. The basic classic kernel-based clustering method is the Kernel-K-Means algorithm. It performs clustering directly in the feature space by computing pairwise distances and updating centroids, using dot products and the kernel trick~\cite{domeniconi2011composite}. The goal of Kernel CC (KCC) is to learn a kernel function that maps ML instances close to each other in feature space, while mapping CL instances far apart, an then perform clustering in the feature space.

Table~\ref{tab:tax:kernel_cc} shows a list of KCC methods. The first KCC method can be found in SSKK~\cite{kulis2005semi,kulis2009semi}, which is built on the basis of the HMRF-K-Means method. Kernel CC methods are all very similar to each other, with one of the few differences being the way in which they include the kernel parameters in clustering process. Some methods, such as ASSKK or SFFA include this parameter in the optimization process. Therefore they do not need to be specified by the user, not withstanding the higher computational cost. Additionally, other methods use more than one kernel, such as TRAGEK and ENPAKL. They learn multiple kernels which are later combined in a single one to obtain the global kernel matrix. Some minor differences can be found in ssFS, for example, which is specifically designed to cluster sets of graphs.

\begin{table}[!h]
	\centering
	\setlength{\tabcolsep}{7pt}
	\renewcommand{\arraystretch}{1.3}
	\resizebox{\textwidth}{!}{
		\begin{tabular}{c l l c c c c c c}
		\hline
		$S_{\mathcal{A}}$ & Acronym & Full Name & Penalty & ML & CL & Hybrid & Year & Ref. \\
		\hline
        \cellcolor{scorecolor!100.0}20.98 & SSKK & \makecell[l]{Semi-Supervised Kernel K-Means} & $\times$ & Soft & Soft & \makecell{---} & 2005 & \cite{kulis2005semi,kulis2009semi} \\
\hhline{--}
\cellcolor{scorecolor!29.74199685173884}6.24 & ASSKK & \makecell[l]{Adaptive Semi Supervised Kernel K-Means} & $\times$ & Soft & Soft & \makecell{---} & 2006 & \cite{yan2006adaptive} \\
\hhline{--}
\cellcolor{scorecolor!47.18092139671003}9.90 & BoostCluster & \makecell[l]{---} & $\times$ & Soft & Soft & \makecell{---} & 2007 & \cite{liu2007boostcluster} \\
\hhline{--}
\cellcolor{scorecolor!75.38584059302674}15.81 & CCSKL & \makecell[l]{Constrained Clustering by Spectral Kernel Learning} & $\times$ & Soft & Soft & \makecell{Non Graph-based} & 2009 & \cite{li2009constrained2} \\
\hhline{--}
\cellcolor{scorecolor!11.28317196764987}2.37 & SFFA & \makecell[l]{names} & $\times$ & Soft & Soft & \makecell{---} & 2012 & \cite{almeida2012general} \\
\hhline{--}
\cellcolor{scorecolor!33.22053107707693}6.97 & TRAGEK & \makecell[l]{TRAnsductive Graph Embedding Kernel} & $\times$ & Soft & Soft & \makecell{---} & 2012 & \cite{chen2012non} \\
\hhline{--}
\cellcolor{scorecolor!33.22053107707693}6.97 & ENPAKL & \makecell[l]{Efficient Non-PArametric Kernel Learning} & $\times$ & Soft & Soft & \makecell{---} & 2012 & \cite{chen2012non} \\
\hhline{--}
\cellcolor{scorecolor!19.785647238323605}4.15 & ssFS & \makecell[l]{semi-supervised subgraph Feature Selection} & \checkmark & Soft & Soft & \makecell{---} & 2012 & \cite{huang2012semi} \\
\hhline{--}
\cellcolor{scorecolor!27.721538335364166}5.82 & SSKSRM & \makecell[l]{Semi-Supervised Kernel Switching Regression Models} & $\times$ & Hard & Hard & \makecell{---} & 2013 & \cite{tang2013semi} \\
\hhline{--}
\cellcolor{scorecolor!27.721538335364166}5.82 & SSSeKRM & \makecell[l]{Semi-Supervised Sequential Kernel Regression Models} & $\times$ & Hard & Hard & \makecell{---} & 2013 & \cite{tang2013semi} \\
\hhline{--}
\cellcolor{scorecolor!57.60532678408057}12.08 & SKML & \makecell[l]{Spectral Kernel Metric Learning} & $\times$ & Soft & Soft & \makecell{---} & 2014 & \cite{baghshah2014scalable}  \\
        \hline
		\end{tabular}}
	\caption{Feature table for KCC methods.}
	\label{tab:tax:kernel_cc}
\end{table}

\subsection{Fuzzy CC}

Fuzzy classic clustering represents a hard dichotomy within the clustering area. In fuzzy clustering, instances are allowed to belong to more than one cluster, with a list of probabilities that indicate the likelihood of said instance to belong to every cluster. The set of these lists is called a fuzzy partition, in contrast to crisp (or hard) partitions obtained by non-fuzzy clustering algorithms, where instances are assumed to belong to a single cluster with a 100\% probability~\cite{baraldi1999survey}. Fuzzy clustering algorithms are usually applied over relational information, meaning that the feature vector describing the instances in the dataset are not needed, only the pairwise relations between them, which can be computed as pairwise distances~\cite{ruspini2019fuzzy}. This makes the fuzzy clustering paradigm specially suitable to be extended in order to include constraints, as constraints are a natural type of relational information.

A list of Fuzzy CC (FCC) methods is proposed in Table~\ref{tab:tax:fuzzy_CC}. Even if this is one of the largest categories in clustering-based CC methods, authors have found that no further significant categorizations can be performed over it. The vast majority of methods in this category include constraints by means of a penalty term, which can be computed with different confidence degrees~\cite{maraziotis2012semi}. The confidence degree of the penalty refers to the number of possible assignations in which constraints violations are checked. Some methods examine all possible assignations for all constraints, such as SSCARD or AFCC, while others only examine the most probable assignation, like SSFCA, which is equivalent to perform violations checks in the crisp partition. Many methods are based on the PCCA method (see Section~\ref{subsec:tax:hcc}), such as AFCC, ACC, SS-CARD, PCsFCM and SS-CLAMP. These methods modify the objective function of PCCA to make it fuzzy or borrow its objective function to use it in a new optimization scheme. Besides, it is worth noting that some fuzzy methods are not relational but evidential. The evidential clustering framework is built around the concept of credal partition, which extends the concepts of crisp and fuzzy partitions and makes it possible to represent not only uncertainty, but also imprecision with respect to the class membership of an instance. Methods such as CECM, CEVCLUS and k-CEVCLUS include constraints into the evidential clustering paradigm, with CEVCLUS and k-CEVCLUS combining both relational en evidential fuzzy clustering features.

\begin{table}[!h]
	\centering
	\setlength{\tabcolsep}{7pt}
	\renewcommand{\arraystretch}{1.3}
	\resizebox{\textwidth}{!}{
		\begin{tabular}{c l l c c c c c c}
		\hline
		$S_{\mathcal{A}}$ & Acronym & Full Name & Penalty & ML & CL & Hybrid & Year & Ref. \\
		\hline
        \cellcolor{scorecolor!33.48210686022766}7.09 & SSCARD & \makecell[l]{Semi-Supervised Clustering and Aggregating Relational Data} & \checkmark & Soft & Soft & \makecell{Intra-View Constrained} & 2007 & \cite{frigui2007adaptive,frigui2008fuzzy} \\
\hhline{--}
\cellcolor{scorecolor!44.829047826412975}9.50 & AFCC & \makecell[l]{Active Fuzzy Constrained Clustering} & $\times$ & Soft & Soft & \makecell{Active Clustering with Constraints} & 2008 & \cite{grira2008active} \\
\hhline{--}
\cellcolor{scorecolor!0.0}0.00 & ACC & \makecell[l]{Adaptive Constrained Clustering} & \checkmark & Soft & Soft & \makecell{---} & 2008 & \cite{frigui2008image} \\
\hhline{--}
\cellcolor{scorecolor!0.0}0.00 & PCsFCM & \makecell[l]{Pairwise Constrained standard Fuzzy c-means} & $\times$ & Soft & Soft & \makecell{---} & 2009 & \cite{kanzawa2009some} \\
\hhline{--}
\cellcolor{scorecolor!0.0}0.00 & PCeFCM & \makecell[l]{Pairwise Constrained entropy Fuzzy c-means} & $\times$ & Soft & Soft & \makecell{---} & 2009 & \cite{kanzawa2009some} \\
\hhline{--}
\cellcolor{scorecolor!29.145884242329373}6.17 & SCAP & \makecell[l]{Semi-supervised fuzzy Clustering Algorithm with Pairwise constraints} & \checkmark & Soft & Soft & \makecell{---} & 2011 & \cite{gao2011new} \\
\hhline{--}
\cellcolor{scorecolor!22.73061661670279}4.82 & SSFCA & \makecell[l]{Semi-Supervised Fuzzy Clustering Algorithm} & $\times$ & Soft & Soft & \makecell{---} & 2012 & \cite{maraziotis2012semi} \\
\hhline{--}
\cellcolor{scorecolor!35.08663757669944}7.43 & CECM & \makecell[l]{Constrained Evidential C-Means} & $\times$ & Soft & Soft & \makecell{---} & 2012 & \cite{antoine2012cecm} \\
\hhline{--}
\cellcolor{scorecolor!30.273694803426075}6.41 & SS-CLAMP & \makecell[l]{Semi-Supervised fuzzy C-medoids CLustering Algorithm of relational data with Multiple Prototype representation} & \checkmark & Soft & Soft & \makecell{---} & 2013 & \cite{de2013semi} \\
\hhline{--}
\cellcolor{scorecolor!58.75222504785468}12.45 & SS-FCC & \makecell[l]{Semi-Supervised - Fuzzy Co-Clustering} & \checkmark & Soft & Soft & \makecell{Co-Clustering} & 2013 & \cite{yan2013semi} \\
\hhline{--}
\cellcolor{scorecolor!65.03530100434583}13.78 & SSeFCMCT & \makecell[l]{Semi-Supervised entropy Fuzzy C-Means for data with Clusterwise Tolerance by opposite criteria} & $\times$ & Soft & Soft & \makecell{---} & 2013 & \cite{hamasuna2013semi} \\
\hhline{--}
\cellcolor{scorecolor!65.03530100434583}13.78 & SSsFCMCT & \makecell[l]{Semi-Supervised standard Fuzzy C-Means for data with Clusterwise Tolerance by opposite criteria} & $\times$ & Soft & Soft & \makecell{---} & 2013 & \cite{hamasuna2013semi} \\
\hhline{--}
\cellcolor{scorecolor!83.2658462596397}17.64 & PC-eFCM-NM & \makecell[l]{Pairwise Constrained - entropy Fuzzy C-Means - Non Metric} & \checkmark & Soft & Soft & \makecell{---} & 2014 & \cite{endo2014hard} \\
\hhline{--}
\cellcolor{scorecolor!83.2658462596397}17.64 & PC-sFCM-NM & \makecell[l]{Pairwise Constrained - standard Fuzzy C-Means - Non Metric} & \checkmark & Soft & Soft & \makecell{---} & 2014 & \cite{endo2014hard} \\
\hhline{--}
\cellcolor{scorecolor!27.429062238187758}5.81 & CEVCLUS & \makecell[l]{Constrained EVidential CLUStering} & $\times$ & Soft & Soft & \makecell{---} & 2014 & \cite{antoine2014cevclus} \\
\hhline{--}
\cellcolor{scorecolor!79.13553791008631}16.77 & CMKIPCM & \makecell[l]{Constrained Multiple Kernels Improved Possibilistic C-Means} & $\times$ & Soft & Soft & \makecell{Active Clustering with Constraints} & 2015 & \cite{abin2015active} \\
\hhline{--}
\cellcolor{scorecolor!30.126897919386824}6.38 & SFFD & \makecell[l]{Semi-supervised Fuzzy clustering with Feature Discrimination} & \checkmark & Soft & Soft & \makecell{---} & 2015 & \cite{li2015semi} \\
\hhline{--}
\cellcolor{scorecolor!100.0}21.19 & AAA & \makecell[l]{names} & $\times$ & Soft & Soft & \makecell{Active Clustering with Constraints} & 2016 & \cite{abin2016clustering} \\
\hhline{--}
\cellcolor{scorecolor!32.299948317325075}6.84 & k-CEVCLUS & \makecell[l]{k - Constrained EVidentialCLUStering} & $\times$ & Soft & Soft & \makecell{---} & 2018 & \cite{li2018k}  \\
        \hline
		\end{tabular}}
	\caption{Feature table for FCC methods.}
	\label{tab:tax:fuzzy_CC}
\end{table}

\subsection{Mixture Model-based CC}

Mixture models are parametric statistical models which assume that a dataset originates from a weighted sum of several statistical sources. These sources can typically be Gaussian distributions, originating the Gaussian Mixture Model (GMM). However, other statistical distributions (like the Dirichlet distribution) can be used in the mixture models paradigm. In GMM-based clustering, each cluster is associated to a parameterized Gaussian distribution. These parameters are optimized for the final distribution to explain its associated cluster. The result of GMM-based clustering is not a crisp partition, but the probability for each instance to be generated by each of the available optimized Gaussian distributions. EM schemes are one of the most widely used methods to optimize the parameters of the distribution~\cite{dempster1977maximum}.

Table~\ref{tab:tax:MM_CC} gathers Mixture Model-based CC (MMbCC) methods. The most common way to include constraints in GMM-based clustering is by discarding unwanted distributions. For the case of hard CC, this is done by removing all distributions which violate any constraints from the addition of Gaussians, as in Constrained EM or DPMM. In the case of soft CC, the summation of distributions is modified to take these distributions into account to a higher or lower extent depending on the number of violated constraints and on the relevance of constraints themselves. This can be achieved by means of a penalty-style objective function, as in sRLe-GDM-FFS, MCGMM, SCGMM, or by assigning a level of confidence to each constraint, as in MAA or PPC. Some methods use different statistical models (not Gaussian), like DPMM, which uses Diritchlet Processes, or they allow a single cluster to be represented by more than one distribution, as in MCGMM. SPGP is not based on Gaussian mixture models but on Gaussian process classifiers (GPC). Given its similitude with GMM, authors have decided to include it in this category. The main difference between these two paradigms is that GMM are generative models, while GPC are discriminative models.

\begin{table}[!h]
	\centering
	\setlength{\tabcolsep}{7pt}
	\renewcommand{\arraystretch}{1.3}
	\resizebox{\textwidth}{!}{
		\begin{tabular}{c l l c c c c c c}
		\hline
		$S_{\mathcal{A}}$ & Acronym & Full Name & Penalty & ML & CL & Hybrid & Year & Ref. \\
		\hline
        \cellcolor{scorecolor!26.469651641194318}5.45 & MAA & \makecell[l]{names} & $\times$ & Hybrid & Hybrid & \makecell{---} & 2004 & \cite{law2004clustering} \\
\hhline{--}
\cellcolor{scorecolor!85.99543328712873}17.70 & Constrained EM & \makecell[l]{Constrained Expectation-Minimization} & $\times$ & Hard & Hard & \makecell{---} & 2004 & \cite{shental2004computing} \\
\hhline{--}
\cellcolor{scorecolor!22.677948133134198}4.67 & PPC & \makecell[l]{Penalized Probabilistic Clustering} & $\times$ & Hybrid & Hybrid & \makecell{---} & 2005 & \cite{lu2005semi,lu2007penalized} \\
\hhline{--}
\cellcolor{scorecolor!100.0}20.58 & MCGMM & \makecell[l]{Multiple-Component Gaussian Mixture Model} & $\times$ & Soft & Soft & \makecell{---} & 2005 & \cite{zhao2005mixture} \\
\hhline{--}
\cellcolor{scorecolor!98.39715048975958}20.25 & SCGMM & \makecell[l]{Single-Component Gaussian Mixture Model} & $\times$ & Soft & Soft & \makecell{---} & 2005 & \cite{zhao2005mixture} \\
\hhline{--}
\cellcolor{scorecolor!43.786624804898324}9.01 & SPGP & \makecell[l]{Semi-supervised Pairwise Gaussian Process classifier} & $\times$ & Hybrid & Hybrid & \makecell{---} & 2007 & \cite{lu2007semi} \\
\hhline{--}
\cellcolor{scorecolor!7.028367892125681}1.45 & CDPMM08 & \makecell[l]{Constrained Dirichlet Process Mixture Models 2008} & $\times$ & Hard & Hard & \makecell{---} & 2008 & \cite{vlachos2008dirichlet} \\
\hhline{--}
\cellcolor{scorecolor!35.47592907086226}7.30 & sRLe-GDM-FFS & \makecell[l]{semi-supervised Robust Learning of finite Generalized Dirichlet Mixture models and Feature Subset Selection} & $\times$ & Soft & Soft & \makecell{---} & 2012 & \cite{ismail2012automatic} \\
\hhline{--}
\cellcolor{scorecolor!22.910080417646604}4.71 & C$_4$S & \makecell[l]{Constrained Clustering with a Complex Cluster Structure} & $\times$ & Soft & Soft & \makecell{---} & 2017 & \cite{smieja2017constrained} \\
\hhline{--}
\cellcolor{scorecolor!67.90370225348988}13.97 & FIECE-EM & \makecell[l]{Feasible-Infeasible Evolutionary Create \& Eliminate - Expectation Maximization} & $\times$ & Hard & Hard & \makecell{Genetic Algorithm} & 2018 & \cite{covoes2018classification} \\
\hhline{--}
\cellcolor{scorecolor!39.74314703919892}8.18 & JDG & \makecell[l]{names} & $\times$ & Soft & Soft & \makecell{---} & 2018 & \cite{covoes2018classification} \\
\hhline{--}
\cellcolor{scorecolor!75.60837595707358}15.56 & FIECE-EM+BFCU & \makecell[l]{FIECE-EM + Best Feasible Classification Uncertainty} & $\times$ & Hard & Hard & \makecell{Active Clustering with Constraints \& Genetic Algorithm} & 2020 & \cite{fernandes2019active,fernandes2020improving} \\
\hhline{--}
\cellcolor{scorecolor!75.60837595707358}15.56 & FIECE-EM+FCU & \makecell[l]{FIECE-EM + Feasible Classification Uncertainty} & $\times$ & Hard & Hard & \makecell{Active Clustering with Constraints \& Genetic Algorithm} & 2020 & \cite{fernandes2019active,fernandes2020improving} \\
\hhline{--}
\cellcolor{scorecolor!75.60837595707358}15.56 & FIECE-EM+DVO & \makecell[l]{FIECE-EM + Distance to Violated Objects} & $\times$ & Hard & Hard & \makecell{Active Clustering with Constraints \& Genetic Algorithm} & 2020 & \cite{fernandes2019active,fernandes2020improving} \\
\hhline{--}
\cellcolor{scorecolor!75.60837595707358}15.56 & FIECE-EM+LUC & \makecell[l]{FIECE-EM + Largest Unlabeled Clusters} & $\times$ & Hard & Hard & \makecell{Active Clustering with Constraints \& Genetic Algorithm} & 2020 & \cite{fernandes2019active,fernandes2020improving}  \\
        \hline
		\end{tabular}}
	\caption{Feature table for MMbCC methods.}
	\label{tab:tax:MM_CC}
\end{table}

\subsection{Hierarchical CC} \label{subsec:tax:hcc}

Details in classic hierarchical clustering and hierarchical CC have been already introduced in Section~\ref{subsec:classic_clustering}. Let us remember that hierarchical clustering methods produce a dendrogram, instead of a partition. Affinity criteria are used to determine cluster merges in every lever of the dendrogram.

Table~\ref{tab:tax:hierarchical_CC} presents a list of Hierarchical CC (HCC) methods. Many strategies designed to include constraints into hierarchical clustering can be found in this category. Some on them modify the clustering engine of existing methods to include constraints in the process of selecting the clusters to merge, such as COP-COBWEB, C-DBSCAN, CAC1, Cons-DBSCAN or SDHCC. Others transform the dataset in some way for it to include the information contained in the constraint set, such as COBRA, COBRAS, C-DenStream. AHC-CTP includes constraints in the computation of the dissimilarities without using a penalty term, while methods such as 2SHACC and PCCA have to use one. The only divisive hierarchical CC method found by the authors is SDHCC, all of the rest perform hierarchical agglomerative CC.

\begin{table}[!h]
	\centering
	\setlength{\tabcolsep}{7pt}
	\renewcommand{\arraystretch}{1.3}
	\resizebox{\textwidth}{!}{
		\begin{tabular}{c l l c c c c c c}
		\hline
		$S_{\mathcal{A}}$ & Acronym & Full Name & Penalty & ML & CL & Hybrid & Year & Ref. \\
		\hline
        \cellcolor{scorecolor!21.237324777787066}4.78 & COP-COBWEB & \makecell[l]{COnstrained Partitional - COBWEB} & $\times$ & Hard & Hard & \makecell{---} & 2000 & \cite{wagstaff2000clustering} \\
\hhline{--}
\cellcolor{scorecolor!85.69928985536278}19.30 & IDSSR & \makecell[l]{names} & $\times$ & Hard & Hard & \makecell{---} & 2005 & \cite{davidson2005agglomerative,davidson2009using} \\
\hhline{--}
\cellcolor{scorecolor!19.70607401010503}4.44 & PCCA & \makecell[l]{Pairwise Constrained Competitive Agglomeration} & $\times$ & Soft & Soft & \makecell{---} & 2005 & \cite{grira2005semi,grira2006fuzzy} \\
\hhline{--}
\cellcolor{scorecolor!17.205507040578304}3.87 & C-DBSCAN & \makecell[l]{Constraint-driven - DBSCAN} & $\times$ & Hard & Hard & \makecell{Density-based CC} & 2007 & \cite{ruiz2007c} \\
\hhline{--}
\cellcolor{scorecolor!19.88500763835673}4.48 & C-DenStream & \makecell[l]{Constrained - Density Stream} & $\times$ & Hard & Hard & \makecell{Density-based CC \& Online CC} & 2009 & \cite{ruiz2009c} \\
\hhline{--}
\cellcolor{scorecolor!0.0}0.00 & AHC-CTP & \makecell[l]{Agglomerative Hierarchical Clustering – Clusterwise Tolerance Pairwise} & $\times$ & Soft & Soft & \makecell{---} & 2010 & \cite{hamasuna2010semi,hamasuna2011semi} \\
\hhline{--}
\cellcolor{scorecolor!18.506737954282947}4.17 & CAC1 & \makecell[l]{Constrained Active Clustering 1} & $\times$ & Hard & Soft & \makecell{Active Clustering with Constraints} & 2011 & \cite{biswas2011large} \\
\hhline{--}
\cellcolor{scorecolor!22.529101051691367}5.07 & SDHCC & \makecell[l]{Semi-supervised Divisive Hierarchical Clustering of Categorical data} & $\times$ & Hard & Soft & \makecell{---} & 2011 & \cite{xiong2011semi} \\
\hhline{--}
\cellcolor{scorecolor!15.266625396514499}3.44 & AHCP & \makecell[l]{Agglomerative Hierarchical Clustering with Penalties} & $\times$ & Hard & Soft & \makecell{---} & 2011 & \cite{miyamoto2011constrained} \\
\hhline{--}
\cellcolor{scorecolor!14.622912709352875}3.29 & SGID & \makecell[l]{names} & $\times$ & Hard & Hard & \makecell{SAT} & 2011 & \cite{gilpin2011incorporating} \\
\hhline{--}
\cellcolor{scorecolor!25.470322245507028}5.74 & Cons-DBSCAN & \makecell[l]{Constrained - DBSCAN} & $\times$ & Hard & Hard & \makecell{Density-based CC \& Active Constraint Acquisition} & 2012 & \cite{zhao2012effective} \\
\hhline{--}
\cellcolor{scorecolor!15.856398497242258}3.57 & Active-HACC & \makecell[l]{Active - Hierarchical Agglomerative Constrained Clustering} & $\times$ & Soft & Soft & \makecell{Active Clustering with Constraints} & 2014 & \cite{biswas2014active} \\
\hhline{--}
\cellcolor{scorecolor!18.6349499751057}4.20 & COBRA & \makecell[l]{COnstraint-Based Repeated Aggregation} & $\times$ & Soft & Soft & \makecell{Active Clustering with Constraints} & 2017 & \cite{van2017cobra} \\
\hhline{--}
\cellcolor{scorecolor!86.89835040454639}19.57 & COBRAS & \makecell[l]{COnstraint-Based Repeated Aggregation and Splitting} & $\times$ & Soft & Soft & \makecell{Active Clustering with Constraints} & 2018 & \cite{craenendonck2018cobras} \\
\hhline{--}
\cellcolor{scorecolor!73.2998136883375}16.51 & 2SHACC & \makecell[l]{2 -Stages Hybrid Agglomerative Constrained Clustering} & $\times$ & Soft & Soft & \makecell{Constraint Propagation} & 2020 & \cite{gonzalez2020agglomerative} \\
\hhline{--}
\cellcolor{scorecolor!100.0}22.52 & 3SHACC & \makecell[l]{3 -Stages Hybrid Agglomerative Constrained Clustering} & \checkmark & Soft & Soft & \makecell{Constrained Distance Transformation} & 2022 & \cite{gonzalez20213shacc}  \\
        \hline
		\end{tabular}}
	\caption{Feature table for HCC methods.}
	\label{tab:tax:hierarchical_CC}
\end{table}

\subsection{Density-based CC} 

In density-based classic clustering, a cluster is considered to be a set of instances spread in the data space over a contiguous region with high density of instances. Density-based methods separate clusters by identifying regions in the input space with low density of instances, which are usually considered as noise or outliers~\cite{kriegel2011density}.

Table~\ref{tab:tax:density_CC} presents a list of Density-based CC (DbCC) methods. Two main strategies are used to include constraints into density-based clustering methods. The first one consists of modifying the assignation rule for instances to cluster, taking constraints into account, similarly to the way in which it is done in cluster engine-adapting methods. This strategy is used in methods like C-DBSCAN, C-DenStream, Cons-DBSCAN, SDenPeak or SSDC. On the other hand, some methods use constraints to redefine the density computation method to take them into account, such as in SemiDen or YZWD. In addition, there are methods that simply use constraints to modify the similarity measure or the dataset on the basis of the constraints, which run classic density-based clustering algorithm over them afterwards, such as SSDPC or fssDBSCAN.

\begin{table}[!h]
	\centering
	\setlength{\tabcolsep}{7pt}
	\renewcommand{\arraystretch}{1.3}
	\resizebox{\textwidth}{!}{
		\begin{tabular}{c l l c c c c c c}
		\hline
		$S_{\mathcal{A}}$ & Acronym & Full Name & Penalty & ML & CL & Hybrid & Year & Ref. \\
		\hline
        \cellcolor{scorecolor!21.71375273216283}3.87 & C-DBSCAN & \makecell[l]{Constraint-driven - DBSCAN} & $\times$ & Hard & Hard & \makecell{Hierarchical CC} & 2007 & \cite{ruiz2007c} \\
\hhline{--}
\cellcolor{scorecolor!25.09534522395188}4.48 & C-DenStream & \makecell[l]{Constrained - Density Stream} & $\times$ & Hard & Hard & \makecell{Hierarchical CC \& Online CC} & 2009 & \cite{ruiz2009c} \\
\hhline{--}
\cellcolor{scorecolor!32.144143031826445}5.74 & Cons-DBSCAN & \makecell[l]{Constrained - DBSCAN} & $\times$ & Hard & Hard & \makecell{Hierarchical CC \& Active Constraint Acquisition} & 2012 & \cite{zhao2012effective} \\
\hhline{--}
\cellcolor{scorecolor!21.449292440092538}3.83 & SDenPeak & \makecell[l]{Semi-Supervised Density Peak} & $\times$ & Soft & Soft & \makecell{---} & 2016 & \cite{fan2016sdenpeak} \\
\hhline{--}
\cellcolor{scorecolor!18.696613336310858}3.34 & SemiDen & \makecell[l]{Semi-supervised Density-based data clustering} & $\times$ & Hard & Hard & \makecell{---} & 2017 & \cite{atwa2017constraint} \\
\hhline{--}
\cellcolor{scorecolor!24.265151939658555}4.33 & YZWD & \makecell[l]{names} & $\times$ & Soft & Soft & \makecell{---} & 2017 & \cite{yang2017adaptive} \\
\hhline{--}
\cellcolor{scorecolor!30.12110497171861}5.37 & SSDC & \makecell[l]{Semi-Supervised DenPeak Clustering} & $\times$ & Soft & Hard & \makecell{---} & 2018 & \cite{ren2018semi} \\
\hhline{--}
\cellcolor{scorecolor!89.35883535465017}15.94 & SSDPC & \makecell[l]{Semi-Supervised Density Peak Clustering} & $\times$ & Soft & Soft & \makecell{---} & 2020 & \cite{yan2021semi} \\
\hhline{--}
\cellcolor{scorecolor!94.14285326077928}16.80 & fssDBSCAN & \makecell[l]{fast semi-supervised DBSCAN} & $\times$ & Soft & Soft & \makecell{Time Series} & 2021 & \cite{he2019fast} \\
\hhline{--}
\cellcolor{scorecolor!96.51271709869663}17.22 & ADPE & \makecell[l]{Active Density Peak Ensemble} & $\times$ & Soft & Soft & \makecell{Active Clustering with Constraints \& Constrained Pool Generation} & 2021 & \cite{shi2021fast} \\
\hhline{--}
\cellcolor{scorecolor!100.0}17.84 & ADP & \makecell[l]{Active Density Peak} & $\times$ & Soft & Soft & \makecell{Active Clustering with Constraints} & 2021 & \cite{shi2021fast}  \\
        \hline
		\end{tabular}}
	\caption{Feature table for DbCC methods.}
	\label{tab:tax:density_CC}
\end{table}

\subsection{Online CC}

Online clustering methods perform clustering over data which varies over time. This is called a data stream. In classic online clustering, new data instances arrive (in the form of chunks or single instances) over time and the goal of the clustering algorithm is to produce a partition of the current set of instances, usually taking into account information obtained from past instances. In online constrained clustering, not only new instances are provided to the method over time, but also constraints and, in some cases, only constraints. Methods performing Online CC (OCC) are gathered in Table~\ref{tab:tax:online_cc}.

The CME algorithm is an online wrapper for the COP-K-Means algorithms. It gradually forgets past constraints, lowering their effect on the current partition as new data from the data stream arrive. TDCK-Means is built on the basis of classic K-Means and also uses a decay term to handle online constraints. It addresses the temporal nature of the data by adapting the Euclidean distance to take into account both the distance in the multidimensional space and in the temporal space. A penalty term is then added to include constraints, which is more severe for instances closer in time and whose magnitude decays over time. CS2GS also uses constraint weight decay to perform online CC and is based on SOM. The architecture of the neural network used by CS2GS features two layers. The between-layer weights and the number of nodes of the first layer are adapted to dynamically correct the violation of constraints. Using the metrics included in these layers as a reference, the violation of constraints is quantified as network's error. Weights are modified over time based on the error to obtain the new weights. The weight update procedure tries to satisfy the currently violated constraints while keeping the new weights close to the old ones to avoid breaking the old constraints.

O-LCVQE and C-RPCL are both online competitive learners. Competitive learning algorithms are characterized by competition among $k$ neurons, which compete to learn instances. This is known as the winner-take-all (WTA) approach. Competitive learning can be seen as performing cluster in the input space, viewing the neurons as centroids and using the Euclidean distance as the competition score. The O-LCVQE is a WTA approach that can only deal with CL constraints by defining the score as in LCVQE. Therefore the winner centroid is computed with regards to the objective function of LCVQE modified to consider only CL constraints. Similarly, the RPCL algorithm~\cite{xu1993rival} can be modified to include only CL constraints within the WTA, resulting in C-RPCL. The intuition behind this modified algorithm is that, if a CL constraint gets violated by assigning the instance to a given centroid, C-RPCL searches for the nearest rival which does not cause any constraint violations. This nearest rival becomes the winner, and the previous winner prototype is moved away from that instance.

C-DenStream is based on the density-based clustering DenStream, which is the online version of DBSCAN. DenStream performs density-based clustering. it uses the micro-cluster density, which is based on weighting areas of instances in a neighborhood as a result of an exponential decay function over time. C-DenStream includes constraints into the DenStream clustering process by translating instance-level constraints into micro-cluster-level constraints using the micro-cluster membership of each instance in each timestamp. SemiStream builds an initial partition using MPCK-Means and updates it as new chunks of data arrive. The update process consist of performing clustering assigning pairs of constrained instances to clusters, minimizing the cost of said assignment. The SCSC algorithm is the only online cc algorithm that keeps the dataset constant over time and consider the time dimension only over the constraint set. It consists of two main components, an offline procedure to build a convex hull, and an online procedure to update the clustering results when new pairwise constraints are received. 

\begin{table}[!h]
	\centering
	\setlength{\tabcolsep}{7pt}
	\renewcommand{\arraystretch}{1.3}
	\resizebox{\textwidth}{!}{
		\begin{tabular}{c l l c c c c c c}
		\hline
		$S_{\mathcal{A}}$ & Acronym & Full Name & Penalty & ML & CL & Hybrid & Year & Ref. \\
		\hline
        \cellcolor{scorecolor!2.8728965557348602}0.35 & CME & \makecell[l]{names} & \checkmark & Hard & Hard & \makecell{---} & 2006 & \cite{moreno2006user} \\
\hhline{--}
\cellcolor{scorecolor!36.70693150489831}4.48 & C-DenStream & \makecell[l]{Constrained - Density Stream} & $\times$ & Hard & Hard & \makecell{Hierarchical CC \& Density-based CC} & 2009 & \cite{ruiz2009c} \\
\hhline{--}
\cellcolor{scorecolor!35.85496707913376}4.37 & SemiStream & \makecell[l]{---} & \checkmark & Soft & Soft & \makecell{Penalty-based Methods} & 2012 & \cite{halkidi2012semi} \\
\hhline{--}
\cellcolor{scorecolor!100.0}12.20 & O-LCVQE & \makecell[l]{Online - Linear Constrained Vector Quantization Error} & \checkmark & --- & Soft & \makecell{Neural Networks-based CC} & 2013 & \cite{covoes2012competitive} \\
\hhline{--}
\cellcolor{scorecolor!100.0}12.20 & C-RPCL & \makecell[l]{Constrained - Rival Penalyzed Competitive Learning} & $\times$ & --- & Soft & \makecell{Neural Networks-based CC} & 2013 & \cite{covoes2012competitive} \\
\hhline{--}
\cellcolor{scorecolor!35.56919545195542}4.34 & TDCK-Means & \makecell[l]{Temporal-Driven Constrained K-Means} & \checkmark & Soft & Soft & \makecell{Penalty-based Methods} & 2014 & \cite{rizoiu2014use} \\
\hhline{--}
\cellcolor{scorecolor!88.07593285073237}10.74 & CS2GS & \makecell[l]{Constrained Semi-Supervised Growing SOM} & $\times$ & Soft & Soft & \makecell{Self Organizing Maps-based CC} & 2015 & \cite{allahyar2015constrained} \\
\hhline{--}
\cellcolor{scorecolor!44.94809239047859}5.48 & SCSC & \makecell[l]{Semi-supervised Clustering with Sequential Constraints} & $\times$ & Soft & Soft & \makecell{Constrained Pool Generation} & 2015 & \cite{yi2015efficient}  \\
        \hline
		\end{tabular}}
	\caption{Feature table for OCC methods.}
	\label{tab:tax:online_cc}
\end{table}

\subsection{Others} 

This section gathers minor CC categories. These categories, shown in Table~\ref{tab:tax:others}, are considered to be minor because of the number of methods belonging to them (5 or less), or because of the restricted applicability or specificity of said methods. A total of 15 minor CC categories are briefly introduced, for a total of 40 methods.

\begin{table}[!h]
	\centering
	\setlength{\tabcolsep}{7pt}
	\renewcommand{\arraystretch}{1.3}
	\resizebox{\textwidth}{!}{
		\begin{tabular}{c l l c c c c c c c}
		\hline
		$S_{\mathcal{A}}$ & Acronym & Full Name & Category & Penalty & ML & CL & Hybrid & Year & Ref. \\
		\hline
        \cellcolor{scorecolor!32.09535531292551}5.49 & SSAP & \makecell[l]{Semi-Supervised Affinity Propagation} & Affinity Propagation & $\times$ & Soft & Soft & \makecell{---} & 2009 & \cite{givoni2009semi} \\
\hhline{--}
\cellcolor{scorecolor!53.381813386223484}9.13 & COALA & \makecell[l]{Constrained Orthogonal Average Link Algorithm} & Alternative Clustering & $\times$ & Any & Any & \makecell{---} & 2006 & \cite{bae2006coala} \\
\hhline{--}
\cellcolor{scorecolor!53.381813386223484}9.13 & COALAcat & \makecell[l]{Constrained Orthogonal Average Link Algorithm (Categorical)} & Alternative Clustering & $\times$ & Any & Any & \makecell{---} & 2006 & \cite{bae2006coala} \\
\hhline{--}
\cellcolor{scorecolor!25.36550243225721}4.34 & ADFT & \makecell[l]{Alternative Distance Function Transformation} & Alternative Clustering & $\times$ & Soft & Soft & \makecell{---} & 2008 & \cite{davidson2008finding} \\
\hhline{--}
\cellcolor{scorecolor!50.04488948110166}8.56 & ClusILC & \makecell[l]{Clustering with Instance-Level Constraints} & Clustering Trees & $\times$ & Soft & Soft & \makecell{---} & 2007 & \cite{struyf2007clustering} \\
\hhline{--}
\cellcolor{scorecolor!72.76167576941758}12.45 & SS-FCC & \makecell[l]{Semi-Supervised - Fuzzy Co-Clustering} & Co-Clustering & \checkmark & Soft & Soft & \makecell{Fuzzy CC} & 2013 & \cite{yan2013semi} \\
\hhline{--}
\cellcolor{scorecolor!69.45533674164236}11.88 & SS-NMF & \makecell[l]{Semi-Supervised Nonnegative Matrix Factorization} & Co-Clustering & $\times$ & Soft & Soft & \makecell{Non-negative Matrix Factorization CC} & 2010 & \cite{chen2009non} \\
\hhline{--}
\cellcolor{scorecolor!12.021859704224998}2.06 & RJFM & \makecell[l]{names} & Co-Clustering & $\times$ & Hard & Hard & \makecell{---} & 2010 & \cite{pensa2010co} \\
\hhline{--}
\cellcolor{scorecolor!27.718397441436625}4.74 & OSS-NMF & \makecell[l]{Orthogonal Semi-Supervised - Non-negative Matrix tri-Factorization} & Co-Clustering & $\times$ & Soft & Soft & \makecell{Non-negative Matrix Factorization CC} & 2010 & \cite{ma2010orthogonal} \\
\hhline{--}
\cellcolor{scorecolor!29.681721770513082}5.08 & CCG & \makecell[l]{Constrained Column Generation} & Column Generation & $\times$ & Hard & Hard & \makecell{---} & 2014 & \cite{babaki2014constrained} \\
\hhline{--}
\cellcolor{scorecolor!67.24327696927602}11.50 & CG+PR+LS & \makecell[l]{Column Generation + Path Relinking + Local Search} & Column Generation & $\times$ & Soft & Soft & \makecell{Single Individual} & 2017 & \cite{de2017comparison} \\
\hhline{--}
\cellcolor{scorecolor!36.90906756551448}6.31 & TKC(17) & \makecell[l]{names} & Constraint Programming & $\times$ & Hard & Hard & \makecell{---} & 2017 & \cite{duong2017constrained} \\
\hhline{--}
\cellcolor{scorecolor!46.92549764586326}8.03 & 3CP & \makecell[l]{Constrained Clustering by Constraint Programming} & Constraint Programming & $\times$ & Hard & Hard & \makecell{---} & 2015 & \cite{calandriello2014semi} \\
\hhline{--}
\cellcolor{scorecolor!13.912469784878184}2.38 & BBMSC & \makecell[l]{Branch-and-Bound Method for Subspace Clustering} & Constraint Programming & \checkmark & Soft & Soft & \makecell{---} & 2015 & \cite{hu2014exploiting} \\
\hhline{--}
\cellcolor{scorecolor!90.3113928266059}15.45 & CMSSCCP & \makecell[l]{ Constrained Minimum Sum of Squares Clustering by Constraint Programming} & Constraint Programming & $\times$ & Soft & Soft & \makecell{---} & 2015 & \cite{dao2015constrained} \\
\hhline{--}
\cellcolor{scorecolor!44.328764193522915}7.58 & CCPP & \makecell[l]{Constrained Clustering via Post-Processing} & Constraint Programming & $\times$ & Hard & Hard & \makecell{---} & 2020 & \cite{nghiem2020constrained} \\
\hhline{--}
\cellcolor{scorecolor!95.83536946620413}16.39 & cut & \makecell[l]{---} & Constraint Programming & $\times$ & Soft & Soft & \makecell{---} & 2009 & \cite{xia2009global} \\
\hhline{--}
\cellcolor{scorecolor!52.944778511590776}9.06 & TKC & \makecell[l]{names} & Constraint Programming & $\times$ & Hard & Hard & \makecell{---} & 2013 & \cite{dao2013declarative} \\
\hhline{--}
\cellcolor{scorecolor!24.096236785452923}4.12 & YLYM & \makecell[l]{names} & Feature Selection & $\times$ & Soft & Soft & \makecell{---} & 2008 & \cite{li2008feature} \\
\hhline{--}
\cellcolor{scorecolor!43.99817532240877}7.53 & SCAN & \makecell[l]{Semi-supervised clustering with Coupled attributes in Attributed \\ heterogeneous information Networks} & HIN & $\times$ & Soft & Soft & \makecell{---} & 2019 & \cite{zhao2019coupled} \\
\hhline{--}
\cellcolor{scorecolor!15.525981303062705}2.66 & SCHAIN & \makecell[l]{Semi-supervised Clustering in Heterogeneous Attributed Information Networks} & HIN & $\times$ & Soft & Soft & \makecell{---} & 2017 & \cite{li2017semi} \\
\hhline{--}
\cellcolor{scorecolor!41.03800698917111}7.02 & EICC & \makecell[l]{Efficient Incremental Constrained Clustering} & Incremental CC & $\times$ & Hard & Hard & \makecell{---} & 2007 & \cite{davidson2007efficient} \\
\hhline{--}
\cellcolor{scorecolor!25.82378005268081}4.42 & 3SMIC & \makecell[l]{Semi-Supervised Squared-loss Mutual Information Clustering} & Information Maximization & $\times$ & Soft & Soft & \makecell{---} & 2014 & \cite{calandriello2014semi} \\
\hhline{--}
\cellcolor{scorecolor!32.422948426557625}5.55 & NLPPC & \makecell[l]{New Label Propagation with Pairwise Constraints} & Label Propagation  & \checkmark & Soft & Soft & \makecell{---} & 2020 & \cite{bai2020new} \\
\hhline{--}
\cellcolor{scorecolor!28.552332367712395}4.88 & PCKMMR & \makecell[l]{Parallel COP-K-Means based on MapReduce} & MapReduce & $\times$ & Soft & Soft & \makecell{---} & 2011 & \cite{lin2011parallel} \\
\hhline{--}
\cellcolor{scorecolor!64.51574494572819}11.04 & MVMC & \makecell[l]{Multi-View Matrix Completion} & Matrix Completion & $\times$ & Soft & Soft & \makecell{Intra-View Constrained \& Spectral CC} & 2017 & \cite{zhao2017multi} \\
\hhline{--}
\cellcolor{scorecolor!69.14653724149946}11.83 & PMMC & \makecell[l]{Pairwise-constrained Maximum Margin Clustering} & Maximum Margin Clustering & \checkmark & Soft & Soft & \makecell{---} & 2012 & \cite{zeng2011semi} \\
\hhline{--}
\cellcolor{scorecolor!100.0}17.11 & CMMC & \makecell[l]{Constrained Maximum Margin Clustering} & Maximum Margin Clustering & $\times$ & Soft & Soft & \makecell{---} & 2008 & \cite{hu2008maximum} \\
\hhline{--}
\cellcolor{scorecolor!76.9437244321038}13.16 & TwoClaCMMC & \makecell[l]{Two Classes Constrained Maximum Margin Clustering} & Maximum Margin Clustering & $\times$ & Soft & Soft & \makecell{---} & 2013 & \cite{zeng2013improving} \\
\hhline{--}
\cellcolor{scorecolor!51.01949889899523}8.73 & DCPR & \makecell[l]{---} & Probabilistic Clustering & $\times$ & Hybrid & Hybrid & \makecell{---} & 2016 & \cite{pei2016comparing} \\
\hhline{--}
\cellcolor{scorecolor!41.76264301837587}7.14 & RHWL & \makecell[l]{names} & Probabilistic Clustering & $\times$ & Soft & Soft & \makecell{Active Clustering with Constraints} & 2007 & \cite{huang2007semi} \\
\hhline{--}
\cellcolor{scorecolor!80.44318160980544}13.76 & d-graph & \makecell[l]{---} & Probabilistic Clustering & $\times$ & Soft & Soft & \makecell{---} & 2018 & \cite{smieja2018semi} \\
\hhline{--}
\cellcolor{scorecolor!17.42342871865417}2.98 & JPBMS & \makecell[l]{names} & SAT & $\times$ & Hard & Hard & \makecell{} & 2012 & \cite{metivier2012constrained} \\
\hhline{--}
\cellcolor{scorecolor!19.24977395222691}3.29 & SGID & \makecell[l]{names} & SAT & $\times$ & Hard & Hard & \makecell{Hierarchical CC} & 2011 & \cite{gilpin2011incorporating} \\
\hhline{--}
\cellcolor{scorecolor!23.488323721857974}4.02 & ISL & \makecell[l]{names} & SAT & $\times$ & Hard & Hard & \makecell{---} & 2010 & \cite{davidson2010sat} \\
\hhline{--}
\cellcolor{scorecolor!18.716114262536646}3.20 & JBMJ & \makecell[l]{names} & SAT & $\times$ & Hard & Hard & \makecell{---} & 2017 & \cite{berg2017cost} \\
\hhline{--}
\cellcolor{scorecolor!75.83778448975451}12.97 & CPSC*-PS & \makecell[l]{Constrained Polygonal Spatial Clustering-Polygon Split} & Spatial CC & $\times$ & Soft & Soft & \makecell{---} & 2012 & \cite{joshi2011redistricting} \\
\hhline{--}
\cellcolor{scorecolor!75.83778448975451}12.97 & CPSC* & \makecell[l]{Constrained Polygonal Spatial Clustering*} & Spatial CC & $\times$ & Soft & Soft & \makecell{---} & 2012 & \cite{joshi2011redistricting} \\
\hhline{--}
\cellcolor{scorecolor!20.675081452260105}3.54 & CPSC & \makecell[l]{Constrained Polygonal Spatial Clustering} & Spatial CC & $\times$ & Hard & Hard & \makecell{---} & 2012 & \cite{joshi2011redistricting} \\
\hhline{--}
\cellcolor{scorecolor!98.20004592431877}16.80 & fssK-Means & \makecell[l]{fast semi-supervised K-Means} & Time Series & \checkmark & Soft & Soft & \makecell{Penalty-based Methods} & 2021 & \cite{he2019fast} \\
\hhline{--}
\cellcolor{scorecolor!98.20004592431877}16.80 & fssDBSCAN & \makecell[l]{fast semi-supervised DBSCAN} & Time Series & $\times$ & Soft & Soft & \makecell{Density-based CC} & 2021 & \cite{he2019fast}  \\
        \hline
		\end{tabular}}
	\caption{Feature table for CC algorithms belonging to minor categories (Others).}
	\label{tab:tax:others}
\end{table}

\subsubsection{Constrained Co-clustering}

Co-clustering methods perform clustering on the column and the rows of a given dataset at the same time, considering them as closely related different sources of information. OSS-NMF extends the classic NMF by introducing constraints and performing clustering by solving a constrained optimization problem. This method is specifically proposed to solve the document clustering task, and does so by performing co-clustering in words and documents simultaneously, and considering both word-level and document-level constraints. RJFM is based on the meta-algorithm called Bregman Co-clustering, which can optimize a large class of objective functions belonging to the Bregman divergences. Its principle is simple: it alternatively refines row and column clusters, while optimizing an objective function that takes both partitions into account. It includes constraints in both columns and rows clustering the same way: it never performs rows or columns assignation breaking CL constraints, and always assigns full cliques of ML constraints. SS-NMF learns a new metric by applying simultaneously distance metric learning and modality selection. The new metric is used to derive distance matrices over which clustering is finally performed. SS-FCC formulates the CC problem as an optimization problem with an objective function built on the basis of the competitive agglomeration cost with fuzzy terms and constraint-based penalties. It introduces cooperation into the co-clustering process by including two fuzzy memberships, one of them related to columns and other related to rows, which are expected to be highly correlated. The amount of cooperation is delivered by the degree of aggregation, which should be maximized among clusters to accomplish the clustering task.

\subsubsection{Alternative Clustering based on constraints}

Constrains can have multiple uses, other than serving as hints for the clustering process. In alternative clustering, constraints are used to produce different partitions of a single datasets. Alternative clustering methods are not strictly CC methods, as they do not use constraints generated from a side source of information (an oracle), but from the current state of a partition in an iterative clustering process. The COALA takes a partition of a dataset as input, and aims to find a different high-quality partition using constraints. In order to make the obtained partition different from the one provided, CL constraints are created between the instances assigned to the same cluster in the original partition. COALA applies agglomerative hierarchical clustering considering two possible merges in each step, one involves the two closest instances and the other involves the two closest instances that do not violate a CL constraint. Which merge is performed depends on a parameter controlling the trade-off between quality and dissimilarity (with respect to the base partition) of the new partition. COALAcat is the categorical version of COALA. Contrary to COALA, ADFT does not take a partition of a dataset as input. The overall ADFT method can be summarized in five steps.  In the first step, a classic clustering method like K-Means is applied to partition the dataset. The second step characterizes this partition by means of ML and CL constraints and obtains a new distance metric based on them (using the CSI method). In the third step, this distance metric is taken as basis to compute an alternative distance measure, by obtaining singular values decomposition of the matrix that defines the metric is obtained, and by computing the Moore-Penrose pseudo-inverse of the stretcher matrix. This effectively flips the stretching and compressing dimensions. After that, the metric matrix is recomposed, multiplying its decomposition. In the fourth step, the newly learned metric is applied on the dataset to transform it, and in the fifth step, the classic clustering algorithm run to obtain the original partition is re-run to obtain a new (and different) one.

\subsubsection{CC based on clustering trees}

Methods which belong to this category perform clustering by using decision trees. ClusILC uses the Top-Down Induction (TDI) approach to build the clustering tree. While most TDI approaches are based on heuristics local to the node that is being built, ClusILC employs a global heuristic which measures the quality of the entire tree and which takes all instances of the dataset into account. ClusILC's heuristic  measures the average variance in the leafs of the tree (normalized by the overall dataset variance) and the proportion of overall violated constraints. ClusILC greedily searches for a tree that minimizes this heuristic using an iterative process that refines the current tree in every iteration by replacing one of its leaves with a subtree consisting of a new test node and two new leaves. This subtree is selected among a set of candidates generated procedurally based on the heuristic described above. 

\subsubsection{Maximum Margin CC}

Maximum Margin Clustering (MMC) uses the maximum margin principle adopted in the supervised learning paradigm. It tries to find the hyperplanes that partition the data into different clusters with the largest margins between them. CMMC includes constraints into MMC by adding concave-convex restrictions to the original MMC optimization problem. These restrictions behave as penalties for constraint violation. PMMC introduces a set of loss functions, featuring a strong penalty to partitions violating constraints while operating under the maximum margin principle at the same time. In order to do so, the classical definition of score used in MMC to determine cluster membership is modified to include high penalties for assignations violating constraints. Both CMMC and PMMC use the constrained concave-convex procedure (proposed in~\cite{zeng2011semi}) to solve the non-convex optimization problem they set to address the CC problem. TwoClaCMMC is proposed to overcome some shortcomings of the CMMC algorithm, although it is limited to two-class problems. It modifies the MMC objetive function with a penalty term which accounts for constraints violations. TwoClaCMMC differs from CMMC in the formulation of the penalty term. In TwoClaCMMC, the position of the hyperplane with respect to the instances involved in a violated constraint is taken into account, weighting the cost of violating such constraints with respect to said position.

\subsubsection{Feature Selection}

Clustering can be used to perform feature selection over high-dimensionality datasets. YLYM is a clustering method designed to perform feature selection making use of a constraint set. It is composed of three steps. The first two steps consist of the classic expectation and maximization steps from an EM optimization scheme, in which feature saliencies are computed in a completely unsupervised way. The third step is called the tuning step (T-step), which refines saliencies to minimize the feature-wise constraint violation measure, which is computed based on the Jensen-Shannon divergence. The three steps (expectation, maximization and tuning) are performed iteratively until they reach convergence. The proposed method outputs a partition of the dataset and the saliency of every feature.

\subsubsection{CC through MapReduce}

The MapReduce paradigm is used to address problems in the context of Big Data. The MapReduce paradigm consists of dividing the computational load associated with processing a dataset among multiple processing nodes, in order to decrease the time required to obtain results. MapReduce is based on two operations: (1) The Map operation processes inputs in the form of key-value pairs and generates intermediate key/value pairs received by the Reduce operation (2) The Reduce operation processes all intermediate values associated with the same intermediate key generated by Map. PCKMMR constitutes the first MapReduce approach to CC. It applies the MapReduce approach on the COP-K-Means algorithm. In order to do so, authors propose a Map function which calculates the distance of each instance to each centroid and assigns it to the one that minimizes this measure and does not violate any constraint. To avoid interdependence between mappers, the constraints are generated locally to each mapper in each call. The map function returns the centroid/instance pair. The Reduce operation takes as input all instances associated to a centroid and updates the value for that centroid by computing the average of those instances.

\subsubsection{SAT-based CC} \label{subsubsec:sat_based_cc}

SAT-based CC approaches formulate the CC problems in terms of logical clauses in conjunctive normal form. They apply general SAT solvers to find a solution, which can find solutions for any problems formulated as a satisfiability problem. This approaches to CC problem can include hard constraints in a very natural way, as well as they usually can handle many types of the constraints described in Section~\ref{sec:Clustering_with_BK}. ISL is limited to clauses of 2 literals. Several types of problems within the CC framework can be expressed in the form of sets of formulas in closed normal form (CNF), implying that they can be approached with ISL and therefore solved optimally. It is worth noting that ISL is limited to two-class problems ($k$=2), although it can find a solution to these problems in polynomial times (if such solutions exists). SGID performs agglomerative hierarchical clustering given a dataset and a set of constraints formulated in terms of logical clauses in its Horn's normal form. In order to produce a dendrogram, the clauses modeling its properties must also be given to SGID, which allows them to vary in the features of the dendrogram it produces. In JPBMS the declarative modeling principle of constrained programming is used to define a CC problem taking into account the constraint set, the description of the clusters and the clustering process itself. Traditionally, clustering algorithms proceed by iteratively refining queries until a satisfactory solution is found. JPBMS includes the stepwise refinement process in a natural way to focus on more interesting clustering solutions. JBMJ performs CC under the correlational clustering paradigm, where a labeled weighted undirected graph is given to perform clustering. The objective function of correlation clustering clusters the nodes of the graph in a way that minimizes the number of positive edges between different clusters and negative edges within clusters. JBMF formulates this problems in terms of clause satisfiability and applies MaxSAT to obtain an optimal solution. It is able to include several types of constraints by modifying the graph structure and applying specific SAT-translation procedure for some of them. Particularly, instance-level constraints are included by just setting the weight of the edges which connect ML related instances to $\infty$ and CL related instances to $-\infty$, there is no need to generate specific clauses.

\subsubsection{CC through constraint programming}

The CC problem can be addressed from the constraint programming (CP) point of view when the classic requirements of a clustering problem formulated as restrictions for a CP are extended with the restrictions regarding ML and CL constraints. TCK does this exactly: it models the classic clustering problem requirements, the constraints, and the clustering optimization criteria, including the within-cluster sums of squares (WCSS), as constraints to be solved by a general constraint programming solver. 3CP extends TCK in the sense that it is more general and it does not need the number of clusters to be specified, only the boundaries of the interval it lies in. Additionally, 3CP is capable of optimizing more than one clustering criteria at the same time, finding the minimal set of nondominated Pareto solutions. TKC(17) also extends TCK, differing from it in two key aspects. Firstly, It does not need the number of clusters to be specified, only bounds need to be given (as in 3CP). Secondly, three optimization criteria are modeled as CP constraints in TKC(17): minimizing the maximal diameter, maximizing the split between clusters, and minimizing WCSS. Besides, CMSSCCP optimizes the WCSS, but it does so via a global optimization constraint. A lower bound for this criterion is computed using dynamic programming, and a filtering algorithm is proposed to filter objective variables as well as decision variables. As usual, instance-level constraints are modeled in terms of constraint programming, along with the classic clustering problem requirements. JPBMS (also in Section~\ref{subsubsec:sat_based_cc}) uses the declarative modeling principles of CP to define the CC problem as a SAT problem, which can be solved with a general SAT solver. BBMSC formulates the CC problem in terms of an integer programming problem, rather than in terms of a constraint programming problem (which can be considered as a subtype of the former). It employs the same procedures as in constraint programming, incorporating ML and CL constraints via a weighted penalty term added to the base classic clustering restrictions, alongside with a regularization term to favor smoothness.

\subsubsection{CC through Column Generation}

In Column Generation (CG), the minimum sum-of-squares (MSS) problem for clustering is solved optimally. This is done by formulating the problem in terms of an integer linear programming problem. In it, a boolean matrix encoding all possible partitions in its columns is explored to find an optimal solution with respect to a cost function (the MSS in this case) that is applied to the boolean matrix by columns. In practice, the boolean matrix is too large to be computed, therefore it is incrementally built when searching for the optimal solution. The column generation approach derives a master problem from a reduced set of restrictions and iterates between two steps: solving the master problem and adding one or multiple candidate columns to the boolean matrix. A column is a candidate to be included in the restricted master problem if its addition improves the objective function. If no such column can be found, one is certain that the optimal solution of the restricted master problem is also the optimal solution of the full master problem. CCG includes constraints into this framework by modifying the MSS formulation. This way a new restriction which enforces all constraints to be satisfied is added to is classic form. Effectively, this is translated into the clustering process by removing all partitions violating any amount of constraints from the boolean matrix. This way, they are discarded from candidate solutions. The CG+PR+LS solves CC similarly to CCG, however it includes two extra steps: path-relinking algorithms are used to intensify and diversify the search in a group of solutions, and a LS procedure is used to locally improve the final solution.

\subsubsection{Information-maximization CC}

Information-maximization clustering techniques address the lack of objective model selection and parameter optimization strategies from which most other clustering techniques suffer. In it, a probabilistic classifier is learned so that some information measure between instances and clusters assignments is maximized. 3SMIC includes constraints into the information-maximization clustering algorithm SMIC. SMIC tries to learn the class-posterior probability in an unsupervised manner so that the mutual information between instances and their class labels (in the final partition) are maximized. Constraints are included into SMIC by modifying the SMI approximator, so that the inner product of the probabilities of constrained instances which belong to the same cluster is maximized in the case of ML and minimized in the case of CL. Furthermore 3SMIC includes a procedure to apply the transitive property of ML efficiently.

\subsubsection{CC through Heterogeneous Information Networks}

Heterogeneous Information Networks (HINs) are graphs which model real world entities and their relationships with objects and links, where objects can be of different types and whose links represent different kinds of relationships. HINs in which objects are described by attributes (features) are called attributed HIN or AHIN abbreviated. The challenge in AHINs is to perform clustering based not only on attribute similarity, but also based on link similarity. The former can be measured with a conventional distance measure, while the latter is measured with graph-oriented distance measures, such as shortest-path length and random-walk-based, although meta-path are commonly used as well. A meta-path is a sequence of node types that expresses a relation between two objects in an AHIN. SCAN includes constraints in AHIN by means of a penalty term in its similarity function. It computes the similarity of every node pair based on their attribute similarity and the connectedness of the nodes network. The former is obtained by an attribute similarity measure which considers coupling relationship among attributes, while the latter is derived based on the meta-paths connecting the object pair. This similarity is then penalized proportionally to the number of violated constraints. SCHAIN includes constraints in AHINs by first composing a similarity matrix that measures the similarity of every object pair based on the attribute similarity and the network connectedness. SCHAIN assigns a weight to each object attribute and meta-path in composing the similarity matrix. To take constraints into account, SCHAIN uses a penalty function which involves the generated weights and cluster memberships. It employs an iterative, staggered 2-step learning process to determine the optimal weights and cluster assignment as output.

\subsubsection{Incremental CC}

In incremental CC, a fixed set of instances and a variable constraint set are provided to perform clustering. Modifications over the constraint sets are given to incremental CC algorithms over time. These modifications include the addition and removal of constraints. The goal of Incremental CC is to efficiently
update a the current partition, without running the base clustering algorithm used to generate the initial partition. EICC is the only proposal belonging to this category. This method takes as input a single constraint at a time, and depending on the properties of the constraint, it will attempt to greedily optimize a given (not specified) objective function. If the constraint does not result in a significant improvement in the objective function, it is passed over and a new one is chosen (by the user). This algorithm only works when a set of preconditions are met, otherwise it will not produce a partition of the dataset, therefore it cannot be used to address a general constrained clustering problem.

\subsubsection{CC through affinity propagation}

In Affinity Propagation (AP) a binary grid factor-graph is used to perform clustering and to determine the most representative instances, which are called exemplars and can be compared with the notion of centroid. In order to do so, a number of hidden variables equal to the number of pairwise dissimilarities is defined. Afterwards, an iterative procedure is performed over the hidden variables, updating their value on the basis of their neighborhood variables and hyperparameters. The final value of the hidden variables determines the exemplar instances and the membership of the rest of instances with respect to the exemplars. SSAP includes constraints in AP by introducing a fictitious meta-points for every chunklet in the transitive closure and for every CL instance which is not part of a chunklet. The meta-points allow explicitly enforcing ML constraints and CL constraints, while they also propagate them.

\subsubsection{Spatial CC}

Spatial clustering is a variant of classical clustering in which instances are not points, but polygons. Classic clustering methods do not work well when applied to spatial clustering because they represent the polygons as points which summarize their features; which are not sufficiently representative of the polygons to obtain a good result. To overcome this problem, the CPSC algorithm is proposed, a spatial clustering algorithm based on the A* algorithm which is able to consider both instance-level constraints (ML and CL) and cluster-level constraints. In order to include the constraints into the clustering process, the heuristic function used by the A* algorithm is designed based on them. CPSC starts by selecting $k$ seeds from the data set (polygons), which will be the initial clusters and will grow through the iterative process. The seeds must be selected in such a way that each of them violates all ML constraints with respect to other seeds, thus ensuring that they will not be grouped in the same cluster. In addition to this, the seeds must satisfy all CL constraints between them. The best $k$ seeds are selected among those that meet these conditions. After that, the A* algorithm starts. The seeds are considered as the initial state, and the target clusters as the goal state. Each cluster is increased by adding polygons to its initial state one by one until it reaches its goal state. At each iteration, the best cluster (with respect to the heuristic) to be augmented and the best polygon to be augmented are selected. This is done to ensure that all clusters grow in parallel and not sequentially, which would affect compactness. The process continues until all polygons have been assigned to a cluster or until a deadlock state is reached, which can occur when two clusters compete for the same polygon and in which case there may be polygons which are not assigned to any cluster in the final partition. To overcome this problem, two other algorithms are proposed: CPSC*, which allows the user to relax the constraints to ensure that all polygons are assigned to a cluster, ensuring convergence, and CPSC*-PS (polygon split), which also allows polygons to be split when strictly necessary.

\subsubsection{Probabilistic CC}

In probabilistic clustering a probabilistic model is used to describe relationships between instances and their cluster memberships. Constraints are included into this framework by also describing them in terms of probabilities. RHWL is an active CC method which uses a basic probabilistic CC procedure as the clustering algorithm. It takes advantage of the probabilities computed by this procedure to later use them to decide the pair of instances to query to the oracle. DCPR uses an objective function maximizing the likelihood of the observed constraint labels composed of two terms: (1) the first is the conditional entropy of the empirical cluster label distribution, which is maximized when the cluster labels are uniformly distributed (the clusters are balanced). (2) the second term is the conditional entropy of instance cluster labels for the unlabeled instances, which is minimized when the formed clusters have large separation margin and high confidence for the cluster memberships of unconstrained instances. A variational EM optimization scheme is used to optimized the proposed objective function.

\subsection{Constrained Distance Transformation}

Distance transformation methods usually take a standard distance measure as their basis (like as the Euclidean distance) and parameterize it. By modifying this parameters, the distance measure is transformed for it to be adapted to the data and the constraints. Generally, the goal of these methods is to learn a new distance metric bringing ML instances together and setting CL instances apart. Constrained Distance Transformation (CDT) methods are presented in Table~\ref{tab:tax:ctd}.

Methods such as CSI and Xiang's learn the weights of a matrix by parameterizing a family of Mahalanobis distances. RCA also learns a Mahalanobis metric. In order to do so it changes the feature space used for data representation by assigning high weights to relevant dimensions and low weights to irrelevant dimensions by means of a global linear transformation. The relevant dimensions are estimated using chunklets. ERCA extends RCA to include CL. It does so by computing a matrix that optimizes the between-class scatter and combining it with the matrix optimizing the within-class scatter. DCA is another extension over RCA to include CL. It does so by looking for a linear transformation which results in an optimal distance metric by maximizing the variance between chunklets and minimizing the variance between instances in the same chunklet. KDCA uses the kernel trick to learn a nonlinear metric distance under the same principles of DCA. MSSB is a modification over RCA that uses a data-dependent regularizer term to avoid the drawbacks that weighting discrimination brings to RCA. From all methods using Mahalanobis distances parameterization, the ITML and A-ITML-K-Means approaches are the only ones using Information Theoretic Metric Learning as its base framework. ITML learns a constrained distance metric by learning a positive-definite matrix that parameterizes a Mahalanobis distance. This matrix is regularized to be as close as possible to a given Mahalanobis distance function, parameterized by another auxiliary matrix. The distance between these two matrices can be quantified via an information-theoretic approach, so it can be computed as the relative entropy between their corresponding multivariate Gaussians. Instance-level constraints are included as linear constraints to the optimization problem, which results in a particular case of the Bregman divergence. Therefore, it can be optimized by the Bregman's method. A-ITML-K-Means is an active clustering with constraints setup for the ITML approach that uses K-means as its clustering algorithm.

It uses a reduced set of selected constraints to perform ITML and applies classic clustering (K-Means) with the newly learned metric in an active clustering setup. ITML learns a Mahalanobis distance metric under a given set of constraints and instances by minimizing the LogDet divergence between the original distance matrix and an objective distance matrix that in built on the basis of constraints.

HMRF-K-Means approaches the CC problem from a hybrid probabilistic framework based on Hidden Markov Random Fields (HMRF), which is developed to find an EM-optimizable objective function derived from the posterior energy, which is defined by the HMRF. This objective function combines an adaptive distance measure, such as the Bregman divergence or directional similarity measures, and a constraint-violation penalty term. The later is controlled by a scaling function that assign more relevance to ML constraints relating distant instances and CL constraints relating close instances. Comraf uses a combinatorial MRF (an MRF in which at least one node is a combinatorial random variable). The Comraf model can be applied to classic clustering by searching for cliques in the Comraf graph and using the mutual information as a potential function. This graph is built on the basis of the interactions between the combinatorial random variables. Comraf can be extended to constrained clustering by incorporating weighted constraints as a penalty term in the objective function that Comraf optimizes.

Other methods such as MPCK-Means iteratively compute cluster-local weights for every feature, effectively creating a new distance metric for every cluster, which can be applied and updated during the clustering process in an EM scheme. LLMA also performs metric learning through locally linear transformations, achieving global consistency via interactions between adjacent local neighborhoods.

SMR builds the graph Laplacian matrix based on pairwise similarities. This matrix is then used to regularize a non-parametric kernel learning procedure in which the learned kernel matrix is forced to be consistent with both pairwise similarities and the constraint sets by minimizing the regularizer that is based on the Laplacian graph. MSBSBS is similar to SMR. However it uses the Karush-Kuhn-Tucker conditions to learn the non-parametric kernel. MSBSBS(10) is an improvement over MSBSBS which combines both the constraint set and the topological structure of the data to learn a non-linear metric. CDJPBY uses a weighted sum to combine multiple kernels. It includes an optimization criterion that allows it to automatically estimate the optimal parameter of the composite Gaussian kernels directly from the data and the constraints.

RDF transforms the metric learning problem into a binary class classification problem and employs random forests as the underlying representation. Constraints are included by replacing the original distance function with a feature map function which transforms each constrained instance, changing their location to reflect the information contained in the constraint set. The transformed data is used as training data for a random forest that evaluates each instance pair. Each tree from the random forest independently classifies the pair as similar or dissimilar, based on the leaf node at which the instance-pair arrives. SSMMHF uses a random forest-based strategy as well. It first builds a model of the data by computing a forest of semi-random cluster hierarchies. Each tree is generated applying a semi-randomized binary semi-supervised maximum-margin clustering (MCC) algorithm iteratively. This way, each tree encodes a particular model of the full semantic structure of the data, so the full structure of the tree can be considered as a weak metric. A final metric model can be produced by merging the output of the forest described above. Constraints are included by modifying the MMC procedure, producing Semi-Supervised MMC. This results in a method that seeks to simultaneously maximize the cluster assignment margin of each point (as in unsupervised MMC) and an additional set of margin terms reflecting the satisfaction of each pairwise constraint.

SCKMM firstly estimates the optimal value for the parameter of a Gaussian kernel by ascending gradient and obtains an initial partition using PCBKM, which is a modification of the classic K-Means algorithm to include constraints. After this, a distance measure is obtained using the assignments of the last partition, which is used by PCBKM to produce a new partition. SCKMM iterates between steps two and three until it converges.

3SHACC includes a metric learning step in 2SHACC. In first place, it determines the relevance of every constraint in a completely unsupervised manner. Said relevances are used as constraint weights by the Weighted-Learning from Side Information (WLSI) DML method, which is a weighted version of CSI.

\begin{table}[!h]
	\centering
	\setlength{\tabcolsep}{7pt}
	\renewcommand{\arraystretch}{1.3}
	\resizebox{\textwidth}{!}{
		\begin{tabular}{c l l c c c c c c}
		\hline
		$S_{\mathcal{A}}$ & Acronym & Full Name & Penalty & ML & CL & Hybrid & Year & Ref. \\
		\hline
        \cellcolor{scorecolor!72.11297761244903}19.17 & CSI & \makecell[l]{Clustring with Side Information} & $\times$ & Soft & Soft & \makecell{---} & 2002 & \cite{xing2002distance} \\
\hhline{--}
\cellcolor{scorecolor!77.51238992538471}20.61 & MPCK-Means & \makecell[l]{Metric Pairwise Constrained K-Means} & \checkmark & Soft & Soft & \makecell{Penalty-based Methods} & 2003 & \cite{basu2003comparing,bilenko2004integrating} \\
\hhline{--}
\cellcolor{scorecolor!33.75117528492475}8.97 & HMRF-K-Means & \makecell[l]{Hidden Markov Random Fields - K-Means} & \checkmark & Soft & Soft & \makecell{Penalty-based Methods} & 2004 & \cite{basu2004probabilistic} \\
\hhline{--}
\cellcolor{scorecolor!60.1510150533486}15.99 & DistBoost & \makecell[l]{---} & $\times$ & Soft & Soft & \makecell{---} & 2004 & \cite{hertz2004boosting} \\
\hhline{--}
\cellcolor{scorecolor!49.576443192564184}13.18 & LLMA & \makecell[l]{Locally Linear Metric Adaptation} & $\times$ & Soft & Soft & \makecell{---} & 2004 & \cite{chang2004locally} \\
\hhline{--}
\cellcolor{scorecolor!60.22588784963381}16.01 & RCA & \makecell[l]{Relevant Components Analysis} & $\times$ & Soft & --- & \makecell{Constrained Data Space Transformation} & 2005 & \cite{bar2005learning} \\
\hhline{--}
\cellcolor{scorecolor!100.0}26.59 & ERCA & \makecell[l]{Extended Relevant Components Analysis} & $\times$ & Soft & Soft & \makecell{---} & 2006 & \cite{yeung2006extending} \\
\hhline{--}
\cellcolor{scorecolor!72.34844661359207}19.24 & DCA & \makecell[l]{Discriminative Component Analysis} & $\times$ & Soft & Soft & \makecell{---} & 2006 & \cite{hoi2006learning} \\
\hhline{--}
\cellcolor{scorecolor!60.48584855819217}16.08 & KDCA & \makecell[l]{Kernel - Discriminative Component Analysis} & $\times$ & Soft & Soft & \makecell{---} & 2006 & \cite{hoi2006learning} \\
\hhline{--}
\cellcolor{scorecolor!21.372398631722334}5.68 & Comraf & \makecell[l]{Combinatorial Markov Random Fields} & --- & Soft & Soft & \makecell{---} & 2006 & \cite{bekkerman2006combinatorial} \\
\hhline{--}
\cellcolor{scorecolor!75.08241689590855}19.96 & DYYHC & \makecell[l]{names} & $\times$ & Soft & --- & \makecell{---} & 2007 & \cite{yeung2007kernel} \\
\hhline{--}
\cellcolor{scorecolor!24.88519251042971}6.62 & ITML & \makecell[l]{Information-Theoretic Metric Learning} & $\times$ & Soft & Soft & \makecell{---} & 2007 & \cite{davis2007information} \\
\hhline{--}
\cellcolor{scorecolor!41.41691592007154}11.01 & SMR & \makecell[l]{names} & $\times$ & Soft & Soft & \makecell{---} & 2007 & \cite{hoi2007learning} \\
\hhline{--}
\cellcolor{scorecolor!56.615366817203096}15.05 & Xiang's & \makecell[l]{---} & $\times$ & Soft & Soft & \makecell{---} & 2008 & \cite{xiang2008learning} \\
\hhline{--}
\cellcolor{scorecolor!44.447939631163344}11.82 & MSSB & \makecell[l]{names} & $\times$ & Soft & Soft & \makecell{---} & 2009 & \cite{baghshah2009metric} \\
\hhline{--}
\cellcolor{scorecolor!63.81355737646045}16.97 & MSBSBS & \makecell[l]{names} & $\times$ & Soft & Soft & \makecell{---} & 2010 & \cite{baghshah2010kernel} \\
\hhline{--}
\cellcolor{scorecolor!24.481153622926527}6.51 & SCKMM & \makecell[l]{Semi-supervised Clustering Kernel Method based on Metric learning} & $\times$ & Soft & Soft & \makecell{Cluster Engine-adapting Methods} & 2010 & \cite{yin2010semi} \\
\hhline{--}
\cellcolor{scorecolor!66.85816376554259}17.78 & MSBSBS(10) & \makecell[l]{names} & $\times$ & Soft & Soft & \makecell{---} & 2010 & \cite{baghshah2010non} \\
\hhline{--}
\cellcolor{scorecolor!18.860572845452005}5.01 & LRML & \makecell[l]{Laplacian Regularized Metric Learning} & $\times$ & Soft & Soft & \makecell{---} & 2010 & \cite{hoi2010semi} \\
\hhline{--}
\cellcolor{scorecolor!70.07259063759888}18.63 & LRKL & \makecell[l]{Low-Rank Kernel Learning} & $\times$ & Soft & Soft & \makecell{---} & 2011 & \cite{baghshah2011learning} \\
\hhline{--}
\cellcolor{scorecolor!27.37597411910461}7.28 & CDJPBY & \makecell[l]{names} & $\times$ & Soft & Soft & \makecell{---} & 2011 & \cite{domeniconi2011composite} \\
\hhline{--}
\cellcolor{scorecolor!26.364587420153068}7.01 & SNN & \makecell[l]{Similarity Neural Networks} & $\times$ & Soft & Soft & \makecell{Classic Neural Network-based CC} & 2012 & \cite{maggini2012learning} \\
\hhline{--}
\cellcolor{scorecolor!29.45392473014754}7.83 & RFD & \makecell[l]{Random Forest Distance} & $\times$ & Soft & Soft & \makecell{---} & 2012 & \cite{xiong2012random} \\
\hhline{--}
\cellcolor{scorecolor!21.131327337379744}5.62 & A-ITML-K-Means & \makecell[l]{Active - Information Theoric Metric Learning - K-Means} & $\times$ & Soft & Soft & \makecell{Active Clustering with Constraints} & 2013 & \cite{rao2013semi} \\
\hhline{--}
\cellcolor{scorecolor!26.574262388314136}7.07 & LSCP & \makecell[l]{Learning Similarity of Constraint Propagation} & $\times$ & Soft & Soft & \makecell{Constraint Propagation} & 2015 & \cite{fu2015local} \\
\hhline{--}
\cellcolor{scorecolor!39.097811010236704}10.40 & SSMMHF & \makecell[l]{Semi-Supervised Max-Margin Hierarchy Forest} & $\times$ & Soft & Soft & \makecell{---} & 2016 & \cite{johnson2016semi} \\
\hhline{--}
\cellcolor{scorecolor!57.70831755684554}15.34 & AMH-L & \makecell[l]{names} & $\times$ & Soft & Soft & \makecell{---} & 2020 & \cite{abin2020learning} \\
\hhline{--}
\cellcolor{scorecolor!57.70831755684554}15.34 & AMH-NL & \makecell[l]{names} & $\times$ & Soft & Soft & \makecell{---} & 2020 & \cite{abin2020learning} \\
\hhline{--}
\cellcolor{scorecolor!84.69146326229607}22.52 & 3SHACC & \makecell[l]{3 -Stages Hybrid Agglomerative Constrained Clustering} & $\times$ & Soft & Soft & \makecell{Hierarchical CC} & 2022 & \cite{gonzalez20213shacc}  \\
        \hline
		\end{tabular}}
	\caption{Feature table for CDT methods.}
	\label{tab:tax:ctd}
\end{table}

\subsection{Distance Matrix Modification}

In Distance Matrix Modification (DMM) methods, the CC problem is approached from the DML point of view. Nevertheless these methods work directly with the distance matrix. Their goal is to modify the entries of the distance matrix for it to reflect the information contained in the constraint set, once again resulting in ML instances being brought closer together and CL instances being set apart. These methods do not provide neither a new distance metric nor a new data space, although these two can be obtained from said distance matrix with classic DML techniques.

\subsubsection{Constraint Propagation}

In constraint propagation methods, entries in the distance matrix which correspond to constrained instances are usually first modified. Then, those changes are propagated to the rest of the matrix to a variable extent and using different strategies. Table~\ref{tab:tax:dmm_cp} gathers methods which use constraint propagation to perform CC.

\begin{table}[!h]
	\centering
	\setlength{\tabcolsep}{7pt}
	\renewcommand{\arraystretch}{1.3}
	\resizebox{\textwidth}{!}{
		\begin{tabular}{c l l c c c c c c}
		\hline
		$S_{\mathcal{A}}$ & Acronym & Full Name & Penalty & ML & CL & Hybrid & Year & Ref. \\
		\hline
        \cellcolor{scorecolor!83.12008933130868}19.16 & CCL & \makecell[l]{Constrained Complete-Link} & $\times$ & Soft & Soft & \makecell{---} & 2002 & \cite{klein2002instance} \\
\hhline{--}
\cellcolor{scorecolor!54.106863346626064}12.47 & LCPN & \makecell[l]{names} & $\times$ & Soft & Soft & \makecell{Non Graph-based} & 2008 & \cite{lu2008constrained} \\
\hhline{--}
\cellcolor{scorecolor!35.64244060038211}8.22 & Lo-NC & \makecell[l]{Local Normalized Cut} & $\times$ & Soft & Soft & \makecell{---} & 2009 & \cite{yan2009pairwise} \\
\hhline{--}
\cellcolor{scorecolor!51.861668720171416}11.96 & E$^2$CP & \makecell[l]{Exhaustive and Efficient Constraint Propagation} & $\times$ & Soft & Soft & \makecell{---} & 2010 & \cite{lu2010constrained,lu2013exhaustive} \\
\hhline{--}
\cellcolor{scorecolor!23.085778181623283}5.32 & SRCP & \makecell[l]{Symmetric graph Regularized Constraint Propagation} & $\times$ & Soft & Soft & \makecell{---} & 2011 & \cite{fu2011symmetric} \\
\hhline{--}
\cellcolor{scorecolor!30.086494596185638}6.94 & MMCP & \makecell[l]{Multi-Modal Constraint Propagation} & $\times$ & Soft & Soft & \makecell{---} & 2011 & \cite{fu2011multi} \\
\hhline{--}
\cellcolor{scorecolor!24.889441982782348}5.74 & LCP & \makecell[l]{Local Constraint Propagation} & $\times$ & Soft & Soft & \makecell{---} & 2012 & \cite{he2012constrained} \\
\hhline{--}
\cellcolor{scorecolor!20.08995599034557}4.63 & SSL-EC & \makecell[l]{Semi-Supervised Learning based on Exemplar Constraints} & $\times$ & Soft & Soft & \makecell{Active Constraint Acquisition} & 2012 & \cite{wang2012exemplars} \\
\hhline{--}
\cellcolor{scorecolor!22.873931378094277}5.27 & UCP & \makecell[l]{Unified Constraint Propagation} & $\times$ & Soft & Soft & \makecell{Inter-View Constrained} & 2013 & \cite{lu2013unified} \\
\hhline{--}
\cellcolor{scorecolor!17.46852597268887}4.03 & MSCP & \makecell[l]{Multi-Source Constraint Propagation} & $\times$ & Soft & Soft & \makecell{Inter-View Constrained} & 2013 & \cite{lu2013exhaustive} \\
\hhline{--}
\cellcolor{scorecolor!24.373072167753165}5.62 & A-ITML-K-Means & \makecell[l]{Active - Information Theoric Metric Learning - K-Means} & $\times$ & Soft & Soft & \makecell{Active Clustering with Constraints} & 2013 & \cite{rao2013semi} \\
\hhline{--}
\cellcolor{scorecolor!44.72340971796131}10.31 & CAF & \makecell[l]{Constraints As Features} & $\times$ & Soft & Soft & \makecell{Constrained Data Space Transformation} & 2013 & \cite{asafi2013constraints} \\
\hhline{--}
\cellcolor{scorecolor!17.51485456150215}4.04 & ACC(14) & \makecell[l]{Adaptive Constrained Clustering} & $\times$ & Soft & Soft & \makecell{---} & 2014 & \cite{he2014constrained} \\
\hhline{--}
\cellcolor{scorecolor!30.651004769088136}7.07 & LSCP & \makecell[l]{Learning Similarity of Constraint Propagation} & $\times$ & Soft & Soft & \makecell{Constrained Distance Transformation} & 2015 & \cite{fu2015local} \\
\hhline{--}
\cellcolor{scorecolor!19.282457803481563}4.45 & CPSNMF & \makecell[l]{Constrained Propagation for Semi-supervised Nonnegative Matrix Factorization} & $\times$ & Soft & Soft & \makecell{Non-negative Matrix Factorization CC} & 2016 & \cite{wang2015semi} \\
\hhline{--}
\cellcolor{scorecolor!100.0}23.05 & ISSCE & \makecell[l]{Incremental Semi-Supervised Clustering Ensemble} & $\times$ & Soft & Soft & \makecell{Constrained Pool Generation} & 2016 & \cite{yu2016incremental} \\
\hhline{--}
\cellcolor{scorecolor!87.83783783783782}20.25 & RSSCE & \makecell[l]{Random Subspace based Semi-supervised Clustering Ensemble} & $\times$ & Soft & Soft & \makecell{Constrained Pool Generation} & 2016 & \cite{yu2016incremental} \\
\hhline{--}
\cellcolor{scorecolor!33.53394564078525}7.73 & C$^3$ & \makecell[l]{Constrained Community Clustering} & $\times$ & Soft & Soft & \makecell{---} & 2016 & \cite{xu2016improving} \\
\hhline{--}
\cellcolor{scorecolor!33.9815192725249}7.83 & CESCP & \makecell[l]{Clustering Ensemble based on Selected Constraint Projection } & $\times$ & Soft & Soft & \makecell{Constrained Pool Generation} & 2018 & \cite{yu2018semi} \\
\hhline{--}
\cellcolor{scorecolor!46.14368143468705}10.64 & DCECP & \makecell[l]{Double-weighting Clustering Ensemble with Constraint Projection} & $\times$ & Soft & Soft & \makecell{Constrained Pool Generation} & 2018 & \cite{yu2018semi} \\
\hhline{--}
\cellcolor{scorecolor!22.587195136277955}5.21 & PCPDAMR & \makecell[l]{Pairwise Constraint Propagation with Dual Adversarial Manifold Regularization} & $\times$ & Soft & Soft & \makecell{---} & 2020 & \cite{jia2020pairwise} \\
\hhline{--}
\cellcolor{scorecolor!71.60214027704119}16.51 & 2SHACC & \makecell[l]{2 -Stages Hybrid Agglomerative Constrained Clustering} & $\times$ & Soft & Soft & \makecell{Hierarchical CC} & 2020 & \cite{gonzalez2020agglomerative} \\
\hhline{--}
\cellcolor{scorecolor!18.045797771589882}4.16 & ILMCP & \makecell[l]{Instance Level Multi-modal Constraint Propagation} & $\times$ & Soft & Soft & \makecell{---} & 2021 & \cite{li2021multi}  \\
        \hline
		\end{tabular}}
	\caption{Feature table for DMM - Constraint Propagation methods.}
	\label{tab:tax:dmm_cp}
\end{table}

The most simple approach in constraint propagation consists of simply setting distances between ML instances to 0 and to a high value for CL instances in the distance matrix. Afterwards, the all-pairs-shortest-path algorithm is run to propagate the changes to the rest of entries. Methods such as CCL, 2SHACC and CAF use this approach.

E$^2$CP propagates constraints in the distance matrix by taking each of its columns as the initial configuration of a two-class semi-supervised learning problem with respect to the instance associated to the column. The positive class contains the examples which should appear in the same cluster as the instance associated to the column, and the negative class contains the examples that should not. In this way, the constraint propagation problem can be decomposed into a number of subproblems equal to the size of the dataset. After that, the same process is repeated, this time taking rows instead of columns. MSCP extends E$^2$CP to consider multi-source data. It decomposes the problem into a series of two-source constraint propagation subproblems, which can be transformed into solving a Sylvester matrix equation, viewed as a generalization of the Lyapunov matrix equation. SRCP also uses decomposition to propagate constraints, although it decomposes the problem taking into account the full dataset, not only columns or rows. It exploits the symmetric structure of pairwise constraints to develop a constraint propagation approach based on symmetric graph regularization. MMCP propagates constraints in rows and columns separately without decomposition using multi-graph propagation methods.

LCP propagates the influence of the constraints to the unconstrained instances in the dataset proportionally to their similarity with the constrained data. LCP first determines the proportion in which constrained instances influence unconstrained instances using a previously proposed label propagation procedure. Then, intermediate structures, called constrained communities, are defined to include the factional instances that are affected by a constrained instance (including itself). These structures are used to find the range of influence of each constraints without any parameter estimation. ACC(14) performs constraint propagation the same way as LCP, the only difference being that ACC(14) allows overlapping between constrained communities. C$^3$ propagates constraints proportionally in constrained communities as well, although directly performing clustering on them through their indicator matrix.

PCPDAMR uses both the similarity and the dissimilarity (distance) matrix to perform constraint propagation. It does so to emphasize the difference between ML and CL constraints, which are usually encoded in the same matrix. The constraint propagation is carried out via manifold embedding, in which the inherent manifold structure among the data instances is mapped to their similarity/dissimilarity codings. A regularization term to consider adversarial relations between the two matrices is used to enhance the discriminability of propagated constraints.

Many methods simply use the E$^2$CP as an intermediate step performed between other CC-related operations. For example, RSSCE, CESCP and DCECP are all ensemble clustering method which use E$^2$CP as the CC method to generate different partitions using the random subspace technique. CPSNMF simply performs E$^2$CP and a penalty-based version of NMF afterwards.

More exotic approaches can be found in LCPN or Lo-NC. LCPN takes the affinity matrix as the covariance matrix of a parameterized Gaussian process with mean 0, effectively connecting the spatial locations of constrained instances and propagating a positive or negative affinity value (depending on whether it is a ML or CL constraint) in those locations. Lo-NC performs a space-level generalization of pairwise constraints by locally propagating the information contained in the constraint set.

\subsubsection{Matrix Completion}

Matrix completion techniques, gathered in Table~\ref{tab:tax:dmm_mc}, are used to fill gaps in relational matrices. Its key concept is found in how to build the matrix over which matrix completion is applied. The reason is that matrix completion algorithm itself does not need to be specifically designed for CC, but it only need a matrix with gaps to be filled, so low-rank matrices are preferred.

\begin{table}[!h]
	\centering
	\setlength{\tabcolsep}{7pt}
	\renewcommand{\arraystretch}{1.3}
	\resizebox{\textwidth}{!}{
		\begin{tabular}{c l l c c c c c c}
		\hline
		$S_{\mathcal{A}}$ & Acronym & Full Name & Penalty & ML & CL & Hybrid & Year & Ref. \\
		\hline
        \cellcolor{scorecolor!61.492533469706935}6.79 & MCCC & \makecell[l]{Matrix Completion based Constraint Clustering} & $\times$ & Soft & Soft & \makecell{---} & 2013 & \cite{yi2013semi} \\
\hhline{--}
\cellcolor{scorecolor!40.3962028924438}4.46 & STSC & \makecell[l]{Self-Taught Spectral Clustering} & $\times$ & Soft & Soft & \makecell{Non Graph-based} & 2014 & \cite{wang2014self} \\
\hhline{--}
\cellcolor{scorecolor!81.09653295833759}8.95 & MCPCP & \makecell[l]{Matrix Completion - Pairwise Constraint Propagation} & $\times$ & Soft & Soft & \makecell{Inter-View Constrained} & 2015 & \cite{yang2014matrix} \\
\hhline{--}
\cellcolor{scorecolor!31.580743961664325}3.49 & LMRPCP & \makecell[l]{Low-rank Matrix Recovery based Pairwise Constraint Propagation} & $\times$ & Soft & Soft & \makecell{---} & 2015 & \cite{fu2015pairwise} \\
\hhline{--}
\cellcolor{scorecolor!100.0}11.04 & MVMC & \makecell[l]{Multi-View Matrix Completion} & $\times$ & Soft & Soft & \makecell{Intra-View Constrained \& Spectral CC} & 2017 & \cite{zhao2017multi}  \\
        \hline
		\end{tabular}}
	\caption{Feature table for DMM - Matrix Completion methods.}
	\label{tab:tax:dmm_mc}
\end{table}

In CC, the constraint matrix is usually a low-rank matrix, perfectly suitable for constraint propagation. STSC takes advantage of this and proposes the first approach to CC from the Self-taught learning paradigm, where side information is generated by the same algorithm that will make use of it later without the need of an oracle (or human). STSC can augment the set of constraints taking advantage of the low-rank nature of the constraint matrix via matrix completion. Matrix completion methods are able to recover low-rank matrices with high probability by using only a small number of observed entries. STSC performs self-taught constraint augmentation and constrained spectral clustering in an iterative manner. Constraint augmentation is performed via matrix completion over a combination of the low-rank constraint matrix and the affinities in the affinity graph. MCPCP also performs matrix completion over the constraint matrix, with each entry in the matrix being a real number that represents the relevance of the two corresponding instances. MCPCP aims to learn the full relation matrix by using a matrix completion algorithm and derives an indicator matrix from it.

Other methods aim to reconstruct and artificially made low-rank similarity matrix. This is the case of MCCC, which assigns similarity 1 for any pair of instances in the same cluster and 0 otherwise, based on the given constraints and the dataset. It can be proven that this is equivalent to finding the best data partition. A convex optimization problem, whose global solution can be efficiently obtained, can be used to solve this problem. Please note that the aim of MCCC is reconstructing the similarity matrix taking constraints into account, not the constraint matrix.

Lastly, LMRPCP performs matrix completion within a transductive learning framework. These approaches make the data matrix and the label matrix jointly low-rank and simultaneously apply a matrix completion algorithms to them. LMRPCP assumes that the data matrix is a clean low-rank matrix, while the constraint matrix is considered to be noisy and low-rank. The resulting problem is a matrix completion problem which can be solved with an augmented Lagrangian multiplier algorithm. The generated constraints are used to adjust pairwise similarities, over which classic spectral clustering is performed to obtain a partition.

\subsection{Constrained Data Space Transformation} \label{subsec:tax:cddt}

Constrained Data Space Transformation (CDST) techniques (gathered in Table~\ref{tab:tax:cdst}) seek to transform the space in which the data is embeded so that the new space can include the information contained in the constraint set. This usually involves reducing or augmenting the number of dimensions of said space. The majority of methods which belong to this category seek to reduce the number of dimensions, thus summarizing (and sometimes losing) information from the original data space. Other methods augment the number of dimensions based on the constraints and without any loss of information, although they produce a larger dataset which is usually harder to process by partitional methods.

\begin{table}[!h]
	\centering
	\setlength{\tabcolsep}{7pt}
	\renewcommand{\arraystretch}{1.3}
	\resizebox{\textwidth}{!}{
		\begin{tabular}{c l l c c c c c c}
		\hline
		$S_{\mathcal{A}}$ & Acronym & Full Name & Penalty & ML & CL & Hybrid & Year & Ref. \\
		\hline
        \cellcolor{scorecolor!90.31494602571198}16.01 & RCA & \makecell[l]{Relevant Components Analysis} & $\times$ & Soft & --- & \makecell{Constrained Distance Transformation} & 2005 & \cite{bar2005learning} \\
\hhline{--}
\cellcolor{scorecolor!99.73314658561375}17.68 & SCREEN & \makecell[l]{Semi-supervised Clustering method based on spheRical k-mEans via fEature projectioN} & $\times$ & Hard & Soft & \makecell{---} & 2007 & \cite{tang2007enhancing} \\
\hhline{--}
\cellcolor{scorecolor!90.64272514558233}16.07 & PCP & \makecell[l]{Pairwise Constraint Propagation} & $\times$ & Soft & Soft & \makecell{---} & 2008 & \cite{li2008pairwise} \\
\hhline{--}
\cellcolor{scorecolor!46.34046136256711}8.22 & RLC-NC & \makecell[l]{Constrained Normalized Cut} & $\times$ & Soft & Soft & \makecell{---} & 2009 & \cite{yan2009pairwise} \\
\hhline{--}
\cellcolor{scorecolor!46.76509035949346}8.29 & GBSSC & \makecell[l]{Graph-Based Semi-Supervised Clustering} & $\times$ & Hard & Soft & \makecell{Graph-based} & 2010 & \cite{yoshida2010performance,yoshida2010graph,yoshida2011pairwise,yoshida2014graph} \\
\hhline{--}
\cellcolor{scorecolor!100.0}17.73 & CPSSAP & \makecell[l]{Constraint Projections Semi-Supervised Affinity Propagation} & $\times$ & Soft & Soft & \makecell{---} & 2012 & \cite{wang2012constraint} \\
\hhline{--}
\cellcolor{scorecolor!58.14706863859447}10.31 & CAF & \makecell[l]{Constraints As Features} & $\times$ & Soft & Soft & \makecell{Constraint Propagation} & 2013 & \cite{asafi2013constraints} \\
\hhline{--}
\cellcolor{scorecolor!65.14304094298038}11.55 & MPHS-Linear & \makecell[l]{Mid-Perpendicular Hyperplane Similarity - Linear} & $\times$ & Soft & Soft & \makecell{---} & 2013 & \cite{gao2013learning} \\
\hhline{--}
\cellcolor{scorecolor!65.14304094298038}11.55 & MPHS-Gauss & \makecell[l]{Mid-Perpendicular Hyperplane Similarity - Gaussian} & $\times$ & Soft & Soft & \makecell{---} & 2013 & \cite{gao2013learning} \\
\hhline{--}
\cellcolor{scorecolor!65.14304094298038}11.55 & MPHS-PCP & \makecell[l]{Mid-Perpendicular Hyperplane Similarity - Pairwise Constraint Propagation} & $\times$ & Soft & Soft & \makecell{---} & 2013 & \cite{gao2013learning} \\
\hhline{--}
\cellcolor{scorecolor!39.08920421574006}6.93 & CNP-K-Means & \makecell[l]{Constraint Neighborhood Projections - K-Means} & $\times$ & Soft & Soft & \makecell{Non Graph-based} & 2014 & \cite{wang2014constraint} \\
\hhline{--}
\cellcolor{scorecolor!24.983801851623213}4.43 & CCC-GLPCA & \makecell[l]{Convex Constrained Clustering - Graph-Laplacian PCA} & $\times$ & Soft & Soft & \makecell{---} & 2018 & \cite{jia2018convex} \\
\hhline{--}
\cellcolor{scorecolor!43.66923356139157}7.74 & DP-GLPCA & \makecell[l]{Dissimilarity Propagation-guided - Graph-Laplacian Principal Component Analysis} & $\times$ & Soft & Soft & \makecell{---} & 2021 & \cite{jia2020constrained}  \\
        \hline
		\end{tabular}}
	\caption{Feature table for CDST methods.}
	\label{tab:tax:cdst}
\end{table}

Most methods in this category perform dimensionality reduction over the original dataset. Some of them are based on the classic PCA algorithm: RCA, CCC-GLPCA and DP-GLPCA are some examples of this. RCA is an ML constrained version of the classic PCA algorithm. It seeks to identify and down-scale global unwanted variability within the data. In order to do so, it changes the feature space used for data representation by assigning high weights to relevant dimensions and low weights to irrelevant dimensions by means of a global linear transformation. The relevant dimensions are estimated using chunklets. CCC-GLPCA includes a regularized term in the objective function of GLPCA (a Graph-Laplacian variant of PCA) to include constraints. This regularizes the similarity between the multiple low-dimensional representations used by GLPCA. DP-GLPCA simply performs classic GLPCA over the modified distance matrix to set distances between instances related by ML and CL to 0 or 1, respectively. Besides, it includes a dissimilarity regularizer which emphasizes CL to expand their influence.

Other methods use diverse procedures to perform dimensionality reduction taking constraints into account. SCREEN includes a step where a constraint-guided feature projection method (called SCREEN$_{PROJ}$) is used to project the original data in a low-dimensional space. RLC-NC seeks a low-dimensional representation of the data through orthogonal factorizations in which the clustering structure defined by the prior knowledge is strengthened. GBSSC first obtains the chunklet graph, over which a Laplacian process is applied for the graph to reflect CL constraints. The entire resulting graph is then projected onto a lower-dimensional space. CPSSAP uses the constraint projection methods to produce a faithful representation of the constraint set in a lower-dimensional space. The affinity matrix is computed on the basis of the new data space and a classic affinity propagation algorithm is used to produce a partition of the dataset. CNP-K-Means seeks to project the original dataset into a lower-dimensional space, preserving the information contained in the constraint set. In order to do so, it defines a neighborhood for every instance based on a parameterized radius value, used to propagate the influence of constraints from constrained instances without augmenting the constraint set.

The three variants of MPHS perform dimensionality reduction in a more exotic constraint-oriented way. It is inspired by the maximum margin hyperplane of SVM. Intuitively, CL constraints can be used to define hyperplanes between pairs of instances. Given two instances with related by CL constraint, we can compute its mid-perpendicular hyperplane which is perpendicular to the line across the two instances, which is also the maximum margin hyperplane. Since there is more than one CL constraint in the constraint set, multiple mid-perpendicular hypeprlanes can be obtained. MPHS contains three main steps. Firstly, it learns a new data representation using the mid-perpendicular hyperplane corresponding to each cannot-link constraint, which can also be regarded as dimensionality reduction. Secondly, it learns individual similarity matrix according to the new data representation corresponding to each CL. In the end, individual similarity matrices are aggregated into a similarity matrix and then perform kernel k-means. Three variants of MPHS are proposed in~\cite{gao2013learning}: MPHS-linear (for simple and well-structured data), which is performed on original data space, MPHS-Gauss (for complex data) which is performed on Gaussian-kernel induced feature space, and MPHS-PCP which first learns a data-dependent kernel similarity (PCP-kernel~\cite{li2008pairwise}) and performs MPHS in PCP-kernel induced feature space.

PCP is one of the few methods which projects the original data into a higher-dimensional space to include constraints. To do so, it learns a mapping over the data graph and maps the data onto a unit hypersphere, where instances related by ML are mapped into the same point and instances related by CL are mapped to be orthogonal. This can be achieved using the kernel trick via semidefinite programming, and has to be done in a high dimensional space, as implementing it in the input space is hard if not unfeasible. Another method which projects the original data into higher dimensions is CAF. It augments the initial space with additional dimensions derived from CL constraints. The instances are augmented with additional features, each of which is defined by one of the given CL constraints. ML constraints are included by modifying the initial distance matrix so that the distance between instances related by ML is the lowest among all pairwise distances (greater than 0) and restoring metricity and the triangle inequality afterwards. Pairwise distances between instances in the augmented space combines both the distances in the original space (modified to include ML) and the distances of instances according to each CL constraint. The distance derivation for the new dimension is based on diffusion maps. The actual clustering is then performed by any classic clustering technique.

\section{Statistical Analysis of the Taxonomy} \label{sec:statistic_analysis}

\begin{figure}[!ht]
	\centering
	\includegraphics[width=\linewidth]{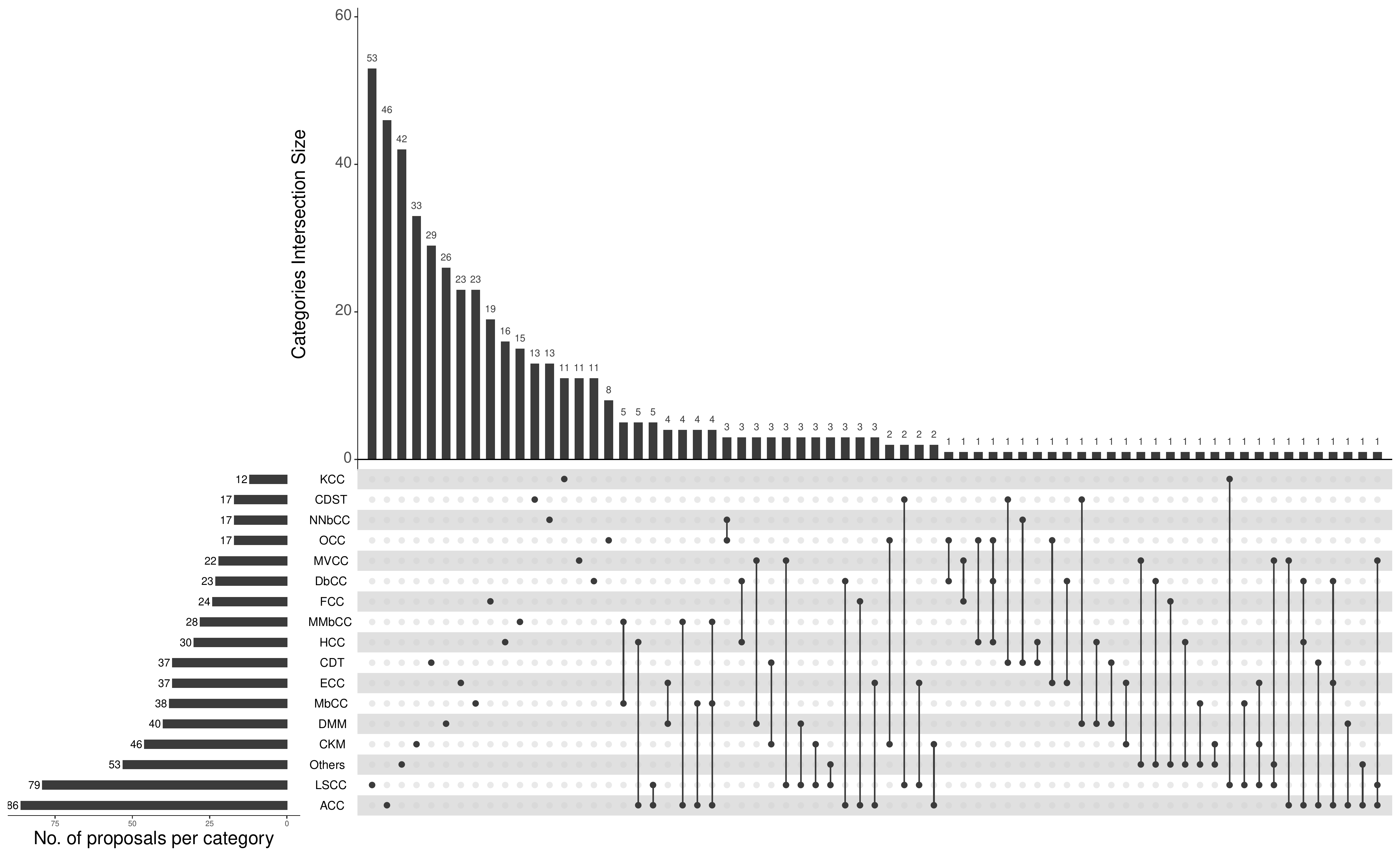}
	\caption{Summary of taxonomy categories and hybridizations.}
	\label{fig:upsetR_figure}
\end{figure}

This section presents relevant statistics on the ranked taxonomy proposed in Section~\ref{sec:Taxonomy}. The \texttt{UpSetR} package provides the perfect tool to obtain a visualization of the overall taxonomy in the form of a statistical summary, presented in Figure~\ref{fig:upsetR_figure}. The left histogram represents the number of methods which belong to every category, while the top histogram considers all hybridizations found between said categories. The categories involved in hybridizations are indicated by the central dot matrix. For example, the left histogram shows that a total of 12 methods belong to the KCC category, and the top histogram shows that 11 of them are purely KCC methods, with a single hybrid method, which the central dot matrix indicates that belongs to both KCC and LSCC category thanks to (which is consistent with the information provided in Table~\ref{tab:tax:kernel_cc}).

\begin{figure}[!h]
	\centering
	\subfloat[Proportion of Constrained Clustering and Constrained DML methods]{\includegraphics[width=0.5\linewidth]{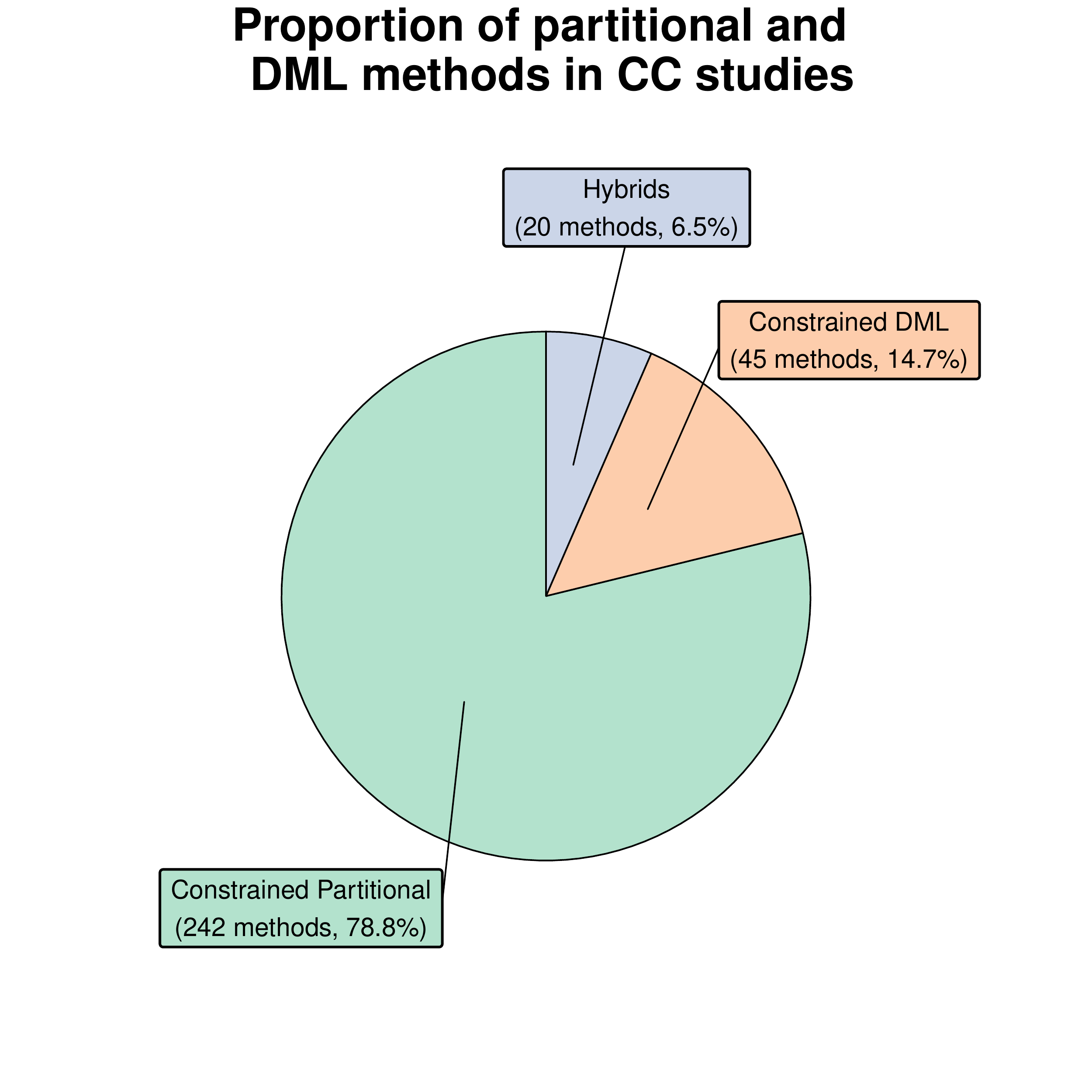}
	\label{fig:pie_taxonomy_proportions}}
	\subfloat[Proportion of constraint types used in CC methods]{\includegraphics[width=0.5\linewidth]{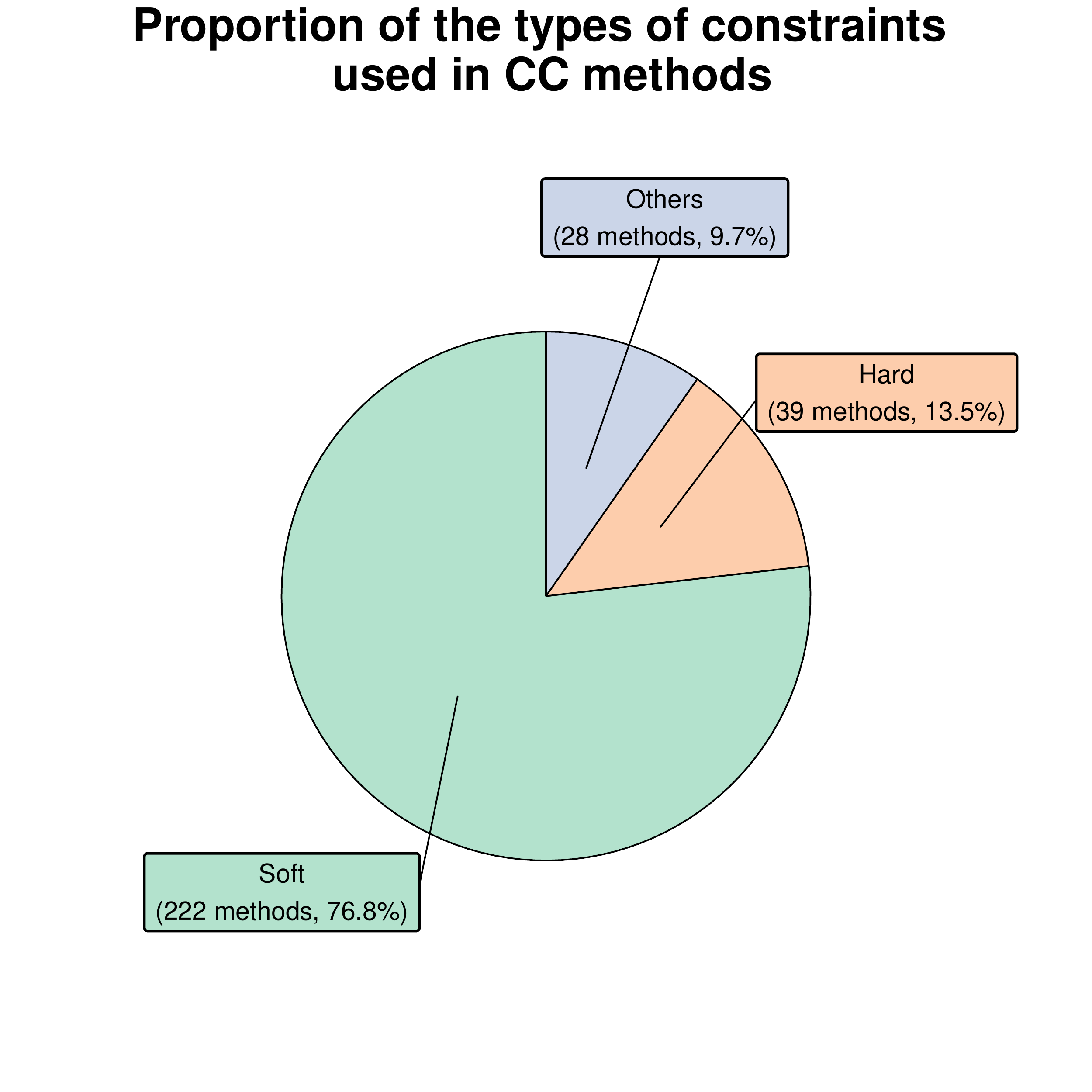}
	\label{fig:pie_constraints_types}}
	
	\caption{Piecharts on the proportions of methods in the highest dichotomy of the taxonomy and on the types of constraints.}
	\label{fig:taxonomy_statistics}
\end{figure}

Figure~\ref{fig:upsetR_figure} shows that the ACC category is the most prominent one, while also being the one which represents the most hybridizations. The rationale behind this is that all methods which belong to the active clustering with constraints category (Table~\ref{tab:tax:active_clustering_with_constraints}) always use a CC method from another category as their core CC method. In addition to this, it can be observed that the more successful hybridization are found in MMbCC+MbCC, HCC+ACC, and LSCC+ACC, with up to five hybrids methods in each combination. Please note that, for Constrained DML categories (CDT, CDST and DMM) hybridization never refer to the method used to eventually obtain a partition from their outputs, as these method are never considered to determine their category memberships. The proportion of methods which belong to the classes of the highest level dichotomy in the taxonomy (constrained partitional versus constrained DML) is presented in Figure~\ref{fig:pie_taxonomy_proportions}, which is introduced next to Figure~\ref{fig:pie_constraints_types}, depicting the proportion of the types of constraints used by these methods. From these figures is clear that the vast majority of methods belong to the constrained partitional category and that the use of soft constraints is greatly preferred over any other type of constraints. The ``Others'' portion in Figure~\ref{fig:pie_constraints_types} gathers methods that consider any combination of constraints that are not purely soft nor hard. For example, methods considering only one type of constraints, or using hard ML and soft CL are included in said portion.

\begin{figure}[!ht]
	\centering
	\includegraphics[width=0.7\linewidth]{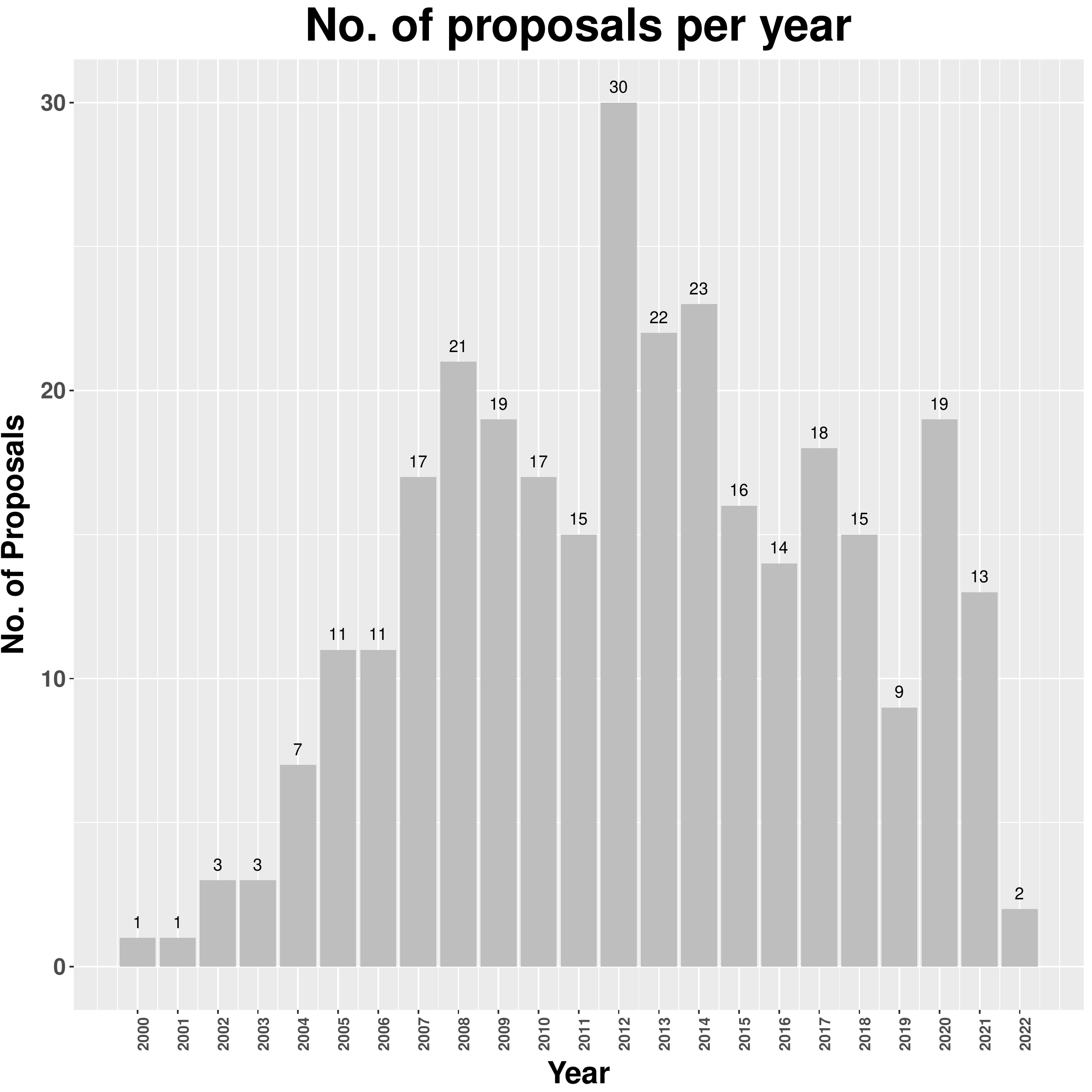}
	\caption{Number of proposals per year.}
	\label{fig:prop_per_year}
\end{figure}

Another interesting statistic is presented Figure~\ref{fig:prop_per_year}, which shows a histogram for the number of publications in the CC area sorted by year  (please note that only three months of the year 2022 are included in this figure). It is clear that this number increases consistently from 2001 to 2008, when the proficiency of the area becomes inconsistent. Authors firmly believe that this is due to the lack of a solid, general reference in the area, which may help new researchers get a general understanding on the problems and the points of view proposed to tackle them.

A relevant concern within any clustering task is the determination of the number of clusters $k$. The vast majority of both classic clustering and CC methods simply leaves this step out of the clustering process and takes $k$ as an input hyperparameter. However there are other alternatives to approach this problem. Table~\ref{tab:k_needed} gathers CC methods that do not need the number of clusters $k$ to be specified by the user, but they include procedures to determine it in some circumstances. For example, SSFCA can handle both a specification for the number of clusters and the lack of it, as it will try to automatically determine it in such case. The BoostCluster method is basically a wrapper which can be applied to any clustering method in order to include constraints in it, thus handling both cases. Other methods ask the user about an interval in which the $k$ lies in, such as CMSSCCP, 3CP and TKC(17). Besides, there are methods that can only work with a fixed number of clusters (usually $k=2$), such as CSP and ISL. The rest of the methods do not accept the number of cluster of the output partitions in their inputs, and include procedures to determine it through the clustering process.

\begin{table}[!h]
	\centering
	\setlength{\tabcolsep}{7pt}
	\renewcommand{\arraystretch}{1.3}
		\begin{tabular}{l c c l c c}
		\hline
		Acronym & Type & Ref. & Acronym & Type & Ref. \\
		\hline
SSFCA & Hybrid &~\cite{maraziotis2012semi} & URASC & Not Needed &~\cite{xiong2017active} \\
BoostCluster & Hybrid &~\cite{liu2007boostcluster} & PCCA & Not Needed &~\cite{grira2005semi,grira2006fuzzy} \\
CMSSCCP & Bounded &~\cite{dao2015constrained} & SemiDen & Not Needed &~\cite{atwa2017constraint} \\
3CP & Bounded &~\cite{calandriello2014semi} & FIECE-EM+LUC & Not Needed &~\cite{fernandes2019active,fernandes2020improving} \\
TKC(17) & Bounded &~\cite{duong2017constrained} & CECM & Not Needed &~\cite{antoine2012cecm} \\
CSP & Fixed &~\cite{wang2010flexible} & En-Ant & Not Needed &~\cite{yang2012semi} \\
ISL & Fixed &~\cite{davidson2010sat} & CAC & Not Needed &~\cite{xu2011constrained,xu2013improving} \\
SSDC & Not Needed &~\cite{ren2018semi} & CELA & Not Needed &~\cite{vu2009leader} \\
COBRAS & Not Needed &~\cite{craenendonck2018cobras} & MELA & Not Needed &~\cite{vu2009leader} \\
FIECE-EM & Not Needed &~\cite{covoes2018classification} & MCLA & Not Needed &~\cite{vu2009leader} \\
JDG & Not Needed &~\cite{covoes2018classification} & SSAP & Not Needed &~\cite{givoni2009semi} \\
Semi-MultiCons & Not Needed &~\cite{yang2022semi} & ACC & Not Needed &~\cite{frigui2008image} \\
ssFS & Not Needed &~\cite{huang2012semi} & AFCC & Not Needed &~\cite{grira2008active} \\
SFFD & Not Needed &~\cite{li2015semi} & COP-b-coloring & Not Needed &~\cite{elghazel2007constrained} \\
AAVV & Not Needed &~\cite{abin2020density} & MOCK & Not Needed &~\cite{handl2006semi} \\
FIECE-EM+BFCU & Not Needed &~\cite{fernandes2019active,fernandes2020improving} & MCGMM & Not Needed &~\cite{raghuram2014instance} \\
FIECE-EM+FCU & Not Needed &~\cite{fernandes2019active,fernandes2020improving} & COBRA & Not Needed &~\cite{van2017cobra} \\
FIECE-EM+DVO & Not Needed &~\cite{fernandes2019active,fernandes2020improving} & ASCENT & Not Needed &~\cite{li2019ascent} \\
JBMJ & Not Needed &~\cite{berg2017cost} & - & - & - \\
        \hline
		\end{tabular}
	\caption{Methods that do not need $K$ to be specified.}
	\label{tab:k_needed}
\end{table}

\subsection{Statistic Analysis of Ranked Scores}

\begin{figure}[!ht]
	\centering
	\includegraphics[width=0.9\linewidth]{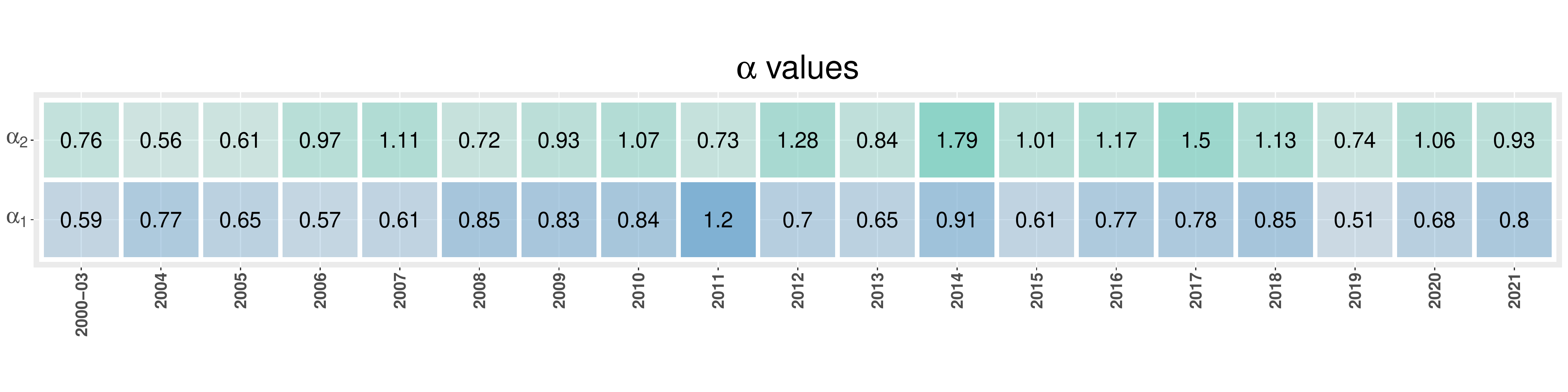}
	\caption{Values for the weighting parameters $\alpha_1$ and $\alpha_2$ used in every year.}
	\label{fig:alphas}
\end{figure}

This section tackles the distribution of the final scores, obtained by applying the methodology introduced in Section~\ref{sec:scoring_system}. Firstly, Figure~\ref{fig:alphas} gives the values for $\alpha_1$ and $\alpha_2$ for every year, so the reader can have a better understanding of how the final scores are obtained. Overall, $\alpha_1$ and $\alpha_2$ do not show sufficiently significant differences between them for the year of publication to be decisive, although enough to be used as a discriminating factor on a reasonable scale.

\begin{figure}[!ht]
	\centering
	\includegraphics[width=0.7\linewidth]{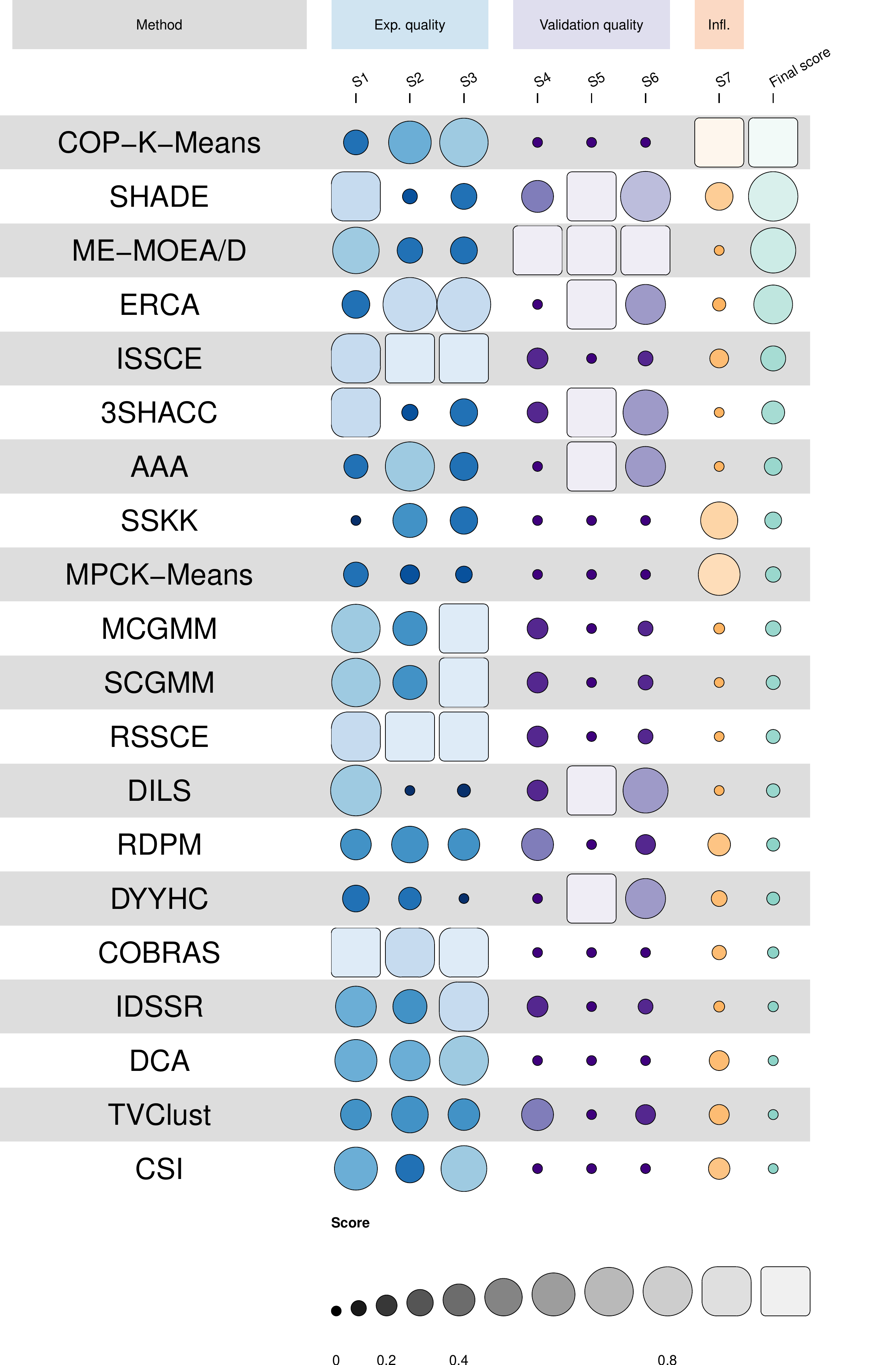}
	\caption{Detailed scorings for the top 20 ranked methods. Every subscore (S1 to S7) has its own meaning and can be understood individually as follows: the weighted year-normalized number of datasets ($\alpha_1^{Y_{\mathcal{A}}}{D'}_{\mathcal{A}}$), the weighted year-normalized number of methods ($\alpha_2^{Y_{\mathcal{A}}}{MS'}_{\mathcal{A}}$), the normalized experimental quality (${EQ'}_{\mathcal{A}}$), the normalized number of validity indices ($V'_{\mathcal{A}}$), the statistical test usage indicator ($T_{\mathcal{A}}$), the normalized validation procedure quality ($VQ'_{\mathcal{A}}$), the normalized influence (${I'}_{\mathcal{A}}$). Finally, ${S}_{\mathcal{A}}$ is given in the final score column.}
	\label{fig:detailed_scorings}
\end{figure}

Figure~\ref{fig:detailed_scorings} gives a visual representation of the detailed scorings for the top 20 best ranked methods. As expected, the best score is obtained by the COP-K-Means method, as it features a high quality experimental setup (S1, S2 and S3) and it is the single most cited CC method ever proposed (S7). This is enough for it to obtain the highest score, even if the paper which proposes it lacks of a proper validation procedure (S4 and S5). Please note that the fact that COP-K-Means is ranked as the best method ever proposed is evidence of the scoring system being reluctant to the number of years the method has been available. As a result, one of the oldest CC method reaches top 1 in the ranking, and is followed by a method which was proposed as recently as in 2021, which is SHADE$_{CC}$. A number of baseline method make it to the top 20, such as ERCA, SSKK, MPCK-Means, DCA or CSI. The reason behind this is the high influence of the ${I'}_{\mathcal{A}}$ term over the final score ${S}_{\mathcal{A}}$. Other methods reach the top 20 by other means, such as a high experimental quality combined with proper validation procedures.

\begin{figure}[!ht]
	\centering
	\includegraphics[width=0.7\linewidth]{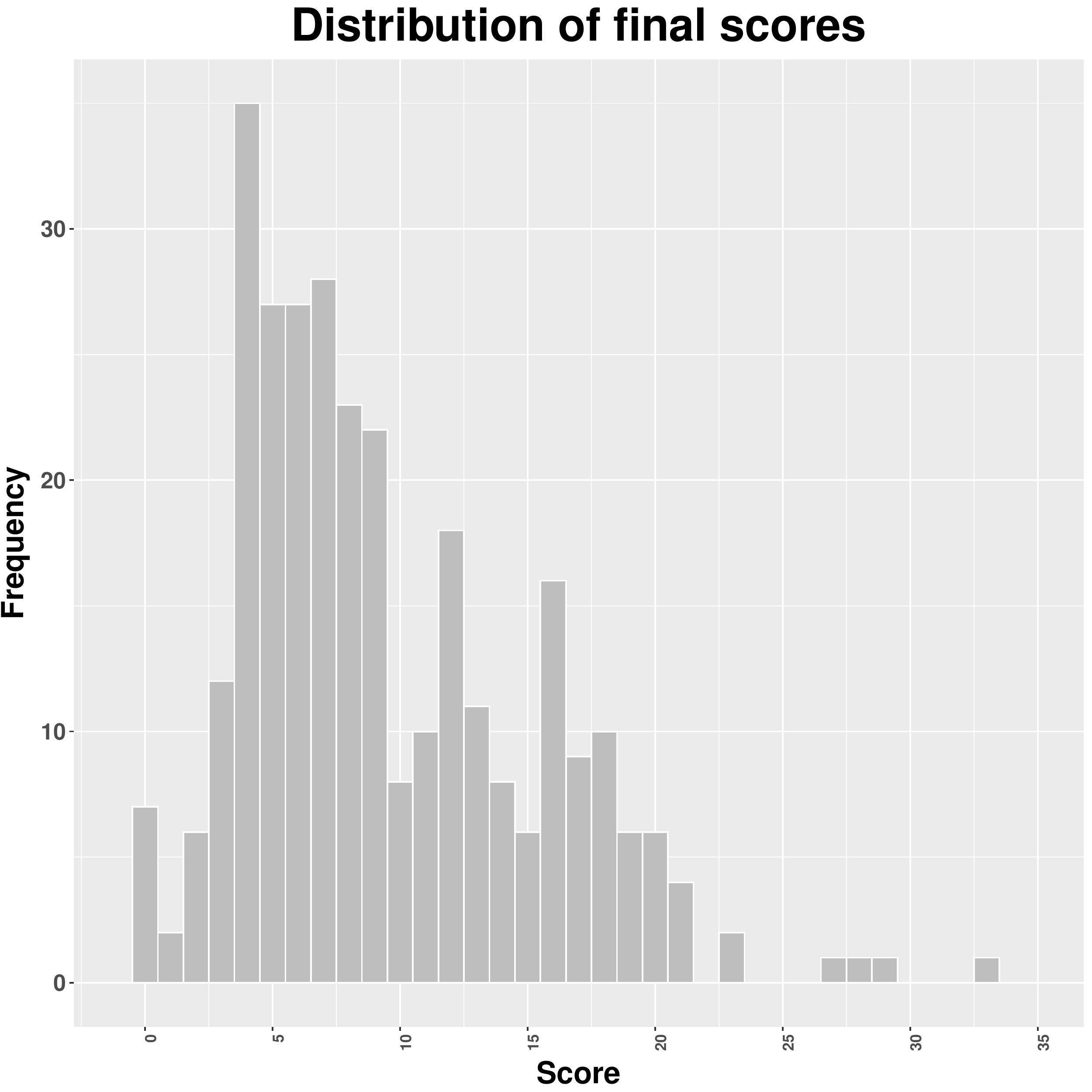}
	\caption{Distribution of all final scores presented in Section~\ref{sec:Taxonomy}.}
	\label{fig:final_score_hist}
\end{figure}

Finally, Figure~\ref{fig:final_score_hist} shows a histogram with the distribution of all final scores presented in Section~\ref{sec:Taxonomy}. It is clear that the majority of methods are scored in $[0, 10]$, fewer methods are in the next higher range of values, which is $(10, 20]$, and very few outlier scores are found in the range $(20, 35]$. Please note that the effective output range of the scoring systems is $[0,33.33]$, as no method fully complies with its standards. This shows the suitability of the proposed scoring system, as any objective raking procedure should place the majority of methods in the low-medium range and few methods in the upper range of the ranking, with those methods being the more remarkable ones for their quality and the robustness of their conclusions.

\section{Conclusions, Criticisms and Future Research Guidelines} \label{sec:conclusions}

This study presented a systematical review on the Constrained Clustering (CC) research domain. Firstly, a general introduction to the Semi-Supervised Learning (SSL) paradigm is given, after which the specific area of semi-supervised clustering is discussed in detail. The discussed methods within this area are capable of including background knowledge (or incomplete information) about the dataset onto the clustering process. To the best of the authors' knowledge, this study provides the first ever overview and taxonomization of the types of background knowledge (in Section~\ref{sec:Clustering_with_BK}) that can be included as constraints in semi-supervised clustering. Afterwards, we motivate why out of all types of background knowledge, the instance-level pairwise Must-Link (ML) and Cannot-Link (CL) constraints, are the most successful and prominent types of constraints. Semi-supervised clustering methods that use ML and CL constraints are known as Constrained Clustering (CC). The CC problem is formalized and presented in later sections, and illustrated by giving examples of several practical fields of applications. Afterwards, advanced CC concepts and structures are described and formalized, allowing to discuss in more detail the advantages and disadvantages of different approaches. Afterwards, a statistical analysis of the experimental elements used in studies which proposes new CC methods has been carried out, revealing the basics research in CC area, as well as the baseline methods, benchmark datasets and suitable validity indices. This overview leads to the proposal of an objective scoring system that captures the potential and relevance of each approach, which is used later to assign every method a score that summarizes its quality, and to produce a ranking of CC methods. Afterwards, a taxonomization of 307 CC methods is conducted, ranking them according to the proposed scoring system and categorizing them in two major families: constrained partitional and constrained DML. These two families are further divided in more specific categories, whose common features and specific methods are described in Sections~\ref{subsec:tax:constrained_kmeans} to~\ref{subsec:tax:cddt}. 

The proposed taxonomy can be used to:

\begin{itemize}

    \item Decide which type of approach and model is best suited to a new constrained clustering problem.
    
    \item Compare newly proposed techniques to those belonging to the same family in this taxonomy, so that in can be determined whether the new method represents an improvement over the current state-of-the-art.

    \item  Identify the proposals which best support their conclusions and propose more robust methods, thanks to the scoring system.
    
\end{itemize}

As any other Computer Science research area, the CC area is not free of flaws and criticism. Having reviewed 270 studies (proposing 307 methods), the authors have identified 5 major problems which affect the vast majority of them. These problems can be summarized as follows:

\begin{itemize}
    
    \item \textbf{The lack of a unified, general reference.} There is not an updated, general reference unifying the overall CC literature. This affects the foundations of new proposals, as it is hard to find the state-of-the-art methods that can be used to compare new techniques with. This is illustrated in Figure~\ref{fig:pieplot_1}, which shows that more than half of the proposed CC methods (52.1\%) are never used in subsequent studies. The aim of this study is to address this deficiency.
    
    \item \textbf{Low number of application studies}. Even if the main purpose of this study is not to review and gather literature concerning CC applications, it has been difficult for the authors to find application studies other than the 95 presented in Section~\ref{subsec:cc_applications}. This is specially significant, given the high number and diversity found in CC studies which propose new methods.
    
    \item \textbf{The lack of extensive experimental comparisons.} From Section~\ref{sec:exp_setup}, it is clear that, unfortunately, the CC research area is consistently poor regarding the number of datasets, validity indices and competing methods used to support their conclusions. Very few methods use more than 10 datasets or 2 validity indices. With respect to the number of competing methods, it is shocking that no study has ever considered more than 9 of them, given that there are, at least, 307. The authors firmly believe that (open-source) code unavailability is responsible for this unsettling fact, as implementation details are rarely given in CC studies.
    
    \item \textbf{Unavailability of specific standardized CC-oriented datasets and constraint sets.} One of the major flaws of the CC area is the lack of specific datasets, which may be justified, provided that the low number of application papers. However, this is not the case when it comes to the constraint sets. In Section~\ref{subsec:generation_methods}, the constraint generation method used in the vast majority of CC studies was presented. It is clear that this procedure is highly dependent of random effects, as constraints are randomly allocated. For this reason, it is necessary to have access to the specific constraint sets used in a given experimentation if it needs to be reproduced. Unfortunately, constraint sets are only rarely published by authors.
    
    \item \textbf{Statistically unsupported experimental conclusions}. This is an effect derived from the two previous criticisms, as a low number of standardized experiments is not significant enough to perform statistical testing procedures or to derive generalized conclusions that are broadly applicable. In fact, a very reduced minority of studies support their conclusions using statistical testing. This may greatly affect the confidence future researches may have towards the reviewed studies, reducing their usability in both the development of new methods and their applications in real-world problems.
    
\end{itemize}

In respect to future research guidelines, new proposals would greatly benefit from avoiding the mentioned flaws as much as possible. New studies should perform extensive experimental comparisons with state-of-the-art methods. A major goal of this overview is to provide easy access to such methods. Similarly, studies would benefit from the use of multiple datasets focusing on a wide range of application domains to make conclusions more generalisable. To this end, authors should always make their datasets and constraint sets public, in order for the result to be reproducible, and preferably also the source code. Supporting conclusions with statistical testing is also essential, as this will increase confidence in the results. 

 Future research based on this study can extend the scoring system to evaluate not only quantitative features regarding the quality of the proposal, but also qualitative features, as their novelty or their applicability. This objective would be arduous to achieve with objective standards in mind, as these features greatly depend on the perception of the researchers. Additionally, the constraint equivalences presented in Section~\ref{subsec:equivalencies} would benefit from formalization. Lastly, the creation of a library of CC baseline algorithms would greatly benefit this research area, making it more accessible to new researchers. Thanks to this study, the mentioned CC baselines algorithms can be chosen using objective criteria.

\section*{Acknowledgements}

Our work has been supported by the research projects PID2020-119478GB-I00, A-TIC-434-UGR20 and PREDOC\_01648.

\section*{Conflict of interest}

The authors declare that there is no conflict of interest.

\clearpage

 \bibliographystyle{unsrt} 
 \bibliography{references}





\end{document}